\documentclass{thesisclass}

\usepackage[english]{babel}
\usepackage{graphicx}
\usepackage{amsmath} %Milo for better maths formulas
\usepackage{amssymb} %Milo for better maths symbols
\usepackage{wrapfig} %Milo to wrap figures in line
\usepackage{tabularx} %Milo to do better tables
\usepackage{multirow} %Milo to merge different rows on certain levels
\newcolumntype{C}[1]{>{\centering\arraybackslash}p{#1}} %Milo Column centering
\DeclareGraphicsExtensions{.pdf,.png,.jpg}
\graphicspath{{./figures/}} %Use curly braces for each path to add and don't
\usepackage[natbibapa]{apacite}
\usepackage{booktabs}

\usepackage{scrhack}

\newenvironment{nscenter}
 {\parskip=0pt\par\nopagebreak\centering}
 {\par\noindent\ignorespacesafterend}

\usepackage{xcolor}
\hypersetup{
    colorlinks,
    linkcolor={blue!75!black},
    citecolor={blue!75!black},
    urlcolor={blue!75!black}
}

\usepackage{nomencl}

\makenomenclature
\usepackage[normalem]{ulem}

\newcommand{\mytype}{\iflanguage{english}{Master Thesis}{Masterarbeit}} 

\newcommand{\myname}{Milo R. Honegger}
\newcommand{\matricle}{1505115}
\newcommand{\mytitle}{\iflanguage{english}{Shedding Light on Black Box Machine Learning Algorithms}{Titel der Arbeit}}
\newcommand{\mysubtitle}{\iflanguage{english}{Development of an Axiomatic Framework to Assess the Quality \\of Methods that Explain Individual Predictions}{Subtitel der Arbeit}}
\newcommand{\myinstitute}{\iflanguage{english}
{Institute of Information Systems and Marketing (IISM) \\
Information \& Market Engineering}
{Institut für Informationswirtschaft und Marketing (IISM) \\
Information \& Market Engineering}}

\newcommand{\reviewerone}{Prof. Dr. rer. pol. Christof Weinhardt}
\newcommand{\reviewertwo}{Prof. Dr. Alexander Maedche}
\newcommand{\advisor}{Rico Knapper}
\newcommand{\advisortwo}{Dr. Sebastian Blanc}

\newcommand{\timeend}{\iflanguage{english}{15th of August 2018}{XX. Monat 20XX}}
\newcommand{\submissiontime}{15.08.2018}

\hypersetup{
 pdfauthor={\myname},
 pdftitle={\mytitle},
 pdfsubject={\mytype},
 pdfkeywords={\mytype}
}

\begin{document}

\frontmatter
\pagenumbering{roman}
%% titlepage.tex
%%

% coordinates for the bg shape on the titlepage
\newcommand{\diameter}{20}
\newcommand{\xone}{-15}
\newcommand{\xtwo}{160}
\newcommand{\yone}{15}
\newcommand{\ytwo}{-253}

\begin{titlepage}
% bg shape
\begin{tikzpicture}[overlay]
\draw[color=gray]  
 		 (\xone mm, \yone mm)
  -- (\xtwo mm, \yone mm)
 arc (90:0:\diameter pt) 
  -- (\xtwo mm + \diameter pt , \ytwo mm) 
	-- (\xone mm + \diameter pt , \ytwo mm)
 arc (270:180:\diameter pt)
	-- (\xone mm, \yone mm);
\end{tikzpicture}
	\begin{textblock}{10}[0,0](4,2.5)
		\iflanguage{english}
			{\includegraphics[width=.3\textwidth]{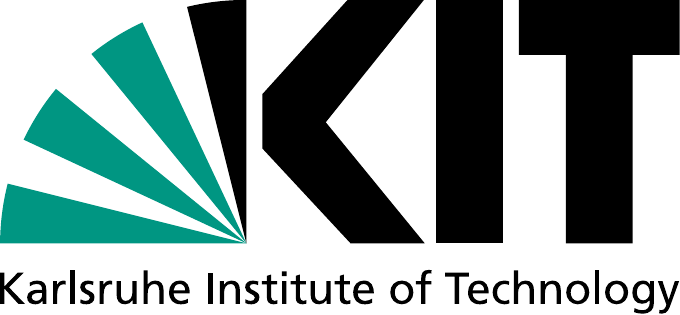}}
			{\includegraphics[width=.3\textwidth]{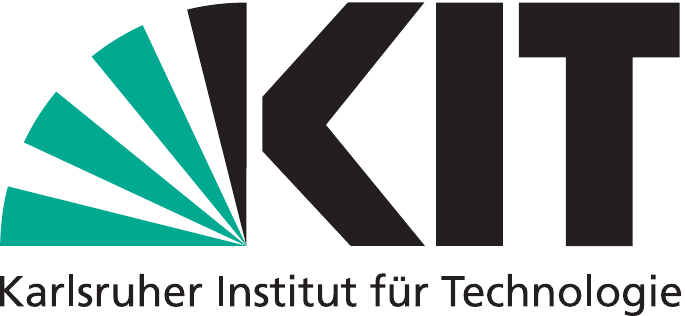}}		
	\end{textblock}
	\begin{textblock}{10}[0,0](10,2.5)
		\includegraphics[width=.6\textwidth]{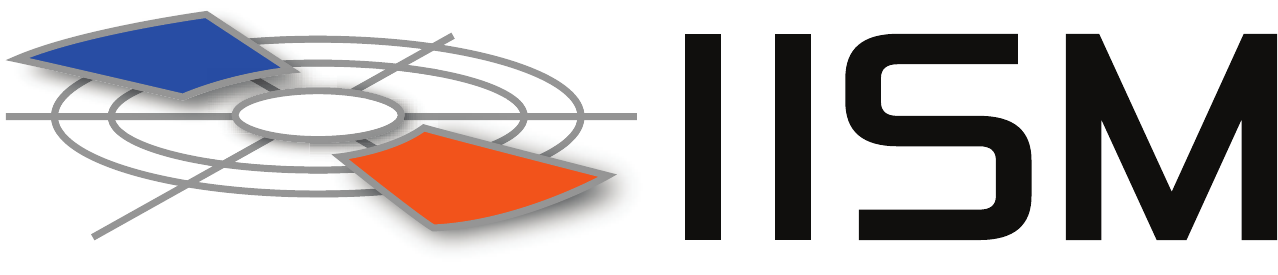}
	\end{textblock}
	\changefont{ppl}{m}{n}	% helvetica	(phv), % IM Style: palatino (ppl) 
	\vspace*{2.0cm}
	\begin{center}
		\Huge{\mytitle}
		\vspace*{0.5cm}\\
		\Large{\mysubtitle}
		\vspace*{2.0cm}\\
		\Large{
			\iflanguage{english}{\mytype}			
								{\mytype}
		}\\
		\vspace*{2.0cm}
		\huge{\myname}\\
			\Large{\matricle}\\
		\vspace*{2.0cm}
		\Large{
			\iflanguage{english}{At the Department of Economics and Management}			
							    {An der Fakult\"at f\"ur Wirtschaftswissenschaften}
			\\
			\myinstitute
		}
	\end{center}
	\vspace*{1.0cm}
\Large{
\begin{center}
\begin{tabular}[ht]{l c l}
  % Gutachter sind die Professoren, die die Arbeit bewerten. 
  %\iflanguage{english}{Reviewer}{Erstgutachter}: & \hfill  & \reviewerone\\
  \iflanguage{english}{Reviewer}{Gutachter}: & \hfill  & \reviewerone\\
  \iflanguage{english}{Second reviewer}{Zweitgutachter}: & \hfill  & \reviewertwo\\
  \iflanguage{english}{Advisor}{Betreuender Assistent}: & \hfill  & \advisor\\
  \iflanguage{english}{Second advisor}{Zweiter betreuender Mitarbeiter}: & \hfill  & \advisortwo\\
  % IM: No second advisor
  % Der zweite betreuende Mitarbeiter kann weggelassen werden. 
\end{tabular}
\end{center}
}

\vspace{1.0cm}
\begin{center}
\large{\timeend}
% \iflanguage{english}{Duration:}{Bearbeitungszeit:} \timestart \hspace*{0.25cm}
% -- \hspace*{0.25cm} 
% \timeend}
\end{center}

\begin{textblock}{10}[0,0](4,16.8)
\tiny{ 
	\iflanguage{english}
		{KIT -- University of the State of Baden-Wuerttemberg and National Laboratory of the Helmholtz Association}
		{KIT -- Universit\"at des Landes Baden-W\"urttemberg und nationales Forschungszentrum der Helmholtz-Gesellschaft}
}
\end{textblock}

\begin{textblock}{10}[0,0](14,16.75)
\large{
	\textbf{www.kit.edu} 
}
\end{textblock}

\end{titlepage}
% * <honegger.milo@gmail.com> 2018-04-23T12:01:36.631Z:
%
% ^.
% * <honegger.milo@gmail.com> 2018-04-23T12:01:35.159Z:
%
% ^.

%\blankpage % IM Style: No additional blank page

%% -------------------
%% |   Directories   |
%% -------------------

%Inserting an abstract:
\blankpage
\chapter*{\centering Abstract}
From self-driving vehicles and back-flipping robots to virtual assistants who book our next appointment at the hair salon or at that restaurant for dinner – machine learning systems are becoming increasingly ubiquitous. The main reason for this is that these methods boast remarkable predictive capabilities. However, most of these models remain black boxes, meaning that it is very challenging for humans to follow and understand their intricate inner workings. Consequently, interpretability has suffered under this ever-increasing complexity of machine learning models. Especially with regards to new regulations, such as the General Data Protection Regulation (GDPR), the necessity for plausibility and verifiability of predictions made by these black boxes is indispensable.
\\Driven by the needs of industry and practice, the research community has recognised this interpretability problem and focussed on developing a growing number of so-called explanation methods over the past few years. These methods explain individual predictions made by black box machine learning models and help to recover some of the lost interpretability. With the proliferation of these explanation methods, it is, however, often unclear, which explanation method offers a higher explanation quality, or is generally better-suited for the situation at hand. In this thesis, we thus propose an axiomatic framework, which allows comparing the quality of different explanation methods amongst each other. Through experimental validation, we find that the developed framework is useful to assess the explanation quality of different explanation methods and reach conclusions that are consistent with independent research.

\noindent\textbf{Keywords:} black box, machine learning, interpretability, explanation methods, explanation quality, axiomatic explanation consistency

\blankpage

%Inserting acknowledgements:
\blankpage
\chapter*{\centering Acknowledgements}
I would like first to express my gratitude and thank my thesis advisor Rico Knapper, for his untiring support and invaluable guidance. He consistently allowed this thesis to be my work but also steered me in the right direction when needed. Also, I would like to thank anacision for making this research possible, and for entrusting me to conduct this thesis. Furthermore, I take this opportunity to thank my second advisor Dr. Sebastian Blanc, and all team members at anacision who provided vital insights and expertise that greatly assisted this research.
\\I would also like especially to thank my reviewer Prof. Dr. rer. pol. Christof Weinhardt from the IISM institute at the KIT, for his openness towards and interest in industry-oriented research.
\\I would like to further acknowledge Rainer Speicher, Martin Albrecht and Manuel Bauer, for their passionate participation and helpful comments on this thesis.
\\Finally, I express my very profound gratitude to my family and loved ones, who provided me with unfailing support and spiritual encouragement throughout my years of study at the KIT, and through the process of researching and editing this thesis. This achievement would not have been possible without them. Thank you.
\begin{flushright} Milo R. Honegger \end{flushright}
\blankpage

\tableofcontents
% IM Style: No additional blank page
% \blankpage

% Do not include a list of figures, list of tables and list of abbreviations, if the work is a seminar thesis
\iflanguage{english}{
\ifthenelse{\equal{\mytype}{Seminar Thesis}}
{}
{
\listoffigures \addcontentsline{toc}{chapter}{List of Figures} 
\listoftables  \addcontentsline{toc}{chapter}{List of Tables} 
\printnomenclature   \addcontentsline{toc}{chapter}{List of Abbreviations} 
}
}
{
\ifthenelse{\equal{\mytype}{Seminararbeit}}
{}
{
\listoffigures \addcontentsline{toc}{chapter}{Abbildungsverzeichnis} 
\listoftables \addcontentsline{toc}{chapter}{Tabellenverzeichnis} 
\printnomenclature \addcontentsline{toc}{chapter}{Abkürzungsverzeichnis}
}
}

% Execute this command for index creation, i.e., for abbreviations by the nomencl package
% makeindex thesis.nlo -s nomencl.ist -o thesis.nls

%\printnomenclature

%% -----------------
%% |   Main part   |
%% -----------------
\mainmatter
\pagenumbering{arabic}
%% ===========================
%% ===========================
\chapter{Introduction}
\label{ch:Introduction1}
%% ===========================
%% ===========================

\hspace{\parindent}Products and services that we use on a regular basis, increasingly feature some Artificial Intelligence (AI)\nomenclature{AI}{Artificial Intelligence}. Often-so, without us realising that it is there. As an example, when we talk to our virtual personal assistant and ask it about the weather, our schedule or the closest Chinese restaurant, we are engaging in a conversation with an AI. Even when we use an automated online translation service, an AI is running in the background. It compares the given text to billions of other similar texts and contexts, which have already been translated. It tries to combine those words in such a way that the translated text sounds more natural, more human-like in the target language.

Even though AI is currently a frequently used term in technology, there is some confusion on its meaning. Especially how it relates to, and what distinguishes it from machine learning (ML)\nomenclature{ML}{Machine Learning} and deep learning (DL)\nomenclature{DL}{Deep Learning}, which are used just as often. Sometimes,
%%%%%%%%%%%%% Figure %%%%%%%%%%%%%
\begin{wrapfigure}{r}{0.45\textwidth}
  \vspace{-20pt}
  \begin{center}
    \includegraphics[width=0.45\textwidth]{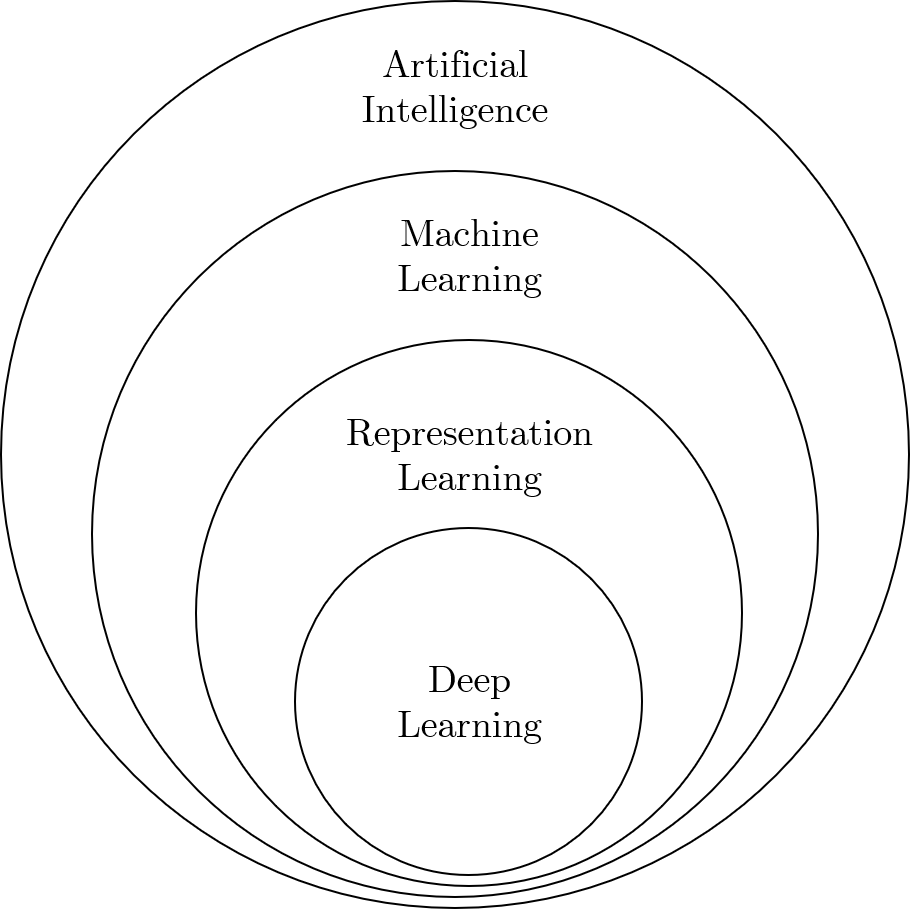}
  \end{center}
  \vspace{-20pt}
  \caption{Venn Diagram of AI, ML, RL and DL [based on \cite{Goodfellow-et-al-2016}]}
  \label{figure 1.1}
  \vspace{-10pt}
\end{wrapfigure}
%%%%%%%%%%%%% Figure %%%%%%%%%%%%%
these terms are even used interchangeably as if they were synonymous. The meaning and breadth of each of these terms become, however, more apparent, when we look at figure \ref{figure 1.1}. As we can see, AI encompasses ML, representation learning (RL)\nomenclature{RL}{Representation Learning} and DL, or in more formal jargon, AI is the superset of ML, RL and DL.
\\The Merriam-Webster dictionary defines AI as "the capability of a machine to imitate intelligent human behaviour"\footnote{\url {https://www.merriam-webster.com/dictionary/artificial\%20intelligence}}. Thereby, a task following a deterministic rule-based process could already be considered an AI. As an example, teaching a manufacturing machine to process and prepare different car parts. A component may e.g. need trimming so that the desired geometry is obtained. This task can be fully and deterministically described with a list of programmed coordinates and settings provided to the machine ahead of time. In artificial intelligence, there is a further distinction to be made between weak AI (focuses on simpler tasks) and strong AI (relates to intelligent systems that can find their own solution to a problem).
\\One of the earliest descriptions of Machine Learning was given by \cite{samuel1959}. The author stated that ML is a set of techniques that enables computer systems to "learn" and thus to progressively improve their performance on a specific task, which eliminates the need for detailed and explicit programming. It is thus a system that can acquire its knowledge from raw data by extracting patterns \citep{Goodfellow-et-al-2016}. As an example, a ML model could learn to predict the failure of a machine based on sensory data.
\\In representation learning, the algorithm does not only learn the mapping from representation to output as in ML, but also the representation itself \citep{Goodfellow-et-al-2016}. Such an algorithm can discover an adequate set of features in hours or days, where it can take an entire research community a decade \citep{Goodfellow-et-al-2016}. An excellent example of a RL algorithm is an autoencoder, which consists of an encoder (converts inputs into a different representation) and a decoder (converts the new representation back to the old format). Autoencoders are e.g. used for compressing data (ZIP files).
\\The innermost circle in our Venn diagram corresponds to deep learning. In DL we implement machine learning algorithms through so-called artificial neural networks (ANN)\nomenclature{ANN}{Artificial Neural Network}, or just Neural Networks (NN)\nomenclature{NN}{Neural Network}\citep{Goodfellow-et-al-2016}. DL is a further extension of representation learning, as it expresses representations through other, more straightforward representations, thereby enabling a computer to build complex concepts out of simpler ones \citep{Goodfellow-et-al-2016}. DL models have leveraged many of the most complex tasks, such as autonomous cars, robots and drones, chat-bots, and object and speech recognition.

The different types of AI introduced above can be used to solve increasingly complicated problems. As an unavoidable consequence, the model complexity grows substantially, which often rewards users with models of higher accuracy and predictive power. However, it usually comes at the expense of human interpretability, which represents a significant pitfall \citep{lundberg2017,ribeiro2016,lou2012,kamwa2012,ruping2006}.
\\Let us imagine a scenario in which an original equipment manufacturer carries out a predictive maintenance project. The desired outcome of that project is a model or system, which accurately predicts the failure and maintenance need of different machines ahead of time. This system could rely on an architecture with, e.g. deep learning (i.e. neural networks), which promises a higher accuracy than traditional approaches such as a linear regression. The problem with the DL approach, however, is that development engineers and further stakeholders of the project may not accept and trust it, as they are not able to follow and understand how the system derives a prediction. Therefore, if we still want to pursue a DL approach, the question is, if we can implement a separate method that is able to explain the predictions of the neural network.
\\Therefore, especially for practical applications, the need for a solution that enables the understanding of predictions made by these so-called black box (BB)\nomenclature{BB}{Black Box} models is indispensable. These are models, which usually yield highly accurate results, however, are not easily interpreted and understood by humans due to their intricate inner workings. Furthermore, there is also regulatory pressure to increase accountability and transparency in AI, by initiatives such as the General Data Protection Regulation (GDPR)\nomenclature{GDPR}{General Data Protection Regulation}, which has entered into force in the European Union (EU)\nomenclature{EU}{European Union} as of the 25\textsuperscript{th} of May 2018 \citep{doshi2017}.

In the following sections, we discuss some background information and further motivate this thesis. After that, we formulate the problem and derive the research question. Moreover, we introduce the research methodology and outline the structure of this study.

%% ===========================
\section{Background and Motivation}
\label{ch:Introduction1:sec:Section1}
%% ===========================
This thesis project has been researched and developed by the author and the help of the advisers at \textit{\href{https://www.anacision.de/}{anacision GmbH}}, a German data science consulting company, which delivers cutting-edge solutions for predictive maintenance and other areas of application. The company has consulted numerous customers ranging from small and medium-sized enterprises to DAX\nomenclature{DAX}{German Stock Index (Deutscher Aktienindex)}-Corporations. In a large number of proof of concepts and development projects, anacision has been able to create monetary value, increase process understanding and efficiency, develop innovations and point out new business models.
\\Clients of anacision increasingly face the interpretability problem described in the introduction. Even though the demand for ever-more accurate models is high, interpretability is an equally-important goal that clients do not want to sacrifice or abandon.

A possible answer to this interpretability problem are explanation methods (EM)\nomenclature{EM}{Explanation Method}, which make individual predictions understandable and increase the model transparency. In figure \ref{figure 1.2}, we illustrate the concept of EMs with two scenarios. In the first one (on the left), no explanation method is used. The user thereby gives some input $X$ to the BB model, which returns the prediction output $\hat{y}$. The smaller the difference is between $\hat{y}$ and
%%%%%%%%%%%%% Figure %%%%%%%%%%%%%
\begin{wrapfigure}{l}{0.50\textwidth}
  \vspace{-20pt}
  \begin{center}
    \includegraphics[width=0.50\textwidth]{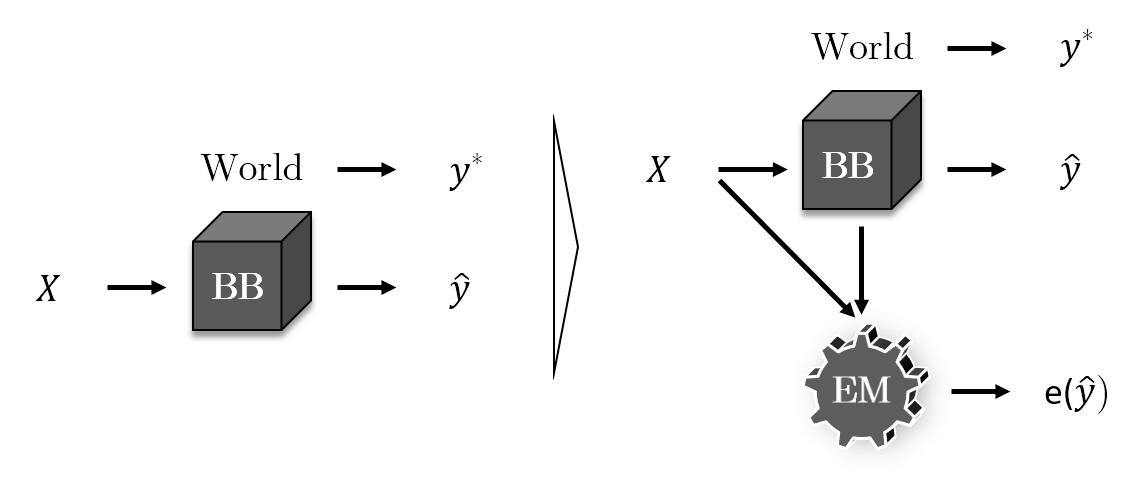}
  \end{center}
  \vspace{-20pt}
  \caption{Black Box ML Model without vs with an Explanation Method}
  \label{figure 1.2} %Serves to create a link to this image later in the text
  \vspace{-10pt}
\end{wrapfigure}
%%%%%%%%%%%%% Figure %%%%%%%%%%%%%
the real value $y^*$, the smaller is the prediction error, and thus, the more accurate and realistic would the prediction be. However, even if the prediction $\hat{y}$ is very close to the real value $y^*$, due to the intricate inner workings of the BB model, we would not be able to understand how the model arrived at its prediction value $\hat{y}$. In the second scenario, we add an explanation method to the picture, which based on the input $X$ and the prediction $\hat{y}$ produces the corresponding explanation $e(\hat{y})$. So instead of only getting the prediction $\hat{y}$, now the user also gets an explanation $e(\hat{y})$ for the prediction, which dramatically improves her understanding and trust in the model \citep{ribeiro2016,lundberg2017}. Such an explanation $e(\hat{y})$ can take on many forms, as we see later on. Let us assume for now that an explanation is a statement that tells us which features of $X$ most influenced the prediction $\hat{y}$ of our BB model. If our model, e.g. predicts house prices, an explanation could tell us something such as "the above-average size of the house raised its price by 50k and its excellent location by another 30k". Using EMs to provide explanations for BB predictions thereby offers the following two main advantages: (1) They facilitate the detection of faulty model behaviour, and of biased input data; and (2) They allow to extract insights and knowledge that the prediction model has learned from hidden structures and patterns in the data \citep{doshi2017_rigorous_science}.
\\Explanation methods can therefore potentially be a solution to our interpretability problem. However, many different EMs exist, which generate different explanations for the same predictions. So the question that arises is – how do we know if an EM and its generated explanations are of good quality? The short answer to that question is that explanations should fulfil specific quality criteria. In this thesis, we conduct preliminary research, in which we take the first steps for deriving such quality criteria that allow us to assess and compare different EMs concerning their quality.

Having discussed the fundamental concepts and problems with regards to EMs, and knowing that these are potentially indispensable tools to increase the interpretability and transparency of BB models, we continue to the next section. Hereafter, we formally define the research problem and question of this thesis.

%% ===========================
\section{Problem Statement}
\label{ch:Introduction1:sec:Section2}
%% ===========================
Over the last few years, the interpretable ML research community has been active in researching and developing new EMs. With a plethora of new methods being published, there is, however, no consensus on how to assess the quality of these methods, application areas in which they excel, and other fields in which they demonstrate weaknesses.
\\To evaluate and compare the performance of different classifiers, there are many metrics available. In classification, we can e.g.\nomenclature{E.g.}{Exempli Gratia (For Example)} calculate measures such as the area under the curve (AUC)\nomenclature{AUC}{Area Under the Curve} in the receiver operating characteristic curve (ROC)\nomenclature{ROC}{Receiver Operating Characteristic}, or we can simply compute the confusion matrix (CM)\nomenclature{CM}{Confusion Matrix}. In regression, we can e.g. calculate measures such as the mean squared error (MSE)\nomenclature{MSE}{Mean Squared Error} or the coefficient of determination (COD)\nomenclature{COD}{Coefficient of Determination (or R-Squared)}, i.e.\nomenclature{I.e.}{Id Est (That Is)} R\textsuperscript{2} measure.
\\When it comes to evaluating the quality of an explanation given by an EM, and comparing the quality of explanations given by different EMs, there are no such metrics available. This is mainly due to the fact that the concept of interpretability, by which we measure the quality of an explanation is somewhat dependent on external factors (e.g. experience of the user) and subjective (e.g. a "good" explanation for one user may not be an "understandable" explanation for another user) \citep{lipton2016,ribeiro2016,bibal2016,ruping2006}. We, therefore, define our \textbf{research problem} as follows:

\vspace{-7pt}
\begin{center}
\textit{"How can we assess the quality of different explanation methods, and thus, compare \\them to each other in terms of their strengths and weaknesses?"}
\end{center}
\vspace{-7pt}

With our research problem defined, we proceed to derive our research question, and an appropriate methodology to answer the question in the following subsections.

%% ===========================
\subsection{Research Question}
\label{ch:Introduction1:sec:Section2:Subsection1}
%% ===========================
In the previous section, we have learned that there are already several different types of EMs, with new ones being frequently published. However, there is little consensus on how to compare different EMs regarding their quality, strengths and weaknesses.
\\In this thesis, we thus strive to develop a framework for exactly that purpose, and define our \textbf{overall research question} as follows:

\vspace{-7pt}
\begin{center}
\textit{"How can an adequate framework be designed, which enables to assess the quality of explanation methods, used to explain predictions made by black box models?"}
\end{center}
\vspace{-7pt}

\noindent This overall research question consists of three components: (1) Black box models, (2) Quality of explanation methods, and (3) Framework.
\\First, we have already established that BB models usually yield highly accurate prediction results. However, the complexity of their inner workings makes them opaque, and consequently, users are most often not able to understand the model's behaviour \citep{ribeiro2016}. This first component represents the \textit{radix causa} – the starting point for our research question.
\\Second, it is imperative that the quality of explanations given by an EM is high, i.e. the explanation should be accurate. Needless to say that it is a trivial task to generate an easily understandable explanation that has no connection to the data and the prediction \citep{ruping2006,bibal2016}. The quality problem of explanations across different EMs is what we wish to address with the present work.
\\Third, with the framework developed in this work, we strive to tackle the quality issue. This framework is the basis on which we can compare any types of explanation methods amongst each other concerning their interpretability. It represents the answer to our overall research question. To assess the suitability of the developed framework, we refine our overall question with the two following research questions.

\noindent\textbf{Research Question 1:} \textit{What characterizes a good approach to compare different explanation methods amongst each other?}
\\The most common method that other researchers have used so far to compare the quality of different explanations amongst each other, are questionnaires with real people \citep{ribeiro2016,lundberg2017}. In these user studies, people are normally shown two or more explanations for the same prediction, generated by different explanation methods, and then prompted to select the one, which is better, i.e. easier to understand and interpret. This process is not only time-consuming but can also be expensive and tedious to be carried out. In our research, we thus strive to take a different avenue, using an approach that can be automated to compare different explanations amongst each other. With this framework, we strive to overcome the shortcomings of the questionnaire method. An example of an approach that could be suitable to compare different explanations is to define a set of appropriate axioms.
\\For the axioms to be considered appropriate, these must be (i) Findable (i.e. can we find axioms to compare different EMs), (ii) Applicable (i.e. the axioms are computationally and otherwise feasible to be used in practice), and (iii) Common practice (i.e. it is non-extraordinary to use axioms in the field of interpretable machine learning). Moreover, the validation of the results of such an axiomatic framework brings us to our second research question, which we define as follows.

\noindent\textbf{Research Question 2:} \textit{How can a framework that enables us to compare different explanations amongst each other be evaluated in terms of yielding meaningful results?}
\\This second question deals with the evaluation of the framework that we develop in this thesis. As a good indicator whether our framework yields meaningful results, we can rely on related literature, which uses questionnaires with people to find the most interpretable EM. If the results from our framework are consistent, i.e. reach the same conclusions as related research, then we consider the results by our framework to be meaningful.
\\In sum, we strive to develop a framework that overcomes the shortcomings of manual explanation quality testing, while ensuring the validity of our framework, by using related research.

With a clear understanding of our research questions, its components and implications, we advance with the formulation of an appropriate and feasible research methodology in the next subsection.

%% ===========================
\subsection{Research Methodology}
\label{ch:Introduction1:sec:Section2:Subsection2}
%% ===========================
In this subsection, we introduce the research methodology, which we used in this thesis project. The method is an iterative cycle of four sequential steps (figure \ref{figure 1.3}).
\\The first phase, as well as all other phases, consists of two consecutive steps. In the first
%%%%%%%%%%%%% Figure %%%%%%%%%%%%%
\begin{figure}[htp]
  \begin{center}
    \includegraphics[width=1.00\textwidth]{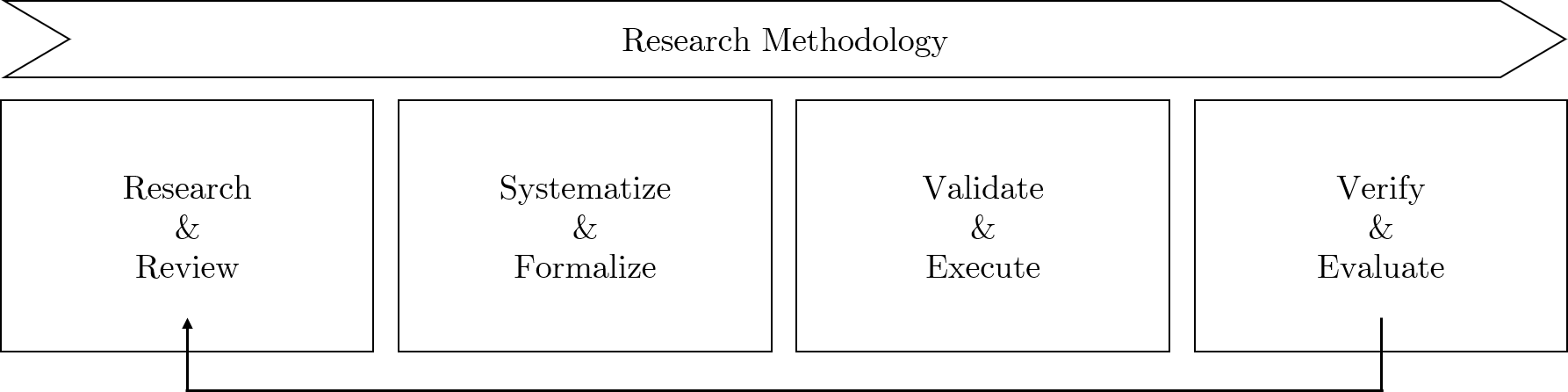}
  \end{center}
  \vspace{-20pt}
  \caption{Research Methodology: an Iterative and Sequential Process}
  \label{figure 1.3}
  \vspace{-10pt}
\end{figure}
%%%%%%%%%%%%% Figure %%%%%%%%%%%%%
step, we conducted research to identify relevant literature by consulting scientific journals, dissertations, books and papers, mainly in the domain of \textit{interpretable machine learning}. Moreover, several conversations with industry experts were conducted to narrow down the research problem of the present work. After that, all collected materials were reviewed and relevant parts selected for further processing.
\\In the second phase, we systematised the reviewed and highlighted literature, by allocating it to different categories. These categories are further discussed in the second chapter. This categorisation helped us to identify: links between different literature, opposing ideas, as well as research gaps. By using our literature categorisation, we proceeded with formalising our research problem and with the derivation of a feasible approach to tackle it.
\\A few validation sanity checks started the third phase. First, we analysed whether the identified problem is of relevance in practice and research, i.e. if there is a need for a solution to the identified issues. Second, we validated whether our approach is feasible, i.e. realisable in limited time and with limited resources. After that, followed the execution, in which we used the Cross-Industry Standard Process for Data Mining (CRISP-DM)\nomenclature{CRISP-DM}{Cross-Industry Standard Process for Data Mining}, as a reference and guidance model \citep{shearer2000}. We planned the longest time for this phase of the thesis.
\\In the fourth phase we started by verifying if the approach was correctly implemented in the execution and if the built solution tackled our problem. With this verification, we proceeded to the evaluation, where we interpreted the experiment results, derived conclusions, identified limitations and future topics of interest.
\\The research method features an iterative loop, which allowed us to return to previous phases, to fine-tune and adjust the focus and direction of the present work. 

With a clear understanding of the applied research methodology, we continue by outlining the structure of the following chapters.

%% ===========================
\section{Thesis Structure}
\label{ch:Introduction1:sec:Section3}
%% ===========================
The remainder of this thesis is mainly structured into four chapters. In chapter 2, we introduce and define all necessary concepts and terms, which serve as knowledge basis to understand the developed framework. Moreover, we present and review related work in the domain of \textit{interpretable machine learning}. The main contribution of the present work, the Axiomatic Explanation Consistency Framework, is introduced and defined in chapter 3. To verify and apply the developed framework, in chapter 4 we conduct a series of experiments. Finally, in chapter 5, we discuss the results of the experiments and of the thesis overall, and conclude by underlining the contribution, limitations and future directions for research.
%% ===========================
%% ===========================
\chapter{Preliminaries}
\label{ch:Content1}
%% ===========================
%% ===========================

\hspace{\parindent}In this preliminaries chapter, we introduce fundamental concepts upon which the developed Explanation Consistency Framework (ECF)\nomenclature{ECF}{Explanation Consistency Framework} is based. This chapter aims to provide the foundation to make the remainder of this thesis comprehensible. We mainly cover four areas. First, we provide the necessary knowledge and concepts in the domains of machine learning and statistics, as these represent the starting point of our research problem and question, as defined in the introduction. Second, we look at the role of interpretability in the context of machine learning. The goal of the second section is thus to highlight why interpretability is crucial in the context of ML. In the third section, we discuss how explanation methods improve the interpretability of black box ML models. This sub-chapter also motivates and further investigates the need for our developed Explanation Consistency Framework. The aim of the third section is thus, twofold – introduce explanation methods and their corresponding explanations as tools to establish interpretability, and highlight the need for the developed ECF. Finally, in the fourth section of the preliminaries, we investigate related work. The goal herein is to describe other paths that researchers have taken to compare the quality of different explanation methods and their given explanations.

%% ===========================
\section{Machine Learning and Statistical Foundations}
\label{ch:Content1:sec:Section1}
%% ===========================
In this first section of the preliminaries, we give an overview of fundamental topics in machine learning and statistics, which are necessary for the understanding of the remainder of this research. We start with the topic of machine learning, defining what it is and of which main branches it is composed. After that, we introduce and review some fundamental notions related to the field of statistics, upon which we base parts of the developed Explanation Consistency Framework.

As defined in the introduction, machine learning is a set of techniques that enables computer systems to "learn" and thus to progressively improve their performance on a specific task, thereby eliminating the need for detailed and explicit programming \citep{samuel1959}. Simply put, machine learning is a technique to teach computers to make and improve predictions based on data \citep{molnar2018}. There are two major types of ML – supervised and unsupervised, which we discuss in the upcoming sections. Moreover, there are also mixed-forms, such as semi-supervised learning and other categories such as reinforcement learning, which are, however, beyond the scope of this thesis.
\\In the following, we briefly highlight the main terminologies and concepts that are crucial for the machine learning process. The basis of a ML model is always an algorithm, which is a set of rules that a machine follows to achieve a specified goal\footnote{\url {https://www.merriam-webster.com/dictionary/algorithm}}. The ML model itself is thereby the output of a ML algorithm, and usually consists of the architecture and learned weights. To build a ML model, we thus need an algorithm, and we need data. There are mainly three types of data: structured (highly organized and usually found in relational databases); semi-structured (less organized, usually not found in a relational database, but still has some organizational properties, such as XML and JSON documents); and unstructured (this represents most of the existing data, such as images, text and video). In this research, we are working with structured datasets, i.e. data that is in a table format. The columns of these datasets are called features (or attributes) and the rows are called objects (or instances / data points). Usually the last column of our dataset, or more generally, the one that we wish to predict with our model is not called a feature, but instead label or target variable. Once we have prepared our dataset, we can specify the machine learning task that we wish to do, which can be of a supervised nature, such as classification and regression, or unsupervised, such as clustering and outlier detection. With the task definition, we can proceed to choose a suitable ML algorithm and then train the ML model, leaving out some data for model validation and testing in the supervised case. We fine-tune our model and its parameters with the validation data until we achieve satisfactory performance. We then use our best-performing model on the test set, to obtain an unbiased performance estimate. After this procedure, we can finally deploy the model and apply it to new data to obtain predictions.

Statistics and machine learning have the common goal of making inferences from data; however, both fields have traditionally different approaches \citep{ruping2006}. Statistics is a discipline with a long history that originated in a time long before sophisticated machines existed. Labour-intensive calculations had to be done by hand, and therefore, statisticians developed very sharp tools to extract the most information from small datasets \citep{ruping2006}. On the other end of the spectrum is machine learning, which originated from computer science, and deals with large amounts of data. A crucial idea herein is to automate as much human work as possible with computationally efficient algorithms. Moreover, as opposed to statistics, in machine learning often ad-hoc solutions without a general supporting theory are chosen \citep{ruping2006}.

After discussing some general notions of machine learning and how it differs from the traditional statistics approach, we continue to deepen our knowledge of ML in the following subsection. We mainly focus on ML techniques that are relevant to the context of this thesis. Thereafter, we elaborate on statistical measures that we adopt in the developed Explanation Consistency Framework.

%% ===========================
\subsection{Supervised Machine Learning}
\label{ch:Content1:sec:Section1:Subsection1}
%% ===========================
In supervised ML, the task is to train a model that accurately maps features to the target variable \citep{molnar2018}. The model should thus extract and learn structures that describe the dependency that objects have on labels \citep{ruping2006}. To perform this task, each of these instances requires a corresponding label. There are fundamentally two tasks that can be realised in supervised ML – description and prediction \citep{ruping2006}. In the former, we fundamentally aim to summarise the given data, and in the latter, we desire to predict the label of new observations as perfectly as possible. Moreover, the task can be of classification (when the label is a categorical variable), or regression (when the target variable is numerical).
\\The algorithm itself is guided by a so-called loss function, which it tries to minimise \citep{molnar2018}. If an algorithm, e.g. learns to predict house prices, the loss function could be to minimise the differences between predicted and actual house prices \citep{molnar2018}. Furthermore, models usually feature a regularisation parameter, which helps to control the complexity of the model \citep{chen2016xgboost}. These two parts – loss function and regularisation, together compose the objective function. The objective function specifies how the ML algorithm proceeds in finding the best parameter or weight combinations that yield the highest model performance given the training data \citep{chen2016xgboost}. The objective function is thereby defined as: $obj(\theta) = L(\theta) + \Omega(\theta)$, whereas $\theta$ represents the undetermined parameters that we need to learn from the data \citep{chen2016xgboost}. The objective function is thus the sum of the training loss function $L(\theta)$ and the regularisation term $\Omega(\theta)$.
\\An example of an often-used loss function is the already-mentioned mean square error, given as $L(\theta) = \frac{1}{n} \sum_{i} (y_i - \hat{y_i})^2$ \citep{friedman2009}. As mentioned above, regularisation helps us to control the complexity of the trained model and thereby to avoid overfitting. The goal of regularisation is to get the balance right between bias and variance \citep{chen2016xgboost}. A biased model is one that has limited flexibility to learn the true signal from training data \citep{friedman2009}, i.e. it usually leads to consistent but often inaccurate predictions on average (underfitting). A model with high variance is one that has a high sensitivity to individual objects in the training data \citep{friedman2009}, i.e. it often leads to inconsistent but usually accurate predictions on average (overfitting). High variance is thereby the result of over-complex models that not only model the general structures and dependencies in the training data but also noisy behaviour and outliers. In sum, regularisation helps us to control this trade-off between bias and variance \citep{chen2016xgboost}.

In the upcoming sections, we delve deeper into the field of supervised ML, after discussing some general ideas and concepts on it above. We highlight three different types of supervised ML algorithms, namely Extreme Gradient Boosting (XGB)\nomenclature{XGB}{Extreme Gradient Boosting}, Multilayer Perceptron (MLP)\nomenclature{MLP}{Multilayer Perceptron} and Long Short-Term Memory Networks (LSTM)\nomenclature{LSTM}{Long Short-Term Memory Network}. We mainly choose to review these three algorithms, as we resort to them in our experiments chapter.

%% ===========================
\subsubsection{Extreme Gradient Boosting (XGB)}
\label{ch:Content1:sec:Section1:Subsection1:Subsubsection1}
%% ===========================
In our daily lives, we all use certain criteria and heuristics to make decisions in all types of situations. Some of these decision-making processes, even if subconsciously, take on the structure of decision trees. As an example, when we decide whether to take an umbrella and / or a coat in the morning when leaving the house. This decision process fundamentally depends on the temperature and the weather condition. If it is raining and cold, we likely take both – an umbrella and a coat to leave the house. If it is raining and warm, we may only take an umbrella but no coat. If it is not raining and warm, we likely do not take either of the two items, and lastly, if it is not raining and cold, we may only take a coat but no umbrella. It are these types of questions, criteria and answers that are the building blocks of decision trees. Gradient boosted trees are based on decision trees, but instead of only using one tree, they use an ensemble of trees.
\\An ensemble more generally means that we do not only have one ML model, but we use several models of the same or different types simultaneously. We could, e.g. have an ensemble that is composed of a decision tree, a multilayer perceptron and a support vector machine \citep{maclin1997empirical}. The goal of ensembles is to achieve a higher predictive performance as compared to when we only use a single model by itself. The main principle in ensembles is that several weak learners together that perform better than random guessing converge to a stronger learner \citep{freund1996experiments}.
\\An ensemble of trees combines multiple decision trees into one model. This is also what XGB fundamentally is – a group of decision trees combined in a specific way, to act as a single model \citep{chen2016xgboost}.

There are three common approaches to build ensembles: bagging, boosting and stacking. We focus on the first two approaches, disregarding the third one for the remainder of this work. Moreover, we discuss these based on the example of decision trees.
\\In bagging or \textit{bootstrap aggregating}, the main idea is to train multiple independent models in parallel on subsets of the data to produce different classifiers \citep{breiman1996bagging}. This is shown in figure \ref{figure 2.1} on the left. To e.g. train an ensemble of $N = 3$ trees, we need to first generate 3 new samples of the original training data through random sampling with replacement \citep{breiman1996bagging}. By using a sampling technique with replacement, some observations can still be repeated over these $N$ new training sets. In the next step, we use
%%%%%%%%%%%%% Figure %%%%%%%%%%%%%
\begin{wrapfigure}{r}{0.60\textwidth}
  \vspace{-20pt}
  \begin{center}
    \includegraphics[width=0.60\textwidth]{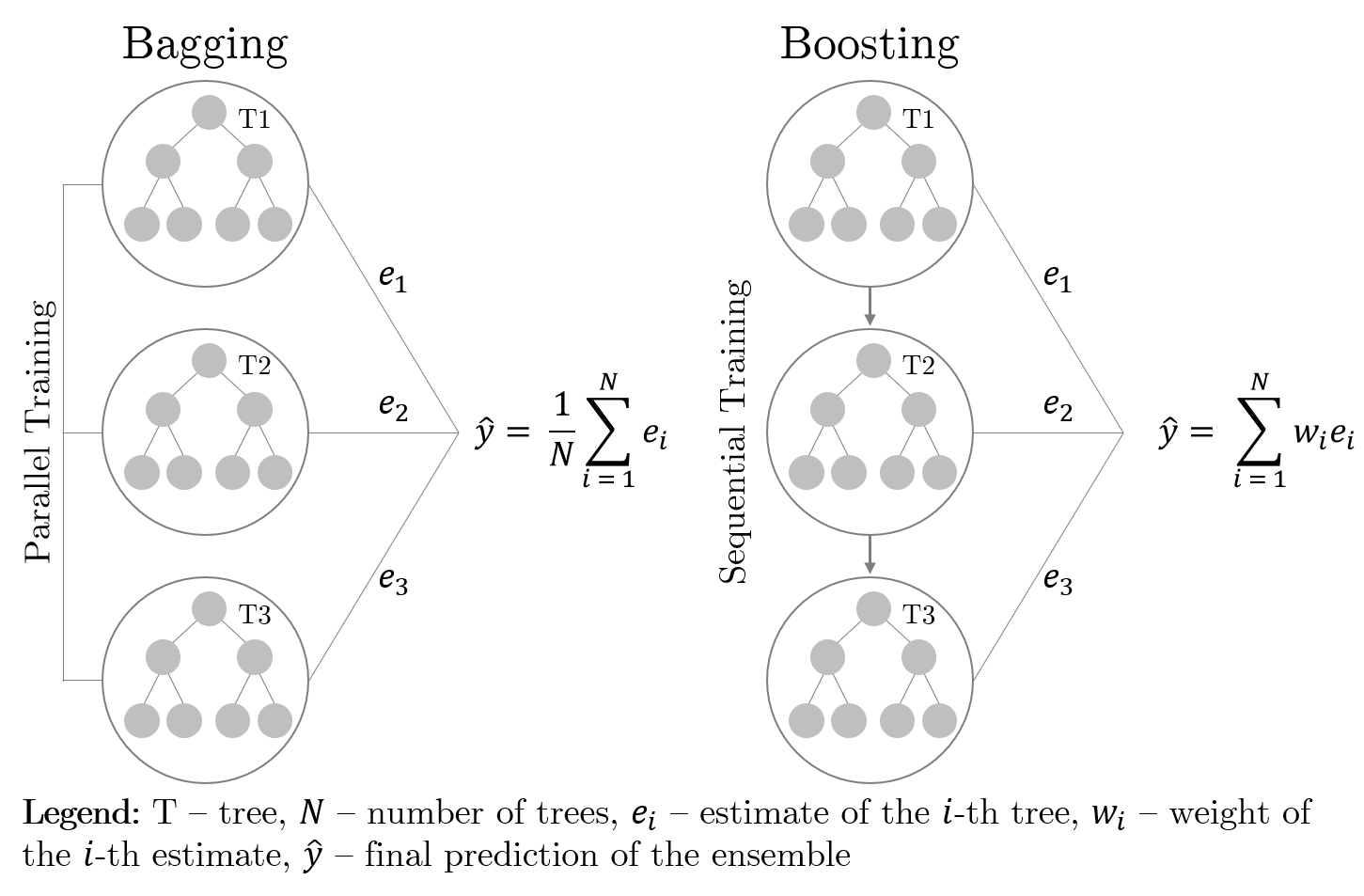}
  \end{center}
  \vspace{-20pt}
  \caption{Tree Ensembles: Bagging vs Boosting}
  \label{figure 2.1}
  \vspace{-10pt}
\end{wrapfigure}
%%%%%%%%%%%%% Figure %%%%%%%%%%%%%
these to train our 3 trees, whereas each tree is assigned its own training set. With this procedure, by training on different data, we produce 3 distinct trees. Once all the trees are trained, we can use them to classify new data, by applying each tree to the new observations. As all trees are somewhat different, they estimate slightly different outcomes. To combine these 3 estimates in bagging, we would apply a majority vote \citep{breiman1996bagging}. This translates into averaging the estimations of the 3 trees into a single prediction, by attributing an equal weight (importance) to each tree. This approach produces models that are especially useful in situations where overfitting (i.e. high variance) is a problem. A well-known representative of the bagging method is the random forest.
\\In boosting, the main idea is still to train multiple models; however, we train new models in sequence, i.e. one after the other \citep{freund1996experiments}. The rationale thereby is to train new trees that perform especially well where previous ones failed to achieve a high predictive performance \citep{freund1996experiments}. In boosting, we still generate $N$ new samples of the original training data by random sampling; however, different observations now receive different weights in subsequent boosting rounds. If e.g. the first two trees have a high predictive success in the first half of the data, then the observations of the second half are attributed with higher weights for the next tree. In this way, the data that have been misclassified by the previous trees have a higher probability to be selected in the random sampling process for the third tree. This is the main difference to bagging, where each observation has the same probability to be selected for a new dataset \citep{freund1996experiments}. This procedure thereby also produces trees that are different from each other. Once our $N$ trees are trained, we can use them to classify new observations. In opposition to bagging, in boosting, we cannot simply take the average of all estimates of the trees. Instead, the sum of weighted estimates must be computed, as can be seen in figure \ref{figure 2.1} on the right. This is a consequence of the boosting stage, in which the algorithm allocates different weights to different trees, based on their predictive performance on the data \citep{freund1996experiments}. A tree with good predictive power on the training data gains a high weight, and conversely, a tree with a bad predictive success receives a small weight or even a weight of zero. This approach is especially useful in situations where underfitting (i.e. high bias) is a problem. The XGB algorithm that we later use in the experiments belongs to this family of boosting methods.

After discussing the differences between bagging and boosting, and describing why the XGB algorithm belongs to the latter approach, we briefly highlight some further characteristics. A primary trait that distinguishes XGB from other boosting algorithms, such as the traditional AdaBoost, is its capability to use any differentiable loss function \citep{chen2016xgboost}. This is achieved by using the gradient descent (GD)\nomenclature{GD}{Gradient Descent} algorithm to minimise the loss function when training subsequent trees, hence the "gradient" term in its name \citep{chen2016xgboost}. Moreover, according to \cite{chen2016xgboost} the XGB algorithm easily scales to huge amounts of data, while using a minimal amount of resources. This is also the reason for the "extreme" term in its name – it pushes the computation limits of machines. Lastly, it often yields highly accurate results in practice, a reason for it to repeatedly be on top of competition-winning algorithms in data science challenges \citep{chen2016xgboost}. For more specific information on ensemble learners and on extreme gradient boosting, we recommend the book by \cite{friedman2009} and the original paper by \cite{chen2016xgboost}.

Having a basic understanding of the XGB model and of the principles by which it works, we proceed to discuss another famous ML algorithm in the next subsection – the multilayer perceptron.

%% ===========================
\subsubsection{Multilayer Perceptron (MLP)}
\label{ch:Content1:sec:Section1:Subsection1:Subsubsection2}
%% ===========================
Multilayer perceptrons are special types of artificial neural networks, which in turn are loosely modelled on the structure of the human brain \citep{schalkoff1997artificial}. MLPs are fully connected and operate in a feedforward fashion \citep{friedman2009}. Fully connected means that each \textit{neuron} or node of one layer in the network is linked to every node in the succeeding layer of the network. Moreover, feedforward means that the information only flows forward in the network, from the input to the output nodes, i.e. there are no loops or backward connections between layers. In figure \ref{figure 2.2}, we can observe these two characteristics and the general structure of a MLP with 2 hidden layers.

The architecture of the MLP consists of three types of layers: input, hidden and output \citep{friedman2009}. The input data determines the dimensionality of the input layer, which is responsible for passing on the information to the hidden layer. As shown in figure \ref{figure 2.2} below, the hidden layer can be composed of multiple levels and has an activation function in each node. If a network has multiple hidden layers, then the outputs of one layer become the inputs of the next layer and so forth. The purpose of the hidden layer is to learn different levels of abstractions of the data and to extract patterns, which are helpful to solve the problem at hand. As an example, a convolutional neural network (CNN)\nomenclature{CNN}{Convolutional Neural Network}, usually used for tasks such as object recognition in pictures, has such a hierarchical organisation in its hidden layers \citep{altenberger2018}. The first few hidden layers of a CNN thereby recognise low-level features such as lines, edges and shadows. Deeper layers recognise higher-level patterns, in the case of a facial recognition CNN, these could be eyes, noses and mouths \citep{altenberger2018}. Finally, the output layer also uses an activation
%%%%%%%%%%%%% Figure %%%%%%%%%%%%%
\begin{wrapfigure}{l}{0.50\textwidth}
  \vspace{-20pt}
  \begin{center}
    \includegraphics[width=0.50\textwidth]{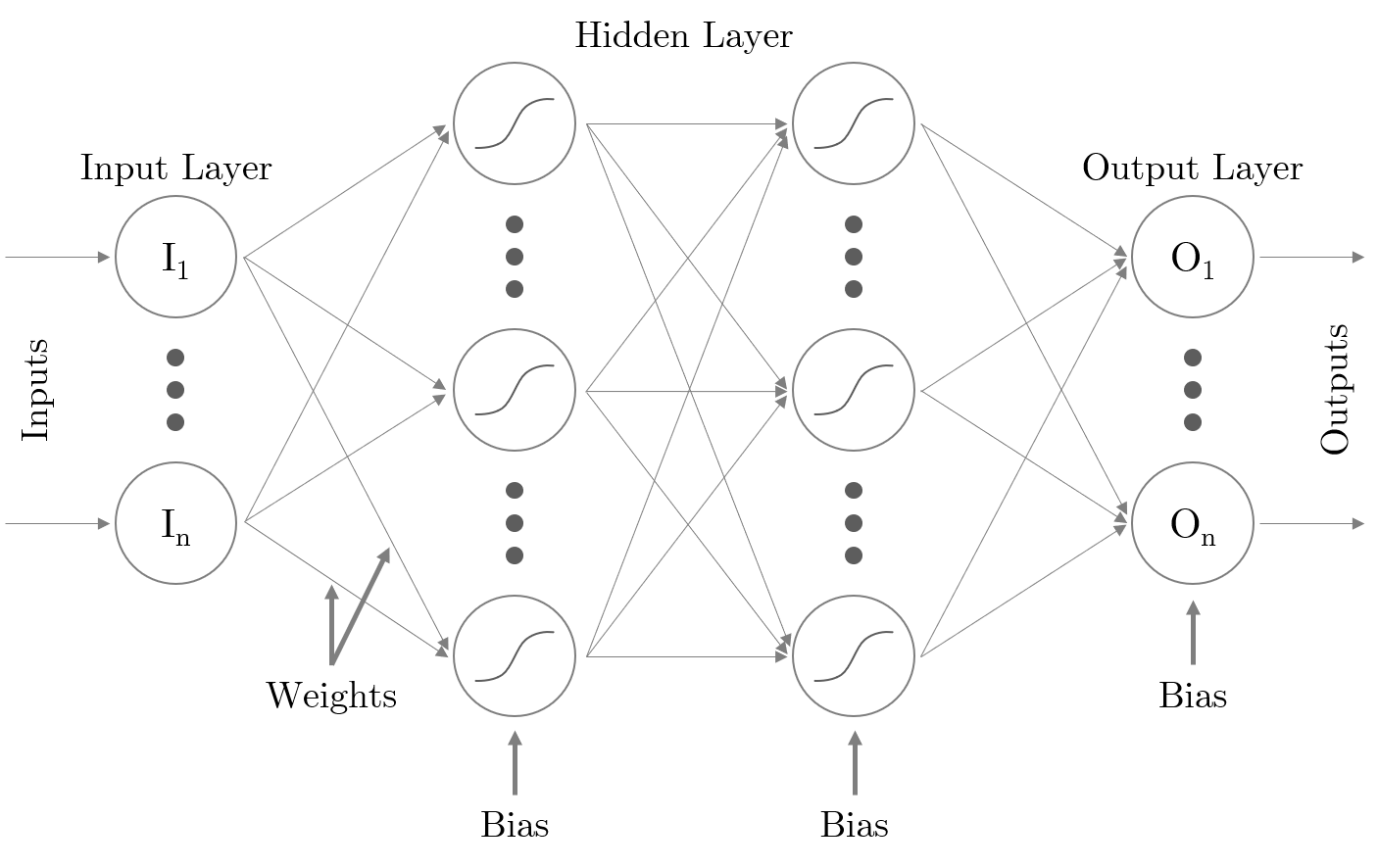}
  \end{center}
  \vspace{-20pt}
  \caption{Multilayer Perceptron: Architecture and Components}
  \label{figure 2.2}
  \vspace{-10pt}
\end{wrapfigure}
%%%%%%%%%%%%% Figure %%%%%%%%%%%%%
function to compute and return the final prediction of the network in the desired format (regression or classification values).
\\The directed arrows between different nodes and layers represent the connections and associated weights that the network learns in the training process. In a standard implementation, the initial weights are randomly chosen by the algorithm and continuously adjusted in each iteration to achieve a better predictive performance on the training data. This process of iteratively updating the weights after evaluating the output error is called backpropagation \citep{friedman2009}.
\\The sigmoidal "S-shaped" curves in the two hidden layers of the network in figure \ref{figure 2.2} represent the previously-mentioned activation functions. These are responsible for computing the output of a certain node, given all inputs that go into that node. Activation functions are chosen problem-specifically, whereas different layers in the network can have different types of activation functions. As an example, if we want to predict a categorical variable, we could use hyperbolic tangent (tanH) activation functions in the first hidden layer, sigmoid activation functions in the second hidden layer, and finally, softmax activation functions in the output layer. In the MLP, non-linear activation functions enable to learn complex and non-linear relations without using many nodes \citep{schalkoff1997artificial}. Moreover, non-linear activation functions serve to control the frequency and intensity with which individual nodes (neurons) fire, i.e. pass on information or signals to subsequent nodes \citep{friedman2009}.
\\The last element of the MLP architecture to discuss are the biases. These are special types of neurons that only have outgoing connections, and they connect to all neurons in a particular layer. So every layer (apart from the input layer) has its bias node. The bias term enables activation functions to shift left or right, to fit the data better \citep{friedman2009}. The weights of the biases are usually initialised as 0 and also learned and updated in the backpropagation step.

With a good understanding of the most important parts of the MLP architecture, we continue to explore how such networks learn. The process fundamentally consists of two phases: the forward pass and the backward pass \citep{friedman2009}.
\\The network always starts with the forward pass, in which inputs are passed forward through the network. In this phase, several transformations are applied to the inputs, by passing them through the weighted edges and activation functions of all neurons. Once the transformed information reaches the output layer, the last activation function is used to compute the prediction of the network for that input object. If the predicted value is equal to the actual value, then the next forward pass with the subsequent instance is initiated. If this is, however, not the case, then the backward pass is started.
\\The backward pass has two sub-steps: error calculation and error minimisation. In the former, the predicted output is compared to the actual output, and the significance of the deviation (i.e. size of the error) is calculated. This can be achieved by, e.g. computing the MSE. Once the network determined how badly the prediction is off, it proceeds with the error minimisation. In the backpropagation or minimisation step, all weights and biases of the network are updated in such way that the total error (also called cost) is minimised, i.e. the predicted output gets closer to the actual output \citep{friedman2009}. This reduces the error of each neuron, and consequently, of the entire network. To achieve this, usually, a variation of the already mentioned gradient descent (GD) optimization procedure, also known as \textit{delta rule} is applied \citep{friedman2009}. In the GD algorithm, the main principle is to minimise the cost function, by iteratively taking steps in the opposite (negative) direction of the gradient of the function \citep{friedman2009}.
\\In sum, the process of training a MLP consists in a cycle of feeding information forward through the network to make predictions, and eventually, propagating necessary adjustments backwards through it. The network improves its performance and thereby "learns", by continuously iterating over a training dataset, while adjusting the discussed parameters of the architecture (weights and biases). For more specific information on the MLP, gradient descent and neural networks in general, we recommend the books by \cite{friedman2009}, \cite{Goodfellow-et-al-2016} and \cite{schalkoff1997artificial}.

After discussing the architectural elements and processes that are at the basis of feedforward neural networks, to which the MLP also belongs, we proceed with an excursus in the upcoming section. We briefly introduce a particular type of recurrent neural network, namely the long short-term memory network.

%% ===========================
\subsubsection{Excursus: Long Short-Term Memory Network (LSTM)}
\label{ch:Content1:sec:Section1:Subsection1:Subsubsection3}
%% ===========================
Long short-term memory networks were first proposed by \cite{hochreiter1997long} and belong to the family of recurrent neural networks (RNN)\nomenclature{RNN}{Recurrent Neural Network}. These are an extension of plain feedforward ANNs. In RNNs, information can flow in all directions, as neurons are not only connected to other neurons in subsequent layers, but also to themselves, to neurons in the same layer and even to neurons in previous layers \citep{altenberger2018}. This enables information to persist in the network \citep{colah2015}. By means of this memory function, RNNs can learn sequential data (with a temporal dimension), such as time-series, musical notes and data ordering problems, such as character sequences in words \citep{medsker2001recurrent}.

A variety of practical applications have however shown that RNNs do not deliver satisfactory results when long-term dependencies are involved \citep{colah2015}. Even though we can tweak RNNs in such a manner that they achieve a good performance for a specific type of long-term dependency, they then lose their generalisation power for other long-term dependencies. This is where the LSTM comes into play. LSTMs were primarily created to learn long-term dependencies \citep{colah2015}. This is one of the main reasons for their current success in many fields such as speech recognition, machine translation, image captioning and voice transcription.
\\Discussing the exact architecture of LSTMs is beyond the scope of this research, for that purpose we recommend the original paper by \cite{hochreiter1997long}, the book by \cite{Goodfellow-et-al-2016}, or the research of famous AI researcher Christopher Olah \citep{colah2015}. We do, however, use LSTMs in our experiments chapter, to assess their predictive performance on one of our datasets as compared to the MLP and XGB. The main advantage of LSTMs is that due to their memory-function, they require no feature engineering (FE)\nomenclature{FE}{Feature Engineering}. This is often a lengthy but necessary process, to achieve a high predictive performance with traditional ML models. As a small side-experiment, we, therefore compare the predictive performance of an LSTM on the same dataset without FE, with the XGB and MLP with FE.

After this brief excursus on RNNs and LSTMs, to which we return in the experiments chapter, we continue to discuss unsupervised ML techniques in the upcoming section.

%% ===========================
\subsection{Unsupervised Machine Learning}
\label{ch:Content1:sec:Section1:Subsection2}
%% ===========================
In unsupervised machine learning, the starting point is quite different to supervised ML, as we have a set of observations without correct answers (labels) \citep{ruping2006}. The goal in unsupervised ML is to infer properties or to extract structures from the data, rather than predicting values \citep{ruping2006, friedman2009}. Unsupervised ML algorithms thereby discover relationships and dependencies without needing a supervisor to give them feedback on the degree of error for each prediction \citep{friedman2009}. As opposed to supervised ML, there is no clear measure of success in unsupervised ML. This means that the quality of results is often judged based on heuristics and consequently includes some level of subjectivity \citep{friedman2009}. 

Many problems can be tackled with unsupervised ML algorithms, such as dimensionality reduction, clustering and association \citep{friedman2009}. The former method is concerned with reducing the size, i.e. compressing data, while still maintaining its essential information and structure. Clustering is used to analyse the similarity of different objects and to partition these into groups. Lastly, associations are used to discover rules that apply to a large number of objects in the dataset. An application of association rules is market basket analysis that discovers relations of the form: "if person A buys item X, the chance that item Y is also purchased is 90\%". For this research, we focus on clustering methods, as we make use of these in the experiments chapter.

There are different approaches to clustering, such as partitioning, hierarchical and density-based methods. In partitioning the idea is to subdivide the feature space with n observations into k clusters, whereas each observation is allocated to the cluster with the closest mean \citep{macqueen1967}. Two famous advocates of this clustering approach are the k-means and k-medoids algorithms \citep{macqueen1967,kaufman1990}. Hierarchical clustering produces a tree-based structure that recursively joins (splits) clusters at the next lower / upper level, based on the merge (split) that results in the smallest / largest intergroup dissimilarity \citep{friedman2009}. Two well-known strategies for this clustering type are agglomerative nesting (AGNES)\nomenclature{AGNES}{Agglomerative Nesting} and divisive analysis (DIANA)\nomenclature{DIANA}{Divisive Analysis}. Finally, in a density-based approach, observations are clustered, if they are tightly packed together, i.e. if points have many close neighbours \citep{ester1996density}. Conversely, observations are marked as outliers, if these lie by themselves in a low-density region. An example of a density-based clustering method is the density-based spatial clustering of applications with noise (DBSCAN)\nomenclature{DBSCAN}{Density-Based Spatial Clustering of Applications with Noise} \citep{ester1996density}.

With a general notion on the differences between supervised and unsupervised ML, the main problems that it addresses, and the essential clustering algorithms, we proceed to the next subsection. We further discuss hierarchical clustering and partitioning methods, namely AGNES and k-means, as we rely on these two in the experiments chapter.

%% ===========================
\subsubsection{Agglomerative Hierarchical Clustering (AGNES)}
\label{ch:Content1:sec:Section1:Subsection2:Subsubsection1}
%% ===========================
As discussed above, AGNES is a hierarchical clustering method that follows a bottom-up (agglomerative) strategy to merge clusters \citep{friedman2009}. The algorithm starts with each object in its own cluster, also called a \textit{singleton} \citep{friedman2009}. At each iteration, the two closest or least dissimilar clusters are merged, thereby forming a new cluster composed of the union of both merged clusters. This means that in each iteration, there is one less cluster than before. The algorithm continues until only one
%%%%%%%%%%%%% Figure %%%%%%%%%%%%%
\begin{wrapfigure}{r}{0.40\textwidth}
  \vspace{-15pt}
  \begin{center}
    \includegraphics[width=0.40\textwidth]{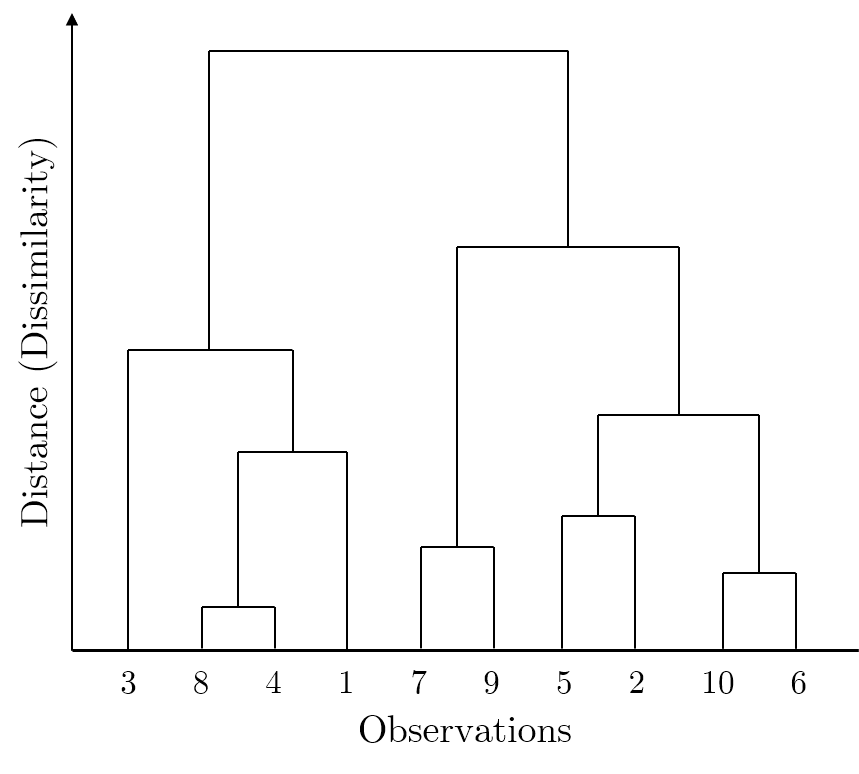}
  \end{center}
  \vspace{-20pt}
  \caption{Dendrogram: Hierarchical AGNES Clustering Example}
  \label{figure 2.3}
  \vspace{-10pt}
\end{wrapfigure}
%%%%%%%%%%%%% Figure %%%%%%%%%%%%%
cluster remains, containing all $n$ observations. Therefore, if not stopped before, the algorithm runs for $n-1$ iterations \citep{friedman2009}.

The AGNES procedure is monotone, which means that the dissimilarity between merged clusters increases with each merger \citep{friedman2009}. We can visualise this property on a so-called dendrogram (figure \ref{figure 2.3}), where we locate all data points on the x-axis and the increasing dissimilarity (distance) on the y-axis. On the displayed diagram we have an example with 10 observations, and the first merger occurs between singletons 4 and 8. These two observations are thus the two most similar ones amongst the 10. The next few iterations merge singletons 6 and 10, then 7 and 9, and then 2 and 5. Once these are merged, in the succeeding iteration a new cluster is formed by merging the already existing cluster of 4 and 8, with the singleton of observation 1. The procedure continues until the last merger in iteration 9, where the cluster containing observations 1, 3, 4, and 8, is merged with the cluster containing observations 2, 5, 6, 7, 9 and 10. In the dendrogram, we can see how the distance increases with each merger, which perfectly portrays the monotonicity property. The dendrogram is also useful in determining how many clusters represent a good choice for the problem at hand. If we look again at figure \ref{figure 2.3}, we, e.g. quickly see that the last merger dramatically increases the distance. It may therefore not be a good solution to have all 10 observations in a single cluster, but better to have two or even more clusters. Lastly, the dendrogram provides an interpretable and complete description of the clustering result in a visual format, which is the main reason for the popularity of hierarchical clustering methods \citep{friedman2009}.

The last aspect that we briefly discuss with regards to AGNES is the dissimilarity function. As previously mentioned, the procedure performs mergers based on the smallest intergroup dissimilarity between clusters \citep{friedman2009}. There are multiple different dissimilarity functions, of which we discuss single linkage and Ward linkage, as we rely on these later on.
\\In single linkage dissimilarity, we define the distance between two clusters as the minimum distance between any single two points, of which one belongs to the first cluster, and the other belongs to the second cluster \citep{friedman2009}. Based on this distance measure, at each merge clusters are combined based on the smallest single linkage distance \citep{friedman2009}. According to \cite{jardine1971}, a hierarchical clustering with single linkage is equivalent to the computation of a minimum spanning tree, which guarantees to find one amongst a set of optimal solutions, as long as we disregard ties. This is a valuable property to which we return in the experiments chapter.
\\The Ward linkage method is not directly based on a distance measure that is defined between two points, but instead on variance \citep{ward1963}. In Ward's method, we merge those two clusters that provide the lowest increase in the combined sum of squares, i.e. the ones that minimise the variance \citep{ward1963}. In practice, using different linkage methods often results in quite distinct results, which is why it is important to evaluate more than only one linkage type or dissimilarity function.

Having a good understanding of the nuts and bolts of the AGNES clustering method, including dendrograms and linkage functions, we proceed to the next unsupervised ML method – k-means, in the upcoming subsection.

%% ===========================
\subsubsection{K-Means Clustering}
\label{ch:Content1:sec:Section1:Subsection2:Subsubsection2}
%% ===========================
The k-means algorithm follows a partitioning approach, in which it tries to find cluster centres in a set of unlabelled data with $n$ observations \citep{friedman2009}. To start the procedure, the user has to define $k$, which is the number of groups that the algorithm is ought to find in $n$. Thereafter, each observation is iteratively assigned to its closest of the $k$ centres and then the centroid is computed \citep{friedman2009}. The centroid is thereby the mean position of all points in a cluster, taking into account every feature.

In figure \ref{figure 2.4}, we show how the procedure works and what steps it follows in a more interpretable visual form. We start at the top left of the diagram with (1), where all observations ($n=44$) are shown in the feature space. As mentioned, we must first define $k$, which we choose to be 3, as the data visually suggests a good fit for 3 clusters. Hence in (2), we randomly place the 3 initial centres on the feature space. In (3), we calculate the distance of each observation to every initial centre, thereby assigning each one to the centre to which it
%%%%%%%%%%%%% Figure %%%%%%%%%%%%%
\begin{wrapfigure}{l}{0.60\textwidth}
  \vspace{-20pt}
  \begin{center}
    \includegraphics[width=0.60\textwidth]{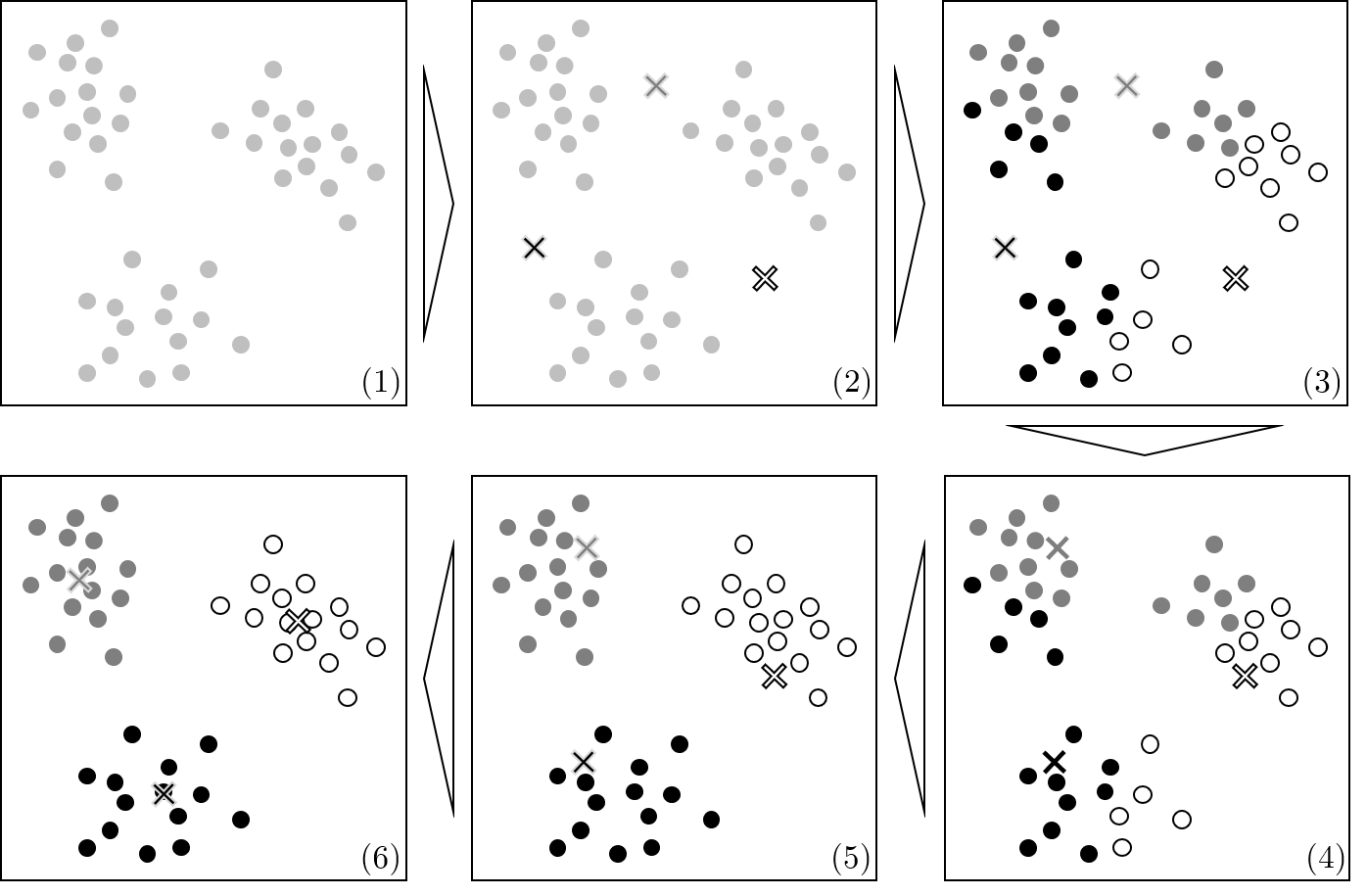}
  \end{center}
  \vspace{-20pt}
  \caption{Partitioning Example: K-Means Clustering}
  \label{figure 2.4}
  \vspace{-10pt}
\end{wrapfigure}
%%%%%%%%%%%%% Figure %%%%%%%%%%%%%
is the closest. We can see that now the initial centres do not represent the mean positions of all observations assigned to them anymore. In (4), we must, therefore, move our initial centres in such a manner that these correspond to the centroids again. Now, because we moved our centres from step (3) to (4), not all observations are assigned to their closest centre anymore, as can be seen in (4). In step (5), we must thus reassign all observations to their new closest centre. Not all observations are affected, i.e. need to be reassigned. Finally, in (6), due to the reassignments in (5), we must move our centres again, so that these correspond to the mean of all observations assigned to them. The algorithm would now go back to step (5), however, as each observation is already assigned to its nearest centre, no further changes take place, and the algorithm terminates.
\\In sum, k-means mainly alternates between the two following steps, as based on \cite{friedman2009}: First, for every centre identify the subset of $n$ that is closer to it than to any other centre; Second, compute the mean of each feature for all $n$ observations and select this mean vector to become the new centre for that cluster. Once no more reassignments take place, the algorithm terminates returning the final cluster centroids and the assignments of each observation to one of the $k$ clusters \citep{friedman2009}.

One of the issues with k-means is that the solution of the algorithm heavily depends on the randomly initiated cluster centres at the beginning of the algorithm \citep{bahmani2012scalable}. It could e.g. happen that the algorithm gets stuck in a similar position as in step (3) of the diagram above if the initial centres were chosen differently. In practice, we usually address this issue with three distinct approaches. The first approach involves running the algorithm several times with random initialisations and comparing the results, selecting the most representative one. Second, the cluster centres are still randomly initialised, however, with the constraint that these are distant from each other, which provably leads to better results than only random initialisation \citep{bahmani2012scalable}. Finally, the third approach is a somewhat informed one – if we already have some idea on where the final centroids could be, we can also initialise the k-means algorithm, providing it with (user-defined) initial cluster centres. In the later experiments of this research, we mainly rely on the second and third approaches.

For further details on the k-means algorithm and its mathematical formulation, we recommend the book by \cite{friedman2009} and the original paper by \cite{macqueen1967}. After discussing the main aspects of k-means clustering – how the algorithm works and how we can bypass some of its pitfalls, we proceed to the next subsection of statistical measures.

%% ===========================
\subsection{Statistical Measures}
\label{ch:Content1:sec:Section1:Subsection3}
%% ===========================
We have already discussed a few topics on machine learning, in which statistics is always a core component. Hereafter, we continue to explore two further statistical models, which we use in the formulation of the Explanation Consistency Framework. These are two measures mainly used in statistics, but also in computer science, namely the Jaccard similarity coefficient and Spearman's rank correlation coefficient. The former is used to compare the similarity of sets and binary vectors, and the latter is employed to measure the rank correlation between two variables, as opposed to Pearson's linear correlation.

The structure of this section thus only features two subsections, whereas we dedicate one to each of the two measures mentioned above. After that, the second part of the preliminaries chapter starts, in which we discuss and motivate the role that interpretability plays in machine learning.

%% ===========================
\subsubsection{Jaccard Similarity Coefficient}
\label{ch:Content1:sec:Section1:Subsection3:Subsubsection1}
%% ===========================
The Jaccard similarity coefficient, also known as Jaccard index, was first introduced by \cite{jaccard1901}. It was originally used to compare the floras of different alpine regions, and has since then, been adopted in many other applications, such as object detection in computer vision. It is defined as follows:
\[J(A,B) = \frac{\mid A \cap B \mid}{\mid A \cup B \mid} = \frac{\mid A \cap B \mid}{\mid A \mid + \mid B \mid - \mid A \cap B \mid}\]
where $A$ and $B$ are two non-empty sets and $J \in [0,1]$. A Jaccard similarity of 1 would thereby mean that the two sets are equal and a similarity of 0 would mean that the two sets have no element in common.
\\To make this measure more graspable, let us imagine the following example: We have two shopping baskets, the first one contains strawberries, ice-cream and water, and the second one holds salad, bread and water. Even though these two baskets are different in most positions, there is still one item that they have in common – water. The Jaccard index between these two baskets can thus be calculated as: $J(Basket_1,Basket_2) = \frac{1}{3+3-1} = \frac{1}{5}$. This means that the similarity between our two baskets is 20\%, as there are 5 unique items (strawberries, ice-cream, salad, bread and water), of which, however, only one (water) is contained in both baskets.

With a good understanding of the computation and utility of the Jaccard index, we proceed to the second statistical measure – Spearman's rho, in the next subsection.

%% ===========================
\subsubsection{Spearman's Rank Correlation Coefficient}
\label{ch:Content1:sec:Section1:Subsection3:Subsubsection2}
%% ===========================
The Spearman rank correlation coefficient, also called Spearman's rho, was first proposed by \cite{spearman1904}. It is a non-parametric measure of association that is based on the ranks (ordering) of values rather than on the values themselves \citep{gregory2009nonparametric}. As opposed to Pearson's linear correlation coefficient, it is applicable to non-linear relations between monotonic and at least ordinal variables \citep{gregory2009nonparametric}. Spearman's rho ($\rho$) is defined as follows (for cases without ties in values / ranks):
\[\rho = 1 - \frac{6 \sum D_i^2}{n(n^2-1)}\]
where $n$ is the number of ranked pairs and $D_i$ are the differences between ranked pairs \citep{gregory2009nonparametric}. We rely on this formula for the illustrative example below, as there are no ties in ranks. For situations with ties, the process remains identical, but a different formula applies that can e.g. be consulted in \cite{gregory2009nonparametric}.

Let us imagine that we have the results of 10 students for two written exams, namely mathematics (maths) and statistics (stats). The scores are represented in table \ref{Table 2.1}, in the
%%%%%%%%%%%%% Table %%%%%%%%%%%%%
\begin{wraptable}{r}{0.45\textwidth}
\vspace{-5pt}
\scalebox{0.77}{ %To resize the table "zoom-out"
\centering
\begin{tabular}{ C{1cm} C{1cm} C{1cm} C{1cm} C{1cm} C{1cm} }
\toprule
\multicolumn {2}{c}{Scores $S_j$} & \multicolumn {2}{c}{Ranks $R_k$} & \multicolumn {2}{c}{Differences $D_i$} \\
\cmidrule(lr){1-2} \cmidrule(lr){3-4} \cmidrule(lr){5-6}
$\mathrm{S}_\mathrm{Maths}$ & $\mathrm{S}_\mathrm{Stats}$ & $\mathrm{R}_\mathrm{Maths}$ & $\mathrm{R}_\mathrm{Stats}$ & $d_i$ & $d_i^2$ \\
\midrule
13	&	52	&	9	&	5	&	4	&	16	\\
22	&	72	&	3	&	2	&	1	&	1	\\
7	&	27	&	10	&	10	&	0	&	0	\\
20	&	43	&	4	&	8	&	-4	&	16	\\
17	&	50	&	6	&	6	&	0	&	0	\\
18	&	39	&	5	&	9	&	-4	&	16	\\
14	&	45	&	8	&	7	&	1	&	1	\\
24	&	87	&	1	&	1	&	0	&	0	\\
23	&	66	&	2	&	3	&	-1	&	1	\\
16	&	58	&	7	&	4	&	3	&	9	\\
\bottomrule
\end{tabular}}
\vspace{-5pt}
\caption{An Example of Spearman's Rank Correlation Coefficient}
\label{Table 2.1}
\vspace{-10pt}
\end{wraptable}
%%%%%%%%%%%%% Table %%%%%%%%%%%%%
first two columns. Each row represents the scores of one student in both exams. As the durations for the exams were different, these also use distinct grading scales. Now, what we want to find out from the data, is whether the fact that a student scores a high value in maths means that her success is also transferable to statistics and vice-versa. In other words, we want to analyse if there is a correlation between scoring high in maths and statistics. This is achieved by computing Spearman's rho. To calculate this value, we must first determine the ranks of each exam for every student. In maths, the highest score was 24, which corresponds to rank 1. Conversely in statistics, the highest score of 87 is also marked as rank 1. We continue to enumerate the remaining scores in both exams and obtain the third and fourth columns of all ranks in the table. In the next step, we compute $D_i^2$, by subtracting the ranks from each other and finally squaring the resulting differences. With all necessary components, we can now insert all values in the formula above and determine the correlation as follows: $\rho = 1 - \frac{6 \cdot 60}{10(10^2-1)} = 0,64$. This value indicates quite a strong positive relationship between the ranks of scores that students achieved in the maths and statistics exams. Consequently, we could expect that if a student scores a high (low) score in either of the two exams that her score in the other exam would likely also be high (low).

After discussing and understanding Spearman's rank correlation coefficient and several other fundamental concepts in machine learning and statistics, we proceed to the next part of the preliminaries chapter, where we highlight the role of interpretability in ML.

%% ===========================
\section{The Role of Interpretability in Machine Learning}
\label{ch:Content1:sec:Section2}
%% ===========================
We start this second part of the preliminaries by defining the term of interpretability, looking at its value, scope and dimensions. In the past section and introductory chapter, we have already used the term without properly defining it, or giving it meaning for the context of this work. The Merriam-Webster dictionary defines the term to interpret as "to present in understandable terms"\footnote{\url {https://www.merriam-webster.com/dictionary/interpretability}}. Therefore, interpretability is related to the capacity of how well humans understand something by looking and reasoning about it.
\\With this notion, we quickly understand that interpretability is a very subjective concept and therefore hard to formalise, which is one of the main reasons why only in recent years it has received significant attention and traction in the field of machine learning \citep{ruping2006}. How well humans understand something heavily depends on various factors. As an example, one person may prefer representations to have various shapes and colours, while this may distract or even prevent another person from understanding the key message. Moreover, one person may prefer or find it easier to interpret examples, while another person would rather look at a formal model. Finally, different people have different backgrounds, levels of education and experiences – while one person may be very experienced with inverting matrices in high-dimensional vector spaces, another person may have more experience in reading candlestick charts. Therefore, what is interpretable to one person may not be interpretable to the next person and vice-versa.
\\Because of the many aspects involved in interpretability, we settle for the simple, yet elegant definition for the context of this work given by \cite{miller2017}. He defines interpretability as "the degree to which an observer can understand the cause of a decision". This means that the interpretability of a model is higher if it is easier for a person to understand and trace back why a decision (or prediction) was made. Comparatively, a model is more interpretable than another model, if its decisions are easier to understand than the decisions of the second model \citep{miller2017}.

Having established a meaningful definition and understanding of the term interpretability, we can now delve into the next subsections, where we investigate its value, scope and goals.

%% ===========================
\subsection{The Value of Interpretability}
\label{ch:Content1:sec:Section2:Subsection1}
%% ===========================
As we have seen in the introduction, in conventional machine learning, where no EM is applied (as in figure \ref{figure 1.2} on the left), the user of the model simply gets the prediction for the given input. This prediction answers the "what" question, i.e. what is the prediction for my input. However, it is not only desirable to get a prediction, but also an explanation for the prediction (as in figure \ref{figure 1.2} on the right). The explanation is responsible for answering the "why" question, i.e. why did the model make this prediction. Interpretability is, thus, established by answering the "why" question of the prediction \citep{molnar2018}.

Many reasons make interpretability a valuable and sometimes even an indispensable feature. However, not all ML systems require interpretability, sometimes being able to guarantee a high predictive performance is sufficient \citep{doshi2017_rigorous_science}. As an example of such systems, we can mention low-risk systems, such as product recommenders or advertising servers, where a mistake has no severe or even fatal consequences \citep{molnar2018}. Another example of systems that do not require interpretability are systems that compute their output without having any human intervention, such as postal code sorting machines or aircraft collision avoidance systems. Based on \cite{doshi2017_rigorous_science}, there are generally two situations where interpretability and thus explanations are not necessary: (1) when there are no significant or severe consequences for unacceptable results, and (2) when a problem is sufficiently well-studied and validated in real-world applications, so that we can trust the system even if it is not always accurate.

Now that we have learned in which situations interpretability is most desired, we can continue by exploring some of its advantages and values. First of all, interpretability is a means to satisfy human curiosity and learning \citep{miller2017}. Of course, we do not need an explanation for everything that we see; most people, e.g. do not need or want to understand how a car or a refrigerator exactly works. However, we are mostly intrigued by unexpected events that make us curious. In the case of an unexpected prediction made by a BB model, we would like to understand why the model made this prediction, as reaching this understanding allows us to update our mental models and thus, to learn. Especially in the research domain, scientific findings often stay in the dark if our models only give predictions without explanations \citep{molnar2018}.
\\Another value of interpretability is that it helps us to find meaning in the world \citep{miller2017}. Decisions based on ML models are increasingly affecting our life, which is why it is important to explain their behaviour. If e.g. a bank's ML model rejects a loan application, the applicant most likely wants to know why it got rejected. Under the new GDPR, in the EU this applicant also has a so-called \textit{right to be informed} \citep{goodman2016,wachter2017}. This means that the applicant has a right to be explained the decision (output of the algorithm), or at least the involved logic, and if the bank fails to comply with this regulation, it is subject to legal penalties. Internet services are also increasingly adding explanations for their recommendations, as in the case of, e.g. online shopping and movie recommendations \citep{molnar2018}. These explanations are usually in the form of "we recommend you to watch this movie or buy this product because other users with similar preferences and profiles also enjoyed the recommended movie or product".
\\The above example also brings us to our next value of interpretability, which is social acceptance. Already a long time ago, \cite{heider1944experimental} have shown that people attribute beliefs and intentions to abstract objects. It is therefore apparent that people more likely accept models that are interpretable. This argument is also supported by \cite{ribeiro2016}, as the authors state that "if the users do not trust a model or a prediction, they will not use it". The fundamental idea is that by establishing interpretability through an explanation, users gain more trust in a model and therefore are more likely to accept and use it.
\\Finally, we highlight another critical value that is enabled by interpretability – safety \citep{miller2017}. Only through interpretability can ML models be debugged and audited, which allows us to increase their safety. In this context, safety can mean multiple things; we could call a model safe if it does what we expect it to do, e.g. in the case of autonomous vehicles, we expect the system to safely take passengers from one point to another, bypassing all obstacles and challenges it may face on its way. However, we could also call a model safe that features no biased behaviour that leads to the discrimination of minority groups, which would make it safe against discrimination. In any case, only through interpretability, can we detect faulty model behaviours, which allows us to fix the system, thereby increasing its safety \citep{molnar2018}.

After establishing some general values of interpretability, in the upcoming subsection, we look at the importance of interpretability in the data mining (DM)\nomenclature{DM}{Data Mining} process in greater detail. Moreover, in the subsection after that, we discuss ethical and societal concerns regarding ML and reinforce the idea that interpretability can be a step in the right direction to address these issues.

%% ===========================
\subsubsection{Value in the Data Mining Process}
\label{ch:Content1:sec:Section2:Subsection1:Subsubsection2}
%% ===========================
In this section, we look at how interpretability changes the conventional ML pipeline, thereby adding value to the DM process, as can be seen in the abstracted and simplified
%%%%%%%%%%%%% Figure %%%%%%%%%%%%%
\begin{wrapfigure}{l}{0.66\textwidth}
  \vspace{-20pt}
  \begin{center}
    \includegraphics[width=0.66\textwidth]{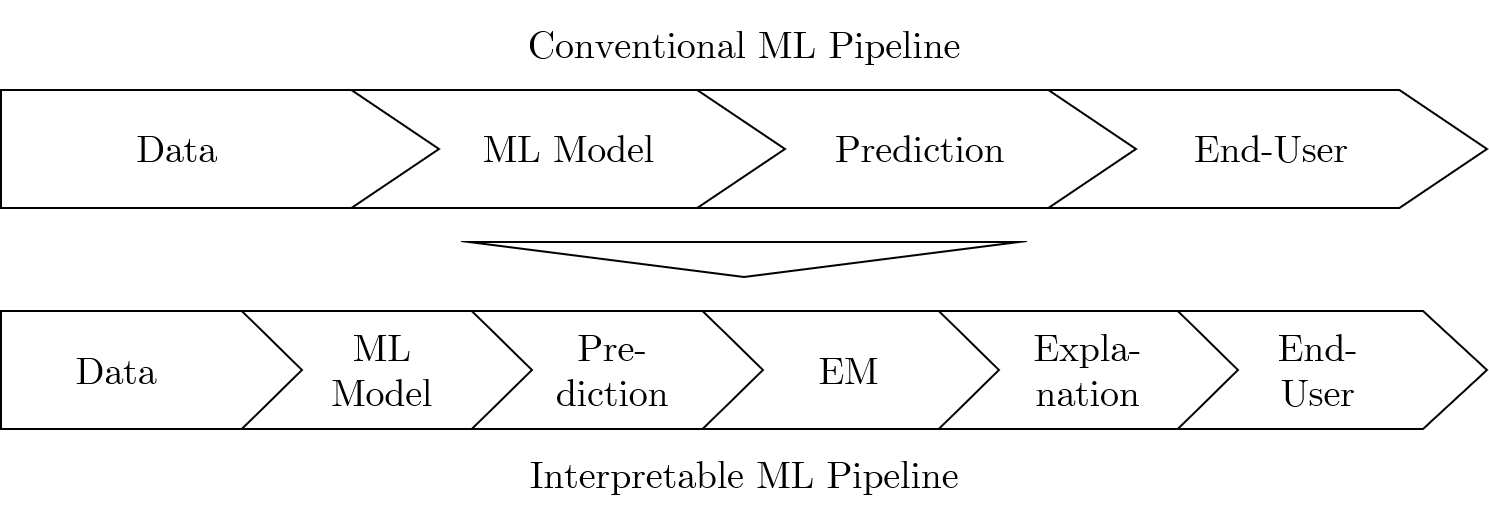}
  \end{center}
  \vspace{-20pt}
  \caption{Paradigm Shift: From a Conventional to an Interpretable ML Pipeline}
  \label{figure 2.5}
  \vspace{-10pt}
\end{wrapfigure}
%%%%%%%%%%%%% Figure %%%%%%%%%%%%%
figure \ref{figure 2.5} to the left. On the top of the figure, we have the conventional ML pipeline which consists of four steps. First, the data is collected, prepared and labelled to then train a machine learning model on the data. Some of the data is kept from the model to assess its predictive performance later. We can change, tweak and adjust the model until we are happy with its results. Once we achieve this, the model is deployed in practice and used to generate predictions on new data. These model predictions can finally be employed by the end-users, to make decisions based on these and on other knowledge that they may have. The question that however remains in this scenario is the following: do end-users trust this BB model enough, to make decisions based on its predictions, i.e. do they blindly follow its recommendations.
\\In the second scenario of the interpretable ML pipeline, we add two additional steps to the pipeline: an explanation method and an explanation. This is a paradigm shift in the sense that it changes the workflow in such a way that the end-user can now interpret and thus understand the predictions of the BB model. This greatly increases trust and willingness to act based on the given data, as we have already seen in the introduction \citep{lundberg2017,ribeiro2016}. So the real value of establishing interpretability is that end-users can make more informed decisions without needing to trust an opaque ML model blindly.

With a broad understanding of the value of interpretability, in the upcoming subsection, we critically analyse ethical and societal implications that result from the increasing use of ML technologies and how interpretability can lessen the negative burden.

%% ===========================
\subsubsection{Ethical and Societal Concerns}
\label{ch:Content1:sec:Section2:Subsection1:Subsubsection1}
%% ===========================
A machine learning model is not inherently "bad", however, it is also only as good as the information that it is trained with. This means that if we feed real-world data to a model, which contains biased or discriminative instances, then it learns these patterns and returns predictions that are inclined towards such behaviour. Consequently, this can lead to the discrimination of several minority groups \citep{beillevaire2016}.

The author of the famous book "Weapons of Math Destruction", Cathy O'Neil, argues that compared to human decision making, which can adapt and thus evolve, machines "stay stuck in time until engineers dive in to change them" \citep{oneil2017weapons}.
\\If we, e.g. would still nowadays use a hypothetical ML model that was established in the 1960s to process college applications, then still many women would not be going to college. This would happen as the model was trained on "old" data that was largely inclined towards successful men.
\\Another illustrative example given by \cite{oneil2017weapons} is related to loan applications. Imagine that a bank creates a ML model that amongst other information, incorporates sensitive data related to race, religion, gender, disease, disability, credit scores, location and so forth. What could, e.g. happen next is that poor people, who are more likely to live in a high-crime neighbourhood, would be less likely to receive a loan from that bank. This could occur even if they were to fulfil all the necessary criteria to be theoretically eligible for a loan. Even worse, these people could specifically be targeted after that with predatory ads for subprime loans, mortgages, car and cellphone loans, and for-profit schools. As a result, these people would likely get more and more indebted over time, going down the vicious "death spiral" \citep{oneil2017weapons}.
\\This is precisely where the problem mentioned above lies – ML systems codify the past, while not having the capability to evolve or invent the future. This is a skill that requires moral and ethical standards and imagination, which, in the narrow sense, so far only humans can provide \citep{oneil2017weapons}.

The first step in the right direction that we must take is to conduct algorithmic audits. So instead of treating ML models as black boxes, we must establish their interpretability and transparency, to understand their inner workings and to judge their fairness. Moreover, we should directly embed ethical values into these models, even it that could sometimes mean to prioritise fairness over accuracy and profit \citep{oneil2017weapons}.
\\The issues discussed above can at least be partially addressed and overcome by using explanation methods, as they show the dominant forces that have the highest influence on a specific decision (or prediction) \citep{beillevaire2016}. This allows us to trace back inputs that went into the model and to understand how they influenced the outcome. Thereby, we could identify models that show a discriminative behaviour and improve them accordingly. Lastly, \cite{oneil2017weapons} mentions that it is also part of the responsibility of the creators of these models and their supervisors, to make moral choices on the data that they integrate into a model and the data that they choose to leave out.
\\To conclude this section in the words of \cite{oneil2017weapons}: "in the end, mathematical models should be our tools, not our masters". We must, hence, continuously invest our efforts to keep it that way, by correcting and improving faulty, biased and discriminatory black box machine learning models.

After studying the value of interpretability and understanding all that it entails, in the upcoming section, we talk about the scope of interpretability.

%% ===========================
\subsection{The Scope of Interpretability}
\label{ch:Content1:sec:Section2:Subsection2}
%% ===========================
In this section, we analyse the most important types of interpretability. For our purpose, we use a division into three major types, which is consistent with \cite{molnar2018}: (1) Algorithm interpretability, (2) Global model interpretability, and (3) Local model interpretability. We describe these three types over the next three subsections.
\\Other authors such as \cite{lipton2016} only consider two major types of interpretability: Transparency and post-hoc interpretability, whereas each of these two types features finer sub-types. As we also use the two terms of transparency and post-hoc interpretability in the present work, we start by defining their meaning. In the broad sense, transparency answers the question of \textit{how the model works}, and post-hoc interpretability answers the question of \textit{what else can the model tell us} \citep{lipton2016}.
\\In the context of ML, we understand transparency as the capacity of the model to convey understandable information regarding the mechanisms by which it works. It is thus the opposite of opaqueness or the already mentioned black box principle. \cite{lipton2016} considers three distinct sub-types of transparency, which we analyse in the next two subsections: algorithmic transparency, simulatability and decomposability.
\\Post-hoc interpretability represents a different approach to extract information from models. It means that we explain the decisions of a model after it has been built and used to compute these decisions (predictions). Compared to the transparency concept, post-hoc interpretations mostly do not show the exact inner workings of a model. However, they have the advantage that we can interpret BB machine learning models after they have been created, without sacrificing their predictive power \citep{lipton2016}. As examples of post-hoc interpretability tools, we can mention text explanations, visual explanations, explanations by example and local explanations.

In the following two subsections, we discuss algorithm interpretability and global model interpretability, which fall into the category of transparency, as defined by \cite{lipton2016}. After that, we look at local model interpretability, which belongs to the category of post-hoc interpretability tools.

%% ===========================
\subsubsection{Algorithm Interpretability}
\label{ch:Content1:sec:Section2:Subsection2:Subsubsection1}
%% ===========================
Algorithm interpretability is one of the possible types of interpretability over which a ML model can dispose. Having an interpretable algorithm means that we can understand how it learns a model from the given input data and what types of relationships it is capable of modelling \citep{molnar2018}. To fulfil this condition, we should thus be able to trace at each step what the algorithm exactly does and understand how it converges towards a solution. This notion also implies that if an algorithm fulfils the transparency condition, we have to understand neither the learned model on a global scale nor how the algorithm computes individual predictions. The focus is solely on understanding the model creation process. This definition given by \cite{molnar2018} can be seen as equal to the definition of algorithmic transparency given by \cite{lipton2016}.
\\As an example of an algorithm that scores high in transparency, we can mention the ordinary least squares (OLS) method, which is well-studied and understood, and used, e.g. in linear regression analysis. As a counterexample, we can mention deep learning methods, such as deep neural networks, as they lack the necessary algorithmic transparency for us to understand how they work fully \citep{lipton2016}. We can thus not guarantee that the application of these to new problems is successful.

After discussing the most critical aspects of algorithm interpretability, we proceed with analysing two types of global model interpretability in the next subsection.

%% ===========================
\subsubsection{Global Model Interpretability}
\label{ch:Content1:sec:Section2:Subsection2:Subsubsection2}
%% ===========================
In global model interpretability, the goal is to understand how a trained model makes predictions and which parts of the model influence this outcome the most. This represents a crucial difference to the previously discussed algorithm interpretability. In literature, we can find different terminologies used for the same two types of global model interpretability. The notion mentioned above of simulatability, as defined by \cite{lipton2016}, corresponds to the notion of global holistic model interpretability, as defined by \cite{molnar2018}. Moreover, decomposability corresponds to global model interpretability on a modular level. In the following, we analyse these two sub-types of interpretability.
\\Simulatability means that the entire ML model is transparent, i.e. a person can grasp the whole model at once \citep{lipton2016}. This implies that the observer can understand how the model makes decisions based on learned features, weights, parameters, structures and interactions of features \citep{molnar2018}. We, therefore, suggest that for a model to fulfil the simulatability condition, it needs to be simple enough. Again, the required simplicity depends on the user of the model, however, given the limited capacity of our cognition, this ambiguity would certainly disappear after a few orders of magnitude \citep{lipton2016}. For all these reasons, simulatability is arguably hard to achieve in practice, as models become complex very quickly, thereby exceeding the average human's short-term memory capacity \citep{molnar2018}. Finally, we cannot generally say that e.g. a linear regression or a decision tree always fulfils simulatability, whereas a neural network is never able to satisfy this condition. We could, however, say that e.g. for a person of normal cognition, a linear regression with at most two dependent variables or a decision tree with a maximum depth of 2 and four child-leaves, fulfil the simulatability condition.
\\Decomposability requires the model to be transparent at the level of individual components, i.e. each part of the model, such as inputs and parameters, admit an intuitive explanation. This notion is in accordance with the definition of \textit{intelligibility} as described by \cite{lou2012}. Decomposability implies that inputs that are used to train a model must be interpretable themselves \citep{lipton2016}. Therefore, models that, e.g. use highly engineered, anonymous or opaque features cannot fulfil this condition. An example of a model that mostly fulfils intelligibility or decomposability is the Generalized Additive Model (GAM)\nomenclature{GAM}{Generalized Additive Model}. As described by \cite{lou2012} GAMs combine single-feature models, based on linear and more complex non-linear shape functions, through a linear linkage function. Consequently, GAMs can become arbitrarily complex by modelling non-linear relationships, thereby outperforming linear models. This is achieved while retaining much of the interpretability of linear models since GAMs do not incorporate interaction effects between features. Finally, the reason why full complexity models, such as neural networks with multiple hidden layers or tree ensembles, can yield more accurate results than GAMs is that they incorporate both – non-linearity and interaction effects \citep{lou2012}.

Having a good understanding of algorithm and global model interpretability, we proceed with discussing what local model interpretability entails in the next subsection.

%% ===========================
\subsubsection{Local Model Interpretability}
\label{ch:Content1:sec:Section2:Subsection2:Subsubsection3}
%% ===========================
The last type of interpretability that we cover in this master thesis is local model interpretability. In this type of interpretability, the general idea is to approximate a small region of interest in a complex and accurate model with a simpler model. This local model usually does not provide an optimal solution, however, a reasonably good approximation, while preserving high interpretability \citep{ruping2006}. We can distinguish between two sub-types: Local interpretability for a single prediction, and local interpretability for a group of predictions \citep{molnar2018}.
\\In the former sub-type, we essentially zoom in on a single instance and try to understand how the model arrived at its decision. This is often achieved by looking at predictions for similar instances that have been slightly perturbed \citep{molnar2018}. A prominent approach that follows this strategy is the Local Interpretable Model-Agnostic Explanations (LIME)\nomenclature{LIME}{Local Interpretable Model-Agnostic Explanations} model by \cite{ribeiro2016}. To make a practical example, imagine that we have a complex and accurate model that predicts house prices. Naturally, the price of houses depends on many factors such as location, size, amenities and so forth; therefore, there may not be a linear relationship between the house price and its size. If we, however, look at a specific category of houses, say the ones that have around 200m$^2$, then there is a good chance that if we only slightly change the size of the house, around 190 and 210m$^2$ that this subregion of our data and thus our model is approximately linear. In that case, the price would linearly go up and down with the house size for that subregion. This behaviour would probably not be observable anymore if we, e.g. allowed a house size variation of 150 to 250m$^2$. Due to this phenomenon, local explanations can be more accurate compared to global explanations, while being more intuitive for us to interpret, as humans also exhibit this type of interpretability in their decision processes \citep{molnar2018,lipton2016}.
\\In the second sub-type of local interpretability for a group of predictions, we can fundamentally proceed in two distinct ways. We can choose to either employ global methods to interpret the group of predictions or local methods \citep{lipton2016, molnar2018}. In the first case, we would take the global model and apply it only to the group of predictions, while pretending that these constitute the entire dataset. After that, we would utilise the global methods on this subset to generate the interpretations. In the second case, we would use the local explanation methods for individual predictions and apply these sequentially to all instances of the group. Afterwards, we could aggregate
%%%%%%%%%%%%% Figure %%%%%%%%%%%%%
\begin{wrapfigure}{r}{0.40\textwidth}
  \vspace{-20pt}
  \begin{center}
    \includegraphics[width=0.40\textwidth]{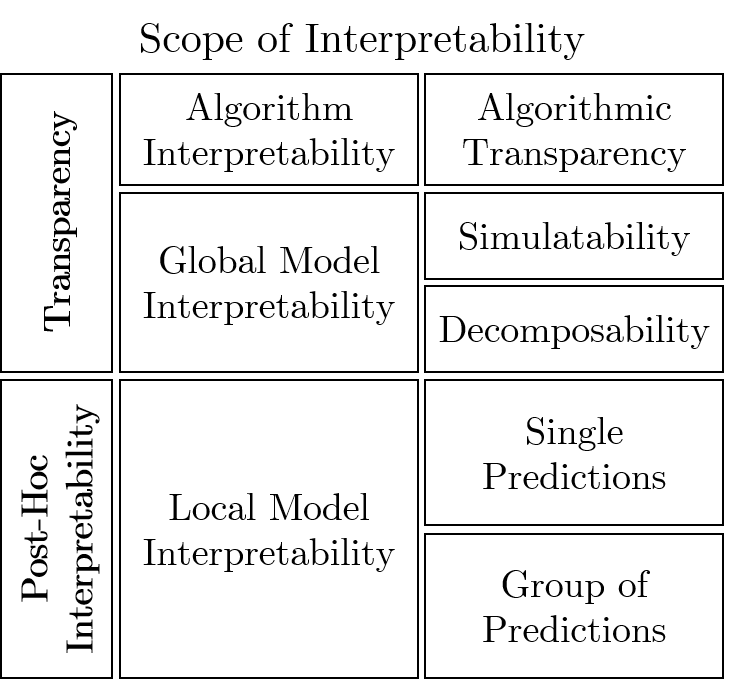}
  \end{center}
  \vspace{-20pt}
  \caption{Overview: Discussed Interpretability Types}
  \label{figure 2.6}
  \vspace{-10pt}
\end{wrapfigure}
%%%%%%%%%%%%% Figure %%%%%%%%%%%%%
and join these interpretations. This second approach is the one that we use later on in the experiments chapter to evaluate our developed Explanation Consistency Framework. In figure \ref{figure 2.6} on the right, we summarised the interpretability types that were covered in this thesis so far. There are more post-hoc interpretability methods, such as the previously-mentioned textual and visual explanations and explanations by example. We occasionally recur to visual explanations in the experiments section; however, we disregard the other two techniques for the remainder of this research. Moreover, we concentrate on post-hoc interpretability tools in general, and thus also disregard the transparency-related algorithm and global model interpretability for the remainder of this work.

With a solid understanding of the different types of interpretability, we continue discussing how we can proceed in its evaluation. Furthermore, we learn about the goals and objectives of interpretability in the upcoming section.

%% ===========================
\subsection{Evaluating Interpretability Quality}
\label{ch:Content1:sec:Section2:Subsection3}
%% ===========================
So far, we have discussed many aspects of interpretability – its value and significance in the DM process, how it helps in addressing ethical and societal concerns and lastly, what types of interpretability there are. In this last section of this topic, we finally discuss how we can evaluate the quality of interpretability and what its corresponding goals are.

Even though the volume of research in ML interpretability is rapidly growing \citep{doshi2017_rigorous_science}, there is no universal consensus on its exact definition and on ways to measure it \citep{bibal2016,molnar2018}. However, in their article "Towards a Rigorous Science of Interpretable Machine Learning", \cite{doshi2017_rigorous_science} conducted some first research on this specific topic and formulated a framework which enables us to categorise different interpretability evaluation approaches. Their proposed taxonomy for interpretability evaluation consists of three levels: (1) Application-grounded evaluation, (2) Human-grounded evaluation, and (3) Functionally-grounded evaluation.
\\In application-grounded evaluation, we fundamentally use the explanation in a real-world application and let domain experts test and evaluate it \citep{doshi2017_rigorous_science}. For example, we could use it in a radiology software that shows where a bone fracture occurred in an x-ray image \citep{molnar2018}. This would then be directly confirmed or corrected by a radiologist. So on this level, we specifically evaluate the quality of an explanation with the end-task in mind. This requires high standards concerning experimental design and domain expert knowledge, which makes it expensive, but also a strong success indicator in case of a positive experiment outcome.
\\On the second level, of human-grounded evaluation, we proceed in fundamentally the same way as in application-grounded evaluation, however, with significant simplifications \citep{doshi2017_rigorous_science}. This means that the experiments are not conducted with domain experts, but with non-specialists. An experiment that we could conduct in this context is, e.g. one of (binary) forced choice. This translates into showing humans different explanations and asking them to choose the best one, i.e. the one that is the most intuitive to them. So on this level, we do not have a specific end-goal in mind. Human-grounded evaluation is, therefore, most appropriately used, when one wishes to test more general notions of the explanation quality. These experiments are less expensive than the previously described ones, as no domain experts with specialised knowledge are required, which also makes it easier to find more participant.
\\For situations in which human experiments may not be feasible or ethical, functionally-grounded evaluations come into play. In experiments on this level, instead of using humans, some formal definition of interpretability serves as a proxy to evaluate the explanation quality \citep{doshi2017_rigorous_science}. As an example, researchers may have found that users best understand decision trees. For this case, a proxy for explanation quality could be the depth of a tree together with the number of its child leaves. A shorter tree with fewer child leaves would then get a higher rating for explanation quality \citep{molnar2018}. However, in this scenario, one would also have to add essential constraints to guarantee that the predictive power of the smaller tree remained on a reasonable level, and would not decrease significantly compared to its bigger version.  This form of evaluation works best when the used model class has already been validated by someone else in at least a human-grounded setting. Moreover, it may also be appropriate to apply when a method has not yet reached a stage of maturity. The tricky task in this type of evaluation amounts to determining reasonable proxies for explanation quality \citep{doshi2017_rigorous_science}. Numerous proxies are imaginable, such as model sparsity, which relates to the number of features that are used by an explanation, out of the total number of features in the model \citep{molnar2018}. Needless to say that an explanation with five features is more straightforward to interpret than an explanation with a hundred. Other examples could, e.g. be uncertainty (i.e. does the explanation let us know when it is unsure), interaction (i.e. is the explanation capable of including and showing interaction effects between features) and processing time (i.e. how long does it take the average user to understand the explanation) \citep{molnar2018}. We base the Explanation Consistency Framework developed in this thesis project on this type of functionally-grounded evaluation methods.

After discussing some critical aspects related to the evaluation of interpretability quality, in the following subsection, we are going to continue analysing what goals and objectives we wish to achieve by establishing interpretability.

%% ===========================
\subsubsection{Goals of Interpretability}
\label{ch:Content1:sec:Section2:Subsection3:Subsubsection1}
%% ===========================
As we have seen before, the quality of interpretability can be evaluated on different levels. Moreover, we have established that there is no consensus on how to exactly measure it \citep{bibal2016,molnar2018}. We can however question ourselves, what we want to achieve with interpretability. To answer this question, \cite{ruping2006} first noted that interpretability is composed of the three following sub-problems or goals: accuracy, understandability, and efficiency (figure \ref{figure 2.7}). \cite{ruping2006} affirm that these three goals are connected and often also competing.
\\Accuracy means that there must exist an actual connection between the given explanation by the EM and the prediction from the ML model \citep{bibal2016}. If this goal
%%%%%%%%%%%%% Figure %%%%%%%%%%%%%
\begin{wrapfigure}{l}{0.40\textwidth}
  \vspace{-20pt}
  \begin{center}
    \includegraphics[width=0.40\textwidth]{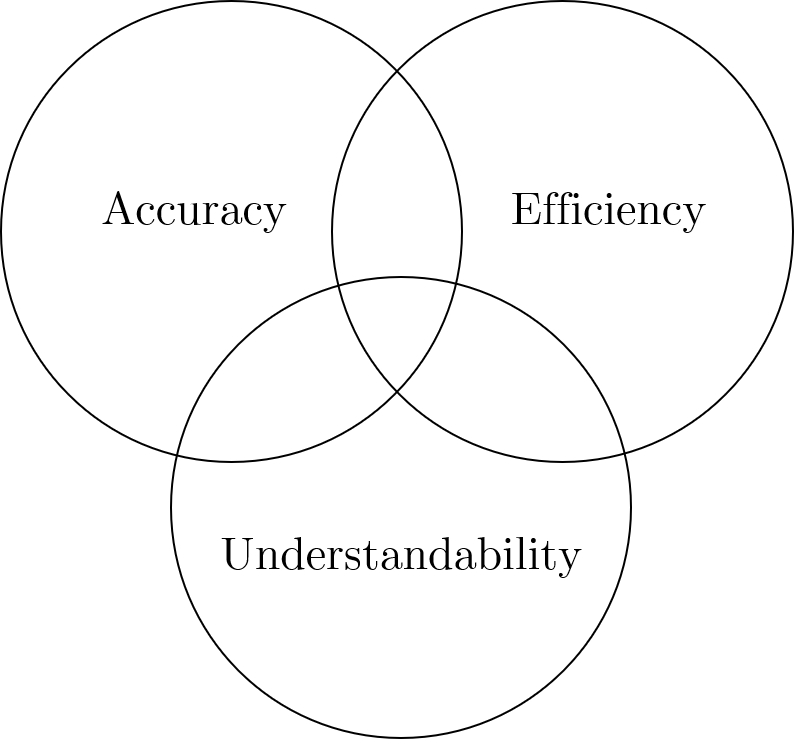}
  \end{center}
  \vspace{-20pt}
  \caption{The Three Goals of Interpretability [based on \cite{ruping2006}]}
  \label{figure 2.7}
  \vspace{-10pt}
\end{wrapfigure}
%%%%%%%%%%%%% Figure %%%%%%%%%%%%%
is not achieved, then the explanation is useless, as it explains something that has no connection to the data and thus to the reality.
\\With regards to understandability, \cite{ruping2006} say that a given explanation must be of such form that an observer can understand it. This is another crucial goal of interpretability because if an explanation is accurate, however, not understandable, it is also not of much use \citep{bibal2016}.
\\Finally, efficiency is related to the time necessary for a user to grasp the explanation \citep{ruping2006}. Obviously, without this condition, it could be argued that almost any model is interpretable, given an infinite amount of time \citep{bibal2016}. An explanation must thus be understandable in a finite and preferably short amount of time, to achieve interpretability.
\\High interpretability would thereby be scored by an explanation that is: accurate to the data and to the model, understandable by the average observer and graspable in a short amount of time.
\\The Explanation Consistency Framework developed in this research, addresses exactly these three dimensions and goals of interpretability, as we see in chapter 4.

All subtopics of interpretability discussed over the previous sections constitute a solid knowledge basis to understand the proposed ECF in this thesis. Hereafter, we delve into the closely related topic of explanations, as they are the necessary means to establish interpretability.

%% ===========================
\section{Explanations as Tools to Establish Interpretability}
\label{ch:Content1:sec:Section3}
%% ===========================
There have been numerous attempts to define the concept of an explanation across different disciplines, reaching from philosophy and sociology, to mathematics and statistics. An appropriate definition for the term is thus dependent on the application domain and should be formulated with regards to the context. The Oxford Dictionary more generally defines an explanation as "a statement that makes something clear"\footnote{\url {https://en.oxforddictionaries.com/definition/explanation}}. As we have seen before, this is exactly what we want to achieve through an explanation – to make something clear, transparent, interpretable and understandable. For the context of this research, we thus use the simple yet goal-oriented definition given by \cite{miller2017}, who states that "an explanation is the answer to a why-question". When a ML model outputs an (in-)comprehensible prediction, we question ourselves why the model made this prediction. An explanation is thus the means by which we explain the decisions made by a machine learning algorithm. This is the crucial difference between an explanation and interpretability. If we recall from the previous section, interpretability could mean either – transparency (i.e. insights on the inner workings of the ML model) or post-hoc interpretability (i.e. insights on the decision processes of the ML model with regards to a single input). We can, therefore, see interpretability as the end-goal that we want to achieve and explanations as tools to get us there. In the case of transparency, the model itself serves as the explanation, as we see later on. In post-hoc interpretability, the explanation has to be generated based on the instance that we want to explain and the ML model used in its prediction. Methods that generate explanations for post-hoc interpretability are thus denoted as \textit{explanation methods} for the remainder of this thesis.

\cite{miller2017} stresses that an explanation is not only a product but also a process that involves a cognitive and a social dimension.
\\In the cognitive dimension the actual explanation is derived by a process of abductive inference \citep{miller2017}. This means that first the causes of an event are identified, often with regards to a particular case, and then a subset of these causes are selected as the explanation. An explanation itself, thus, usually does not contain all influencing factors that contributed to a decision or prediction, but only the most important ones \citep{molnar2018}. Otherwise, we would have a complete causal attribution, which can, however, also make sense if we are, e.g. legally required to state all influencing factors that contributed to a decision of a ML model \citep{molnar2018}.
\\The social dimension of an explanation is related to the social interaction, in which knowledge is transferred from the explainer (i.e. the giver of an explanation) to the explainee (i.e. the receiver of an explanation) \citep{miller2017}. The primary goal of this interaction is that the explainee receives enough information from the explainer to understand the causes of the event or the decision \citep{miller2017}. The social context can thereby have a significant influence on the actual content of an explanation \citep{molnar2018}. The explainer can be both in our case – a human or a machine.

Having a good understanding of the explanation concept and its distinction to interpretability, we proceed with analysing what a "good" explanation constitutes regarding human understandability, in the next subsection. After that, we give two examples where the models themselves serve as the explanation to establish interpretability. Finally, we conclude this sub-chapter with the analysis of model-dependent and model-agnostic explanation methods, to establish post-hoc interpretability.

%% ===========================
\subsection{Recipe for a "good" Human-Style Explanation}
\label{ch:Content1:sec:Section3:Subsection1}
%% ===========================
Now that we have understood and narrowed down the explanation topic to our research, we talk about important ingredients that are necessary to produce good explanations for humans. In his works "Explanation in Artificial Intelligence: Insights from the Social Sciences", \cite{miller2017} affirms that "most work in explainable artificial intelligence uses only the researchers' intuition of what constitutes a 'good' explanation". There are, however, vast bodies of research in philosophy, psychology and the cognitive sciences on how people select and evaluate explanations \citep{miller2017}. In the following, we thus analyse some of these aspects, in a non-exhaustive way, based on the research conducted by various researches from different fields.

Commonly stated as a fundamental property of a good explanation is counterfactual faithfulness \citep{wachter2017counterfactual,doshi2017,molnar2018}. Humans have their mental models with which they guesstimate an outcome. If this outcome is different from the outcome suggested by the ML model, then we start to question it. This means that people usually do not ask why a specific prediction was made, but rather why this prediction was made instead of another one, namely the one we expected \citep{lipton2016}. As an example, if our mortgage application gets rejected, we are more interested in knowing the factors that need to change to get it accepted, rather than the factors that generally lead to a rejection. What we thus want to know is the difference between our application and the would-be-accepted version of our application \citep{molnar2018}. Therefore, explanations should feature some contrast between the instance to explain and a reference point, such as the average over all instances. These contrastive explanations are more natural for us to understand than a complete list of all factors. However, they are also application-dependent, as we require some reference point for comparison \citep{molnar2018}.

Another property of a reasonable explanation concerning human understandability is selectivity. People usually do not expect an explanation to contain all causes of an event \citep{miller2017}. Due to the sometimes infinite number of causes that an event can have, it can be difficult or impossible to gather all of these and to process them in our minds. Humans are therefore inclined to select only one or two causes to constitute \textit{the} explanation \citep{miller2017}. The Rashomon Effect describes precisely these situations, in which different causes can explain an event \citep{molnar2018}. In sum, explanations generated by an EM should, in general, not return a long list of causes but instead focus on the few most important ones.

The third important aspect of a good explanation that we discuss is to focus on abnormality \citep{miller2017}. When we talked about the values of interpretability, we established that especially in situations of unexpected outcomes that violate our current understanding, we would like an explanation. This explanation allows us to update our mental models and thus to learn \citep{miller2017}. As an example, imagine that we are predicting house prices and one particular house is estimated to be very expensive. This house features an exceptional number of three balconies, unlike any of the other houses in the data. Now, the explanation method may find that the above-average size of the house, the excellent location or the recent renovation contribute in equal amounts to the high price as the three balconies. However, the single best explanation for the high price could still be that the house has three balconies, as this is an unseen trait in all other houses \citep{molnar2018}. For explanation methods, we can thereby conclude that they should attribute higher importance in their explanations to unusual features if these have a significant impact on the prediction output.

The last important aspect that we discuss in detail is truthfulness \citep{molnar2018}. We have mentioned before that an explanation is of little use if it is understandable, however, not accurate to the data \citep{bibal2016}. Thereby, a reasonable explanation must prove to be right in reality. Curiously, however, for humans, this is not the most critical factor for a good explanation. As an example, selectiveness, which omits part of the truth is more important than truthfulness, as it dramatically increases the understandability of an explanation \citep{miller2017,molnar2018}. Nevertheless, the explanation should still predict an event as truthfully as possible. Applied to our house problem from above, if the value of one house goes up because it features three balconies, then this should hold true for at least similar houses \citep{molnar2018}.

Other aspects contribute to a good explanation, such as the adaptation to the social context in which it is used. The context determines the content and complexity of the explanation and should therefore also be considered \citep{miller2017}. This requirement is, however, difficult to translate into a generalizable ML framework. Moreover, good explanations should be coherent with the preceding beliefs of the explainee, which is supported by the fact that humans often ignore information that does not correspond to their prior knowledge \citep{molnar2018}. It is however in a trade-off with truthfulness, as prior knowledge of people is often not generally applicable and only valid in a specific domain. We thus argue that it would be counter-productive to include these types of conditions and aspects in an explanation method.

In the Explanation Consistency Framework, developed in chapter 4 of this thesis project, we build on these desirable characteristics of explanations, while further exploring other important ones that relate to the previously discussed goals of interpretability. After familiarising ourselves with the above-mentioned humanly-desirable properties of explanations, we continue to discuss how interpretable ML models can simultaneously be the prediction models and explanations in the upcoming section.

%% ===========================
\subsection{Interpretable Models and Explanations}
\label{ch:Content1:sec:Section3:Subsection2}
%% ===========================
When we talked about global model interpretability, we already stressed that we cannot generally say that e.g. a linear regression or a decision tree are always interpretable models. Nevertheless, as long as they remain on a manageable level of complexity, they are straightforward to interpret and fulfil at least the decomposability criterion, and sometimes even simulatability. Apart from these two methods, there are others, such as the logistic regression and the previously-mentioned GAM's, which can also be counted to the class of algorithms that create interpretable models \citep{molnar2018}.

In the following two sub-sections, we present two simplified examples of interpretable models, which serve as prediction models and explanations simultaneously. We mainly focus on how to interpret these models, but also give some background information on their properties and assumptions.

%% ===========================
\subsubsection{Linear Regression}
\label{ch:Content1:sec:Section3:Subsection2:Subsubsection1}
%% ===========================
Linear models are well-studied, understood and due to their high degree in transparency and interpretability, often used in practice \citep{friedman2009}. The linear regression itself is interpretable on a global level, when it only has a single or a few variables, and on a modular level, when it features multiple variables. This high interpretability is mainly facilitated by its nature of only learning linear relationships. The regression model is moreover monotone as a result of its linearity, which means that an increase in a feature consistently leads to either an increase or a decrease of the target variable, but never both for this feature \citep{molnar2018}. One property that linear regression models do not include out of the box is interactions between features. These can be manually added, but they may hurt the interpretability \citep{molnar2018}.

The linear regression model features several assumptions that we briefly address hereafter. These should be fulfilled to guarantee the validity of its results \citep{molnar2018}. First, we have the assumption of \textit{linearity}, which requires the representability of the target variable as a linear combination of the used features. If we have non-linear relationships, it may make more sense to use a different, non-linear model. Second, \textit{normality} assumes that the target variable follows a normal distribution, given the independent variables. Third, we assume constant variance or \textit{homoscedasticity} in more technical jargon, which means that the variance of the error term must be constant over the entire feature space. Fourth, the \textit{independence} between all input instances is required. The fifth assumption of \textit{multicollinearity} specifies that different features used in the model must not be highly correlated. We thus want the absence of multicollinearity. Finally, the sixth assumption of \textit{fixed features} requires the independent variables to be fixed, i.e. not carrying any error or variation \citep{molnar2018}. Some of these assumptions are easily fulfilled, while some can represent challenging obstacles. In practice, we should always verify these and try to fulfil them as well as possible to avoid modelling mistakes or overestimating the performance of our model.

As our main focus is on interpretability, we assume the fulfilment of the given conditions from above in the following example. Also, rather than presenting a generic formula for a
%%%%%%%%%%%%% Figure %%%%%%%%%%%%%
\begin{wrapfigure}{r}{0.40\textwidth}
  \vspace{-25pt}
  \begin{center}
    \includegraphics[width=0.40\textwidth]{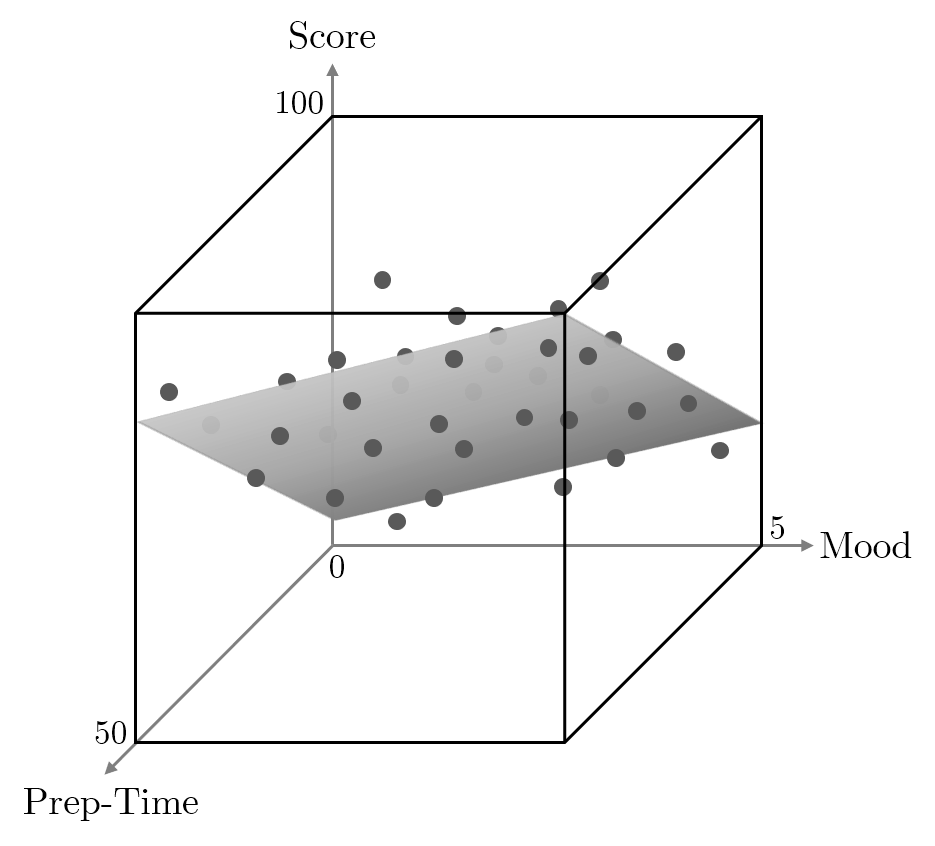}
  \end{center}
  \vspace{-20pt}
  \caption{Interpretable Model 1: Linear Regression}
  \label{figure 2.8}
  \vspace{-15pt}
\end{wrapfigure}
%%%%%%%%%%%%% Figure %%%%%%%%%%%%%
linear regression model, we directly apply it to our context. For further details on linear regression in general and on its calculation, we recommend the book by \cite{friedman2009}.
\\In our example, the situation is the following: we want to predict the exam scores of this year, based on two features – preparation time (prep-time) and mood. The score is thus the dependent variable, and prep-time and mood are the independent variables, which we use to predict the expected score that students achieve in the exam. In figure \ref{figure 2.8}, we have a representation of the described situation. On the x-axis, we have the prep-time feature, whereas students are known to invest between 0 and 50 hours for their preparation. On the y-axis, the mood-level of students right before the exam is shown, which can take on integer values from 0 to 5. A mood of 5 stands for an excellent mood and 0 for a horrific mood. We could argue that the mood of a student depends on her invested time for the preparation, but for the sake of simplicity, we assume that this is not the case, i.e. that these two features are not affected by multicollinearity. Finally, on the z-axis the dependent variable – the score is shown in percentage. We can further see some data points in figure \ref{figure 2.8}, which represent the exam statistics collected during the previous year. The grey-coloured hyperplane represents the linear regression model that we learned from the data of last year. We assume that this model was determined by applying the Ordinary Least Squares (OLS)\nomenclature{OLS}{Ordinary Least Squares} algorithm on the data and that it equates to the following:
\[Score = 10 + 1,5 \cdot Preparation\ time + 5 \cdot Mood\]
Now that we have our prediction model, we proceed with its interpretation. The first value of 10 is called the intercept, which is the number that the model predicts as expected score, given that our two features (prep-time and mood) take on the value of 0. Applied to our exam scenario, this means that even if a student did not prepare at all for the exam, and simultaneously, the student's mood is awful, i.e. 0, we would still expect the score to be no lower than 10, because $Score = 10 + 1,5 \cdot 0 + 5 \cdot 0 = 10$.
\\The other two values that we have in the equation, of 1,5 and 5 respectively for prep-time and mood, are called coefficients. These represent the weights by how much the given feature influences the target variable. As an example, if the preparation time of a student increased from 10 to 20 hours, we would expect the score to increase by 15\%, ceterus paribus (c.p. – other things equal)\nomenclature{C.p.}{Ceterus Paribus (Other Things Equal)}. If the mood increased from 3 to 5 c.p., we would expect the score to improve by 10\%. We may question ourselves, if the fact that the mood feature has a higher coefficient (weight) than the prep-time feature, automatically means that it is of higher importance. The answer to this question is it depends – if the data would be standardised (i.e. mean zero and standard deviation (STD)\nomenclature{STD}{Standard Deviation} equal to one for all features), then the answer would be yes. However, in our case, the data is not standardised, so this statement does not hold true. The reason why it does not hold is because the mood can only take on integer values from 0 to 5, so its maximal influence on the score could be of 25\%. The prep-time feature, on the other hand, varies between 0 and 50 and can thus have a maximum influence of 75\% on the score, which is three times more than the mood-feature.
\\As we can see, the interpretation of this model is straightforward, once we understand the concept of coefficients and the intercept. If someone questioned us what score we expect for a student that, e.g. prepared for 40 hours and has a mood of 4, we could easily answer that we expect the student to achieve a score of 90 ($Score = 10 + 1,5 \cdot 40 + 5 \cdot 4$).
\\With this example, it becomes clear how the model can simultaneously be used as a prediction model and as an explanation. We thus argue that the model from the example above fulfils the simulatability condition as defined earlier. Moreover, the linear model complies with the counterfactual faithfulness property, with its intercept functioning as reference point \citep{molnar2018}. It must be noted, however, that the intercept is more meaningful in a situation where we previously standardised the data, as then it represents the predicted outcome of an instance when all features are at their mean value \citep{molnar2018}. The selectivity property is not implemented in this form of linear regression, as all features have a corresponding non-zero weight. In sparse models, as computed with, e.g. the Lasso method, this shortcoming can be overcome \citep{molnar2018}.

As a closing consideration for the presented example, we stress that the interpretability of these models can be good, but it highly depends on the complexity and extent of the data. If we, e.g. have over 100 features, with interactions between some of them, then it may become complicated to interpret even these models. Moreover, we made the implicit assumption in the example that the prediction model represented by the hyperplane in figure \ref{figure 2.8} is a good approximation for the data. This means that we assumed that the given model has a high $R^2$ score, which is equal to say that it explains most of the variance existing in the data. If a model has a low $R^2$ score, it makes little sense to interpret it, as it does not explain much of the variance and consequently, the weights are not meaningful \citep{molnar2018}.

Having a good understanding of the linear regression as an exemplary model that is interpretable, we proceed with another, more succinct example involving a decision tree, in the upcoming subsection.

%% ===========================
\subsubsection{Decision Tree}
\label{ch:Content1:sec:Section3:Subsection2:Subsubsection2}
%% ===========================
Decision trees are like linear models, well-known and preferred in various fields due to their simplicity, computational speed, and ease of representation and interpretation \citep{friedman2009}. This method shines where the linear regression gets to its limits – it can model non-linear relationships and interactions between features \citep{molnar2018}. On the other hand, it does not return monotone models, which may occasionally make the interpretability a little more challenging. Decision trees work by partitioning the feature space into a set of smaller rectangular subspaces and then fitting a simple model, such as a constant, in each of these \citep{friedman2009}. How these partitions are determined depends on the used algorithm. As examples of commonly-used approaches, we can mention the Classification and Regression Tree (CART)\nomenclature{CART}{Classification and Regression Tree}, and the C4.5 and its successor the C5.0, as described in \cite{friedman2009} and \cite{quinlan1996improved}.

In our example, again, the emphasis is on interpretability and not on the generation of the decision tree itself. We, therefore, limit ourselves to explaining how to interpret it and how to use it to predict new instances. For further details on the computation of a decision tree, we recommend the book by \cite{friedman2009}.
\\The situation in our example is the following: we want to process mortgage applications for a certain bank, based on two features – annual income (in thousands of a monetary unit) and deposit size (as a percentage of the total house price). Fortunately, we have some records of prior mortgage applications shown in figure \ref{figure 2.9}, on the left. The dark-grey circles show people that successfully applied for a mortgage, and the empty circles show unsuccessful applications. Based on this data, we generate the decision tree that is in the figure on the right. This tree is straightforward to interpret: we start at the top, at the so-called root-node, and work our way down through the tree, answering the questions until we reach one of the four bottom nodes, also called leaves.
%%%%%%%%%%%%% Figure %%%%%%%%%%%%%
\begin{wrapfigure}{l}{0.66\textwidth}
  \vspace{-20pt}
  \begin{center}
    \includegraphics[width=0.66\textwidth]{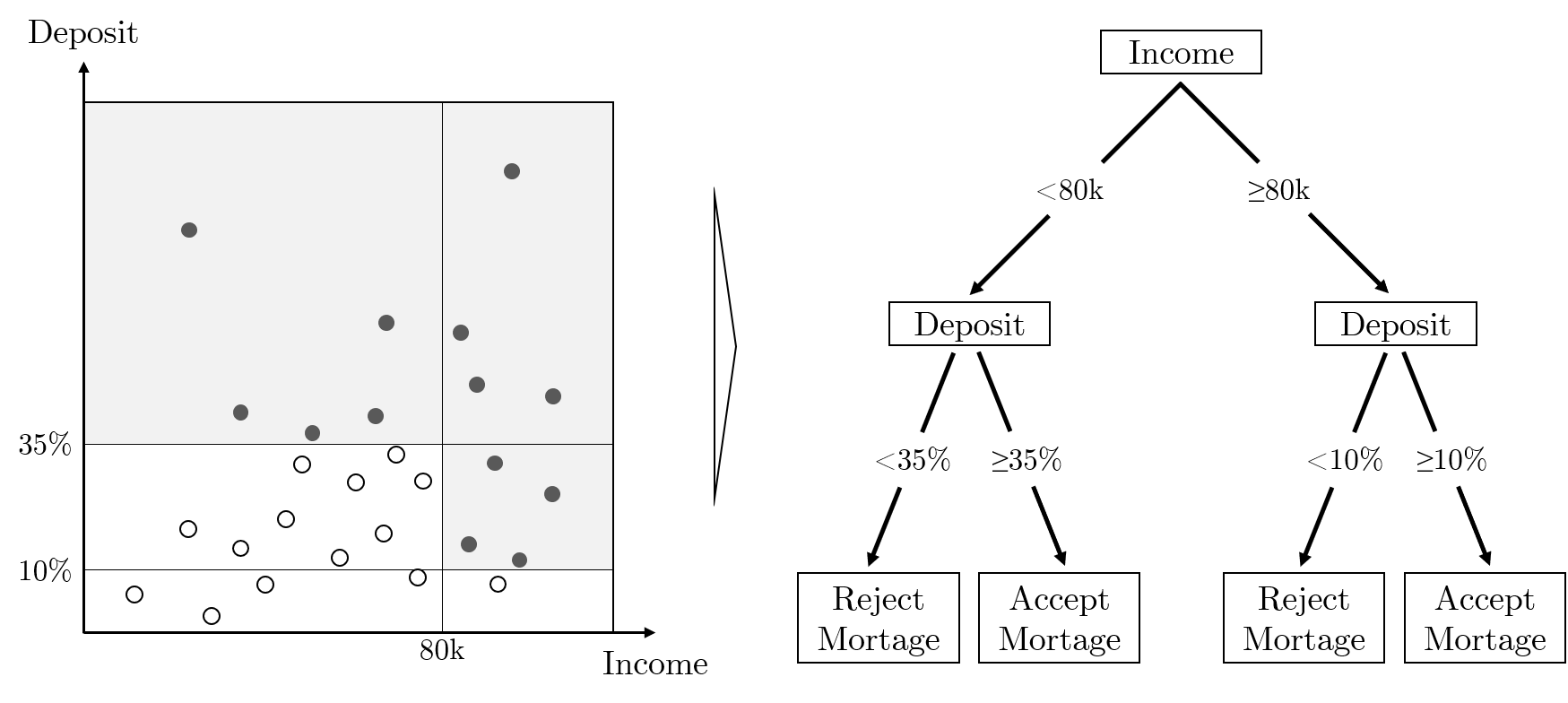}
  \end{center}
  \vspace{-20pt}
  \caption{Interpretable Model 2: Decision Tree}
  \label{figure 2.9}
  \vspace{-10pt}
\end{wrapfigure}
%%%%%%%%%%%%% Figure %%%%%%%%%%%%%
The leaf-node tells us what the result of our prediction is. We could e.g. have the following path: A person applies for a mortgage that earns less than 80k per year, so we go to the left branch of the tree. Fortunately, the house that this person wishes to buy is not very expensive, and thus the person can put down 40\% of the house price as a deposit. This means that the mortgage size would only be 60\% of the house price, and thus this person would be eligible for a mortgage. In the example, there are four possible paths, of which two lead to rejection and the other two to a successful mortgage application. The two options regarding minimum conditions for a person to be eligible for a mortgage are the following. First, the person earns 80k or more per annum (p.a.)\nomenclature{P.a.}{Per Annum (Per Year)} and puts down a deposit equal to 10\% or more of the house price. Second, the person earns less than 80k p.a. and puts down a deposit equal to 35\% or more of the house price. In figure \ref{figure 2.9} on the left, we can see the grey-shaded rectangles, which correspond to the areas where people are eligible for a mortgage. In this simplified example, there is no additional division at, e.g. an income of 30k, which would make sense in reality, as a bank would not issue a mortgage to someone without any income. This would likely only happen if the person had big savings, but then this person would probably not need a mortgage.
\\As before in the linear regression model, the interpretation of the decision tree also becomes intuitive, once we understand how to navigate the tree. If someone at the bank now questioned us, if we would expect a person with, e.g. an income of over 80k p.a. and a down payment of 20\% of the house price to be eligible for a mortgage, we could easily answer positively, as the person fulfils one of the two minimum conditions from above. With this example, it becomes clear how to use a decision tree simultaneously as a prediction model and as an explanation. As long as a tree does not have too many child leaves and long paths, we argue that in cases such as the one above, the tree fulfils the simulatability condition as previously defined.
\\Trees, in general, create good explanations concerning human understandability, as they invite us to think in a counterfactual way: "If the person would put a deposit of 35\% down rather than only 30\% the mortgage would be accepted" \citep{molnar2018}. Trees are also selective, as they prioritise the most important features that yield the highest information gain, and thus they often do not use all available data \citep{molnar2018}. After considering some strengths of trees, we also mention some weaknesses, such as the handling of linear relationships, lack of smoothness and stability in certain situations \citep{molnar2018}. Nevertheless, decision trees with all their relatives and successors, such as ensembles, represent powerful and often still interpretable models.

After discussing in which situations certain model classes can be both – prediction models and explanations simultaneously, we proceed with looking at situations where this is not the case. In these scenarios, we have to rely on explanation methods. We start by recapitulating what an EM is and then discuss the difference between model-dependent and model-agnostic explanation methods, giving examples for each of these.

%% ===========================
\subsection{Explanation Methods}
\label{ch:Content1:sec:Section3:Subsection3}
%% ===========================
We mainly use explanation methods in situations where we have no algorithmic or global model transparency, to interpret predictions of individual instances. We recall the definition given earlier – explanation methods are used to generate explanations to establish post-hoc interpretability. Post-hoc means \textit{after the event}, i.e. we want to establish the interpretability of a model after it has been trained and used to compute the predictions. There are mainly two different types of EMs: model-dependent and model-agnostic (also know as model-independent).
\\Model-dependent explanations are tied to a certain model class, as we see in the following subsection. We can e.g. have an explanation method for boosting (XGB), another one for bagging (random forests) and yet another one for deep neural networks (MLP). These methods build on the specific architecture of the underlying prediction model, which can be advantageous, as model-dependent methods usually have shorter computation times \citep{beillevaire2016}. On the other hand, these methods have two severe weaknesses, which can make them unusable for certain situations. First, as already mentioned and as their name indicates, these do not generalise to other model types. Second, we cannot compare explanations generated by different model-dependent EMs as these most likely do not follow the same standard and logic. Nevertheless, if we only use one type of model for a certain task, it may be a good idea to rely on one of these methods.
\\Model-agnostic explanation methods are independent of a particular model class and can thus, in theory, be applied to any ML algorithm. The weaknesses of model-dependent EMs are the strengths of model-agnostic EMs and vice-versa. This means that model-agnostic EMs often have longer computation times, but we can, however, theoretically apply them to all ML models, and they produce explanations that are comparable for different ML models \citep{molnar2018,beillevaire2016,ribeiro2016,lundberg2017}. If we are working on a scenario where we must use several different ML algorithms, it may thus be more feasible to use a model-agnostic EM.

An alternative to using explanation methods to establish interpretability is only to use interpretable models, such as the ones presented in the previous two sections \citep{molnar2018}. This could, however, bring a significant disadvantage, namely reduced accuracy compared to more complex ML models \citep{molnar2018}. Having a good understanding of explanation methods in general, we continue with looking at model-dependent explanation methods in the next subsection.

%% ===========================
\subsubsection{Model-Dependent Explanation Methods}
\label{ch:Content1:sec:Section3:Subsection3:subsubsection1}
%% ===========================
As described above, model-dependent EMs only work on a specific model class, but usually they do so very successfully in terms of computational speeds. There are numerous approaches in this category, with name-worthy examples such as DeepLift by \cite{shrikumar2017learning} and Layer-Wise Relevance Propagation by \cite{binder2016layer} for deep neural networks; the Random Forest Interpreter by \cite{palczewska2014}; and the Tree Ensemble Interpreter by \cite{lundberg2017consistent}, amongst various others. To gain a better understanding and insight on how these model-dependent EMs work, we pick one of the above-mentioned approaches and discuss it in greater detail. As we use deep neural networks in our experiments chapter to validate our Explanation Consistency Framework, we choose to elaborate on the basic principles of the DeepLift EM in the following.

DeepLift stands for "Deep Learning Important Features", and it is a method for decomposing the output prediction of a neural network and attributing importances to the used features in the prediction \citep{shrikumar2017learning}. To achieve this, the method makes use of the layered architecture and the feed-forward / propagate backwards mechanisms of a neural network. As noted by \cite{shrikumar2017learning}, the central principle consists in explaining the difference in output of a particular instance from the reference output, regarding the difference of the input of this instance from the reference input. A strongly simplified explanation of how this difference is computed is given in the following.
\\First, we must compute the reference activation for all neurons, by propagating the reference input through the network. Once we know what the reference activation of the network looks like, we then propagate the instance that we wish to explain through the network and compute the differences in all neurons between the reference activation and the current activation. We thereby sum all effects, and finally, express the contributions on a feature-level.
\\The challenge is to choose an appropriate reference, which depends on the problem specifics and it should be some default or neutral input \citep{shrikumar2017learning}. In practice, this usually comes down to domain-specific knowledge, and often, the contributions are calculated with multiple distinct reference inputs, averaging the results over these \citep{shrikumar2017learning}.

It becomes clear from the brief explanation above, why the DeepLift method is only applicable to neural networks – it depends on its layered architecture and mechanisms, to calculate the feature importances. Even though there is currently ongoing research in model-dependent EMs, we only focus on model-agnostic EMs in our experiments chapter. Nevertheless, the developed ECF framework in this research is also applicable to model-dependent EMs. In the following, we continue with discussing model-agnostic EMs.

%% ===========================
\subsubsection{Model-Agnostic Explanation Methods}
\label{ch:Content1:sec:Section3:Subsection3:subsubsection2}
%% ===========================
From before, we already know that in theory, model-agnostic EMs are applicable to any ML classifier \citep{ribeiro2016,lundberg2017}. Agnostic in this context means to separate the explanation method from a specific machine learning model. \cite{ribeiro2016} notes that model-agnostic explanation methods should feature the following desirable characteristics: model flexibility, explanation flexibility and representation flexibility. We have already discussed the former one – the EM should work for all types of ML models. Explanation flexibility is related to how we present the explanation to the user. This could be a visual explanation, such as a chart with feature importances, but it could also be a linear formula in a different situation or a simple decision tree in yet another case. For each situation there are probably more and less intuitive ways to transmit an explanation to a human, we should thus have the flexibility to choose the most suitable one. Finally, representation flexibility means that the EM should not have to use the same feature representation as the ML model that is being explained \citep{molnar2018}. As an example, in text classification, often the abstract concept of word embeddings are used, which may not be very intuitive to interpret for a human. In this situation, it may thus be a better choice to present the user with single words as an explanation, rather than some probably less intuitive word vectors \citep{molnar2018}.

In this sub-chapter, we mainly focus on two of these model-agnostic EMs, namely on the already-mentioned Local Interpretable Model-Agnostic Explanations (LIME)\citep{ribeiro2016} and on Shapley Additive Explanations (SHAP)\nomenclature{SHAP}{Shapley Additive Explanations}\citep{lundberg2017}. These two methods are also the ones that we use in our experiments chapter, to validate our Explanation Consistency Framework. LIME and SHAP take two completely different approaches to explain predictions, as we see in the following. There are also other model-agnostic EMs, such as the Descriptive Machine Learning Explanations (DALEX)\nomenclature{DALEX}{Descriptive Machine Learning Explanations}\footnote{\url {https://github.com/pbiecek/DALEX}}, which are, however, beyond the scope of this thesis. In the following, we proceed with discussing LIME and SHAP, by giving an overview of the fundamental principles and ideas on how these two methods work.

%% ===========================
\paragraph{Local Interpretable Model-Agnostic Explanations (LIME)}
\label{ch:Content1:sec:Section3:Subsection3:subsubsection2:Paragraph1}
%% ===========================
LIME is an explanation technique that explains the predictions made by any ML model faithfully and transparently, by locally learning an interpretable model around the prediction \citep{ribeiro2016}. LIME cannot only be applied to tabular data (i.e. data in a table format) but also to textual and image data. The internal processes of explanation generation for these different data types are slightly different, but the main idea, which we explain in the following, remains the same.
\\The rationale behind the process through which LIME computes an explanation is quite intuitive. Based on the simplified explanation given by \cite{molnar2018} there are mainly five steps involved. (1) Instance of interest: the user selects an instance, for which an explanation of its BB prediction is needed. (2) Perturbation and prediction: LIME slightly perturbs the given training dataset, thereby generating new artificial data points, and obtains their corresponding predictions from the BB model. With this step, LIME fundamentally tests what happens to the predictions of the ML model, when we feed it variations of the original data. (3) Weighting: the new artificial data points are weighted based on their distance to our instance of interest (the closer these are to our instance, the higher is their attributed weight). (4) Fit an interpretable model: based on the newly created and weighted artificial dataset, train a new interpretable model, such as a linear regression or a decision tree. (5) Explanation: based on the newly trained interpretable model explain the instance of interest.
\\To deepen our understanding of this process, we explain it based on an example in the following. Imagine that we have a classifier that predicts whether certain credit card transactions are fraud or no fraud. Let us further assume that this classifier is highly accurate and consequently very complex and opaque. The classifier is represented in figure \ref{figure 2.10}, with
%%%%%%%%%%%%% Figure %%%%%%%%%%%%%
\begin{wrapfigure}{r}{0.50\textwidth}
  \vspace{-20pt}
  \begin{center}
    \includegraphics[width=0.50\textwidth]{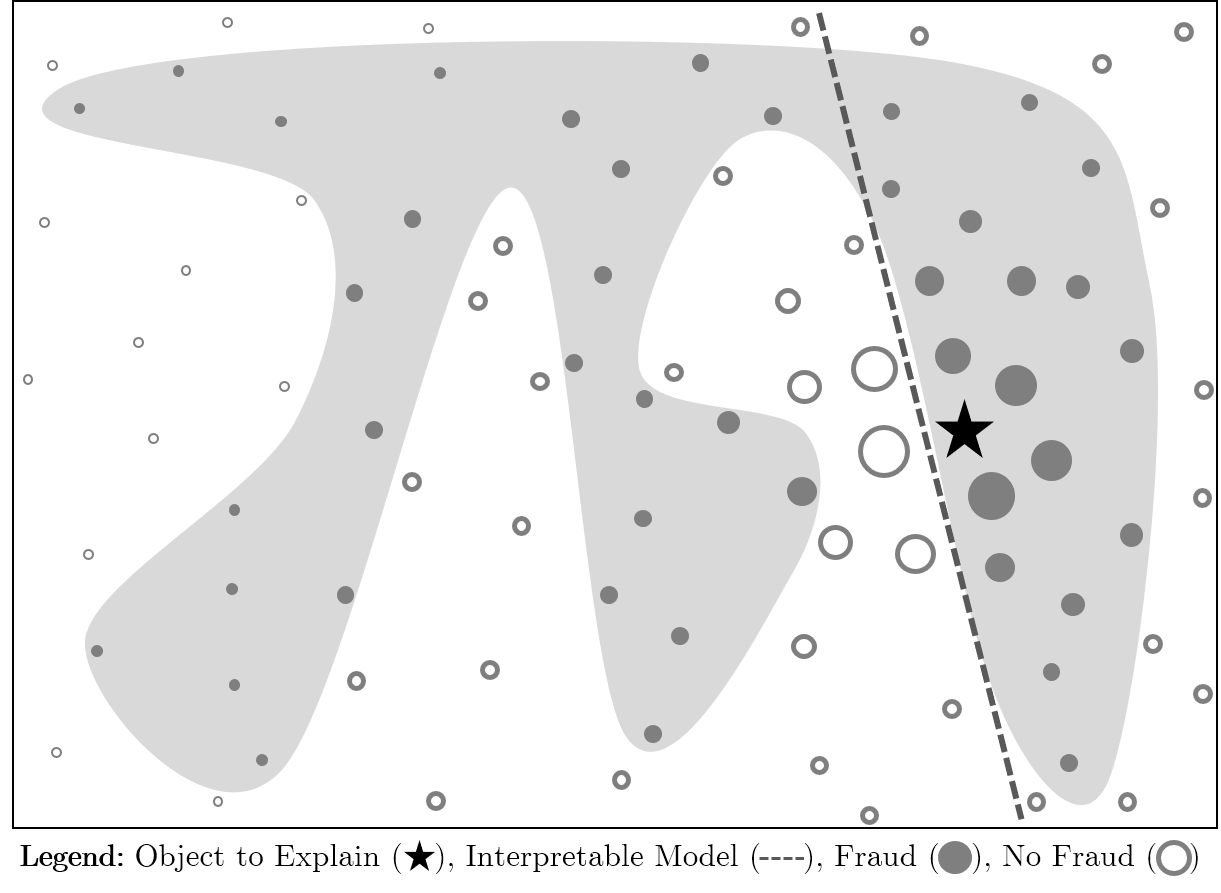}
  \end{center}
  \vspace{-20pt}
  \caption{LIME: Fraud Example [based on \cite{ribeiro2016}]}
  \label{figure 2.10}
  \vspace{-10pt}
\end{wrapfigure}
%%%%%%%%%%%%% Figure %%%%%%%%%%%%%
the grey-shaded area corresponding to the \textit{fraud} class (dark-grey-filled circles) and the white-shaded area corresponding to the \textit{no fraud} class (empty circles). Concerning the process described above, we would now first choose the instance of interest that we wish to explain, which corresponds to the dark-shaded pentagram. In the next step, LIME would generate all perturbed instances as visible in the diagram, and use the classifier to predict their label classes (fraud vs no fraud). In the following, LIME would attribute a higher importance (weight) to those artificially created instances, which are the closest to our instance of interest. In figure \ref{figure 2.10}, we can see that the perturbed data points around the pentagram are of greater size (have greater weight). Moreover, the further away from our instance of interest, the smaller (less important) they get. Once all of the artificial instances have an attributed weight, LIME proceeds with fitting an interpretable model, such as a linear regression in our case (represented by the dashed line). In the shown example, the regression line seems to be a good approximation of that local area around the instance of interest, as it mostly captures the local decision boundary of the BB classifier. This is also called local fidelity, i.e. the model is locally faithful \citep{molnar2018}.

Naturally, many further processes and details are going on behind the scenes of LIME. For the sake of simplicity and general understanding of the principle, we however abstracted from further details. To explore the formal definition of LIME, see section \ref{Appendix:Section:A} of the appendix. In the experiments section, we revisit LIME and test it with our Explanation Consistency Framework. This allows us to draw conclusions about its strengths, weaknesses and best application domains. Having a good understanding of the LIME method, we proceed with exploring another model-agnostic EM, the Shapley Additive Explanations, in the upcoming subsection.

%% ===========================
\paragraph{Shapley Additive Explanations (SHAP)}
\label{ch:Content1:sec:Section3:Subsection3:subsubsection2:Paragraph2}
%% ===========================
In comparison to LIME, Shapley Additive Explanations take a completely different approach to explain predictions by opaque ML models, namely cooperative game theory. Shapley values were introduced and named after Lloyd Shapley, who published the concept in 1953 \citep{shapley1953,cubitt1991}. The main idea of Shapley values is to fairly distribute gains and costs to several players cooperating in a coalition. The concept mainly applies to cases where contributions of players are unequal, ensuring that each player only gains as much or more than they would have gained, if they acted independently \citep{shapley1953}. This is a crucial aspect, as players would otherwise not be incentivized to collaborate with each other.

Shapley values have a broad range of applications, and we can also use these in situations, where we wish to explain the behaviour of a BB machine learning model. This requires to formulate the explanation problem as a cooperative game, as we see in the following with an example based on \cite{molnar2018}. Let us imagine that we have a complex ML model that forecasts house prices for a certain area, based on three features: size, location and condition. These three features are the players of our cooperative game and the predicted price by our ML model is the payout. Our goal is to fairly distribute this payout amongst the three players, based on their contributions towards it. As an example, let us say that the average house price is 500k and we want to explain the prediction of a certain house that is estimated to cost 600k, based on the three mentioned features. An explanation for this prediction could look like the following: the size of 350 m$^2$ compared to the average size of 300 m$^2$, raised the price by 40k; the exquisite location with a rating of 10/10 compared to an average 5/10 further increased the price by 80k; lastly, the condition of the house, with 4/10, below the average of 6/10 decreased the price by 20k. If we sum up all of these contributions, we get 100k, which is exactly the difference between the average predicted house price and the house price that we are currently looking at. The values of 40k, 80k and -20k are nothing else than the Shapley values for our features, respectively, size, location and condition. We can, therefore, specify that the Shapley value
%%%%%%%%%%%%% Figure %%%%%%%%%%%%%
\begin{wrapfigure}{l}{0.50\textwidth}
  \vspace{-20pt}
  \begin{center}
    \includegraphics[width=0.50\textwidth]{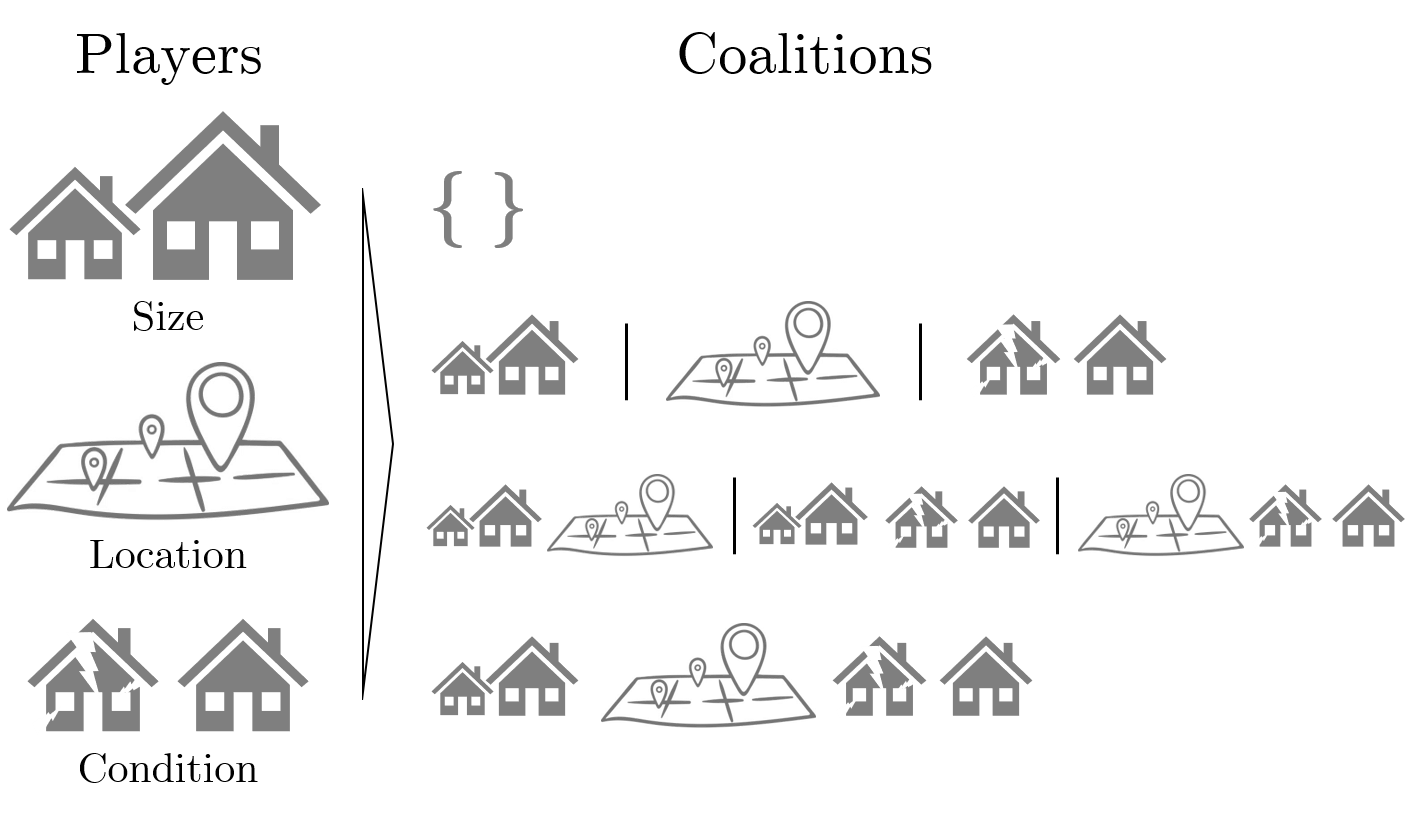}
  \end{center}
  \vspace{-20pt}
  \caption{SHAP: House Prices Example [based on \cite{molnar2018}]}
  \label{figure 2.11}
  \vspace{-10pt}
\end{wrapfigure}
%%%%%%%%%%%%% Figure %%%%%%%%%%%%%
is the average marginal contribution of a feature value (e.g. 350 m$^2$ for the size-feature) to the prediction minus the average prediction, over all possible coalitions \citep{shapley1953}.
\\To explain the computation of Shapley values, let us imagine that there is another feature for our houses – wheelchair-access (yes or no). In figure \ref{figure 2.11}, we represent all possible coalitions over which we must compute the marginal contributions of our wheelchair-feature, to compute its Shapley value. To compute the marginal contribution for one coalition we proceed as follows: we assess the contribution of the "wheelchair-access: yes" feature value, when added to the coalition of "location: 10/10", "condition: 4/10" and a randomly drawn value for the size-feature (320 m$^2$). Then we use our BB model to predict the price of this house, which we estimate at 550k. In the next step, we remove the "wheelchair-access: yes" feature from our coalition and replace it with a randomly drawn value for that feature ("wheelchair-access: no"). Note that the value could have also been yes again. Now, we compute the price for this new coalition, which we estimate at 530k. The contribution of the "wheelchair-access: yes" feature is, therefore, $550k - 530k = 20k$. This value depends on the sampled feature values, and these estimates of marginal contributions get better if we repeat this procedure multiple times \citep{molnar2018}. Also, the 20k only represent the contribution of the "wheelchair-access: yes" feature in this coalition. To calculate the Shapley value of this feature, we would have to repeat the procedure of calculating the marginal contribution over all coalitions, as shown in figure \ref{figure 2.11}, and take the (weighted) average out of these \citep{shapley1953}.

As we can see, the computation of Shapley values is a non-trivial task and very costly in computational terms, as the time complexity increases exponentially with the number of features \citep{molnar2018}. Consequently in practice, often only approximations of the Shapley values are viable and used. The SHAP package used in the experiments chapter, e.g. also computes approximative Shapley values for deep neural networks \citep{lundberg2017}. For further details and the formal definition of the Shapley values, see section \ref{Appendix:Section:B} of the appendix.

To conclude this section, we lastly highlight three desirable properties for this context of feature attributions that the Shapley values satisfy based on \cite{shapley1953,molnar2018}. (1) Dummy: if a feature (player) never changes the predicted value (payout), no matter to which coalition it is added, then it should receive a Shapley value of 0. (2) Symmetry: if two features always add the same marginal value to any coalition to which they are added, then their contribution should be equal. (3) Efficiency: The sum of all Shapley values (of all features) has to sum up to the difference between the predicted value and the average value. These three axioms are provably only satisfiable by the Shapley values and no other attribution method \citep{molnar2018,lundberg2017}. Consequently, we expect that the SHAP method outperforms LIME regarding explanation quality, as later defined by our Explanation Consistency Framework.

With a good understanding of the two post-hoc explanation methods – LIME and SHAP, which are highly relevant for the experiments chapter, we proceed to the next topic of related work. With the related work section, we strive to give a brief overview of other approaches that researchers have taken, to assess the quality of different explanation methods and their corresponding explanations.

%% ===========================
\section{Related Work}
\label{ch:Content1:sec:Section4}
%% ===========================
In sections \ref{ch:Content1:sec:Section2} and \ref{ch:Content1:sec:Section3} of the preliminaries, we have already discussed literature related to the role of interpretability in ML, and to how we can establish interpretability by means of explanation methods. In this last section of the preliminaries, we further explore literature, where other researchers specifically address the topic of analysing and comparing the quality of different explanation methods and their generated explanations.
\\Even though the field of interpretable ML has only gained a significant amount of attention in the last few years, there seems to currently be substantial research on developing new explanation methods. However, there seems to be very little research on developing measures and approaches that allow to compare the quality of already existing explanation methods.

We start with a brief overview of qualitative indicators that other researchers mention in literature and that enable the assessment of explanation quality. \cite{doshi2017_rigorous_science} mainly mention the following five indicators: form, number, level of compositionality, monotonicity and uncertainty.
\\Form relates to the basic units of explanation, i.e. what are the explanations composed of \citep{doshi2017_rigorous_science}. These could e.g. be based on feature importances of the inputs, or on new combined features that were derived in the prediction process, or even on something completely different, such as a collection of pixels for images.
\\Number refers to the amount of explanation units that the explanation uses, i.e. if it is based on input features, does it return all features with their respective importances or only a few most important ones \citep{doshi2017_rigorous_science}.
\\The compositionality indicator assesses the organisation and structure of the explanation \citep{doshi2017_rigorous_science}. This translates, e.g. into hierarchical rules such as "show features with the highest importances first", or "cut feature importances with less than a 5\% impact". These types of rules and abstractions can help humans to process explanations faster.
\\Monotonicity is related to interactions that may occur between different units of explanation \citep{doshi2017_rigorous_science}. These could e.g. be combined in linear, non-linear, monotone or other ways. Some of these types of relations may thus be more intuitive to understand for humans than others.
\\Finally, uncertainty refers to the possibility that the explanation method returns a human-understandable uncertainty measure \citep{doshi2017_rigorous_science}. If the EM, e.g. returns an explanation with an uncertainty of 10\%, or with a confidence interval, the question is if users can interpret and understand it correctly. In case of a good explanation with an intuitive uncertainty measure, this question should be positively answered.
\\All of the above-mentioned indicators mainly impact the two interpretability goals of efficiency and understandability \citep{ruping2006}, as described in section \ref{ch:Content1:sec:Section2:Subsection3:Subsubsection1}. As an example, compositionality, i.e. hierarchically organised and structured explanations are much easier to grasp for humans, thereby augmenting understandability. At the same time, structured information can also be processed faster in our minds, thereby increasing the efficiency of an explanation with compositionality.

Having discussed some qualitative indicators that can be used to assess the quality of explanations, we proceed to review some quantitative ones in the following. For this research, the quantitative indicators are of greater relevance, as the foundation of our Explanation Consistency Framework are axioms, which we verify quantitatively.
\\The review of quantitative indicators is mainly based on the paper by \cite{sundararajan2017}, who choose a similar axiom-based approach. Their research is, however, not to compare explanation qualities, but to develop a new explanation method for deep networks called \textit{integrated gradients}. Nevertheless, there are still some parallels between their research and ours, which we explore. 
\\The problem that \cite{sundararajan2017} study is how to attribute a prediction of a deep network to the original input features. As a solution to their problem, they propose two axioms that explanation methods for deep networks should fulfil, namely \textit{sensitivity} and \textit{implementation invariance} \citep{sundararajan2017}.
\\The sensitivity axiom relates to feature importances on an individual feature-level and is composed of two parts. In the first part, \cite{sundararajan2017} propose that if we have two different predictions for two inputs that only differ in one particular feature, then this feature should have a non-zero attribution. This seems logical because if only one feature between the two inputs is different, and if that difference was enough to change the prediction, then clearly that particular feature had an influence on the outcome. It should, therefore, receive a feature importance greater than zero. The second part of the sensitivity axiom is identical to the dummy property of the Shapley values, discussed earlier in section \ref{ch:Content1:sec:Section3:Subsection3:subsubsection2:Paragraph2}. It states that if the deep network never depends on some input for its predictions, then the attribution or feature importance of that input should always be zero \citep{sundararajan2017}.
\\The second axiom on implementation invariance argues that, if two networks are equal, i.e. they return identical predictions to the same inputs, then the corresponding attributions of these two networks must also be identical, even if these have different implementations \citep{sundararajan2017}. This is crucial because if an attribution method does not satisfy this axiom, then it is potentially sensitive to irrelevant aspects of the models, such as its architecture, which is undesirable.
\\The authors defined the two axioms above as desirable characteristics for their model-dependent explanation method for deep networks \citep{sundararajan2017}. We argue, however, that those two axioms cannot only be the basis of such an EM. They can also be part of a larger framework of axioms, utilised to assess and compare the explanation quality across different explanation methods. This larger framework could include the two proposed axioms modified in such way that these are applicable in a model-agnostic context. Therefore, in the next chapter, we establish a connection between the axioms stated above with the axioms proposed in this research.

With the end of this section, and a broad overview of recent and related work in the field of axiomatic explanation quality assessment, and more generally interpretable ML, we conclude the preliminaries chapter. In the next chapter, we introduce the Explanation Consistency Framework, which constitutes the main contribution of this research.

%% ===========================
%% ===========================
\chapter{Axiomatic Explanation Consistency}
\label{ch:Content3}
%% ===========================
%% ===========================

\hspace{\parindent}In this chapter, we introduce the axiomatic explanation consistency framework, which is a functionally-grounded evaluation method as earlier defined by \cite{doshi2017_rigorous_science}. It consists of three proxies for explanation quality, which we, however, call axioms \citep{sundararajan2017} from here onwards. Moreover, we can apply it to all post-hoc interpretability explanation methods, which work with feature importance values in absolute or relative terms. The framework works for both – regression and classification, whereas each case has a separate definition.
\\The naming of the framework consists of two parts: (1) axiomatic, and (2) explanation consistency. The meaning of the first part is straightforward – as the basis of the framework consists of axioms, it is an axiomatic framework. The second part of explanation consistency represents the objective that an explanation method should attain and that we wish to measure. This idea is reinforced by \cite{lundberg2017}, as they mention that good model explanations should be consistent with humans explanations, i.e. they should be aligned with and based on human intuition. We incur that the defined axioms in this research ground on human intuition.

The axioms defined hereafter, always relate an object to its corresponding explanation. By object, we mean an individual row in a tabular dataset (i.e. a single data point) with its corresponding features (i.e. attributes) and prediction by the ML model. By explanation we mean the corresponding feature importances for all features of that object. With that, we define our three axioms for explanation consistency as follows:

\textit{1. Identity: Identical objects must have identical explanations}

\textit{2. Separability: Non-identical objects can not have identical explanations}

\textit{3. Stability: Similar objects must have similar explanations}

\noindent We explain the identity axiom as follows: if we prompt the explanation method multiple times to explain the same object with its corresponding prediction, we expect it to generate the same explanation multiple times. If this is not the case, then some form of randomisation must be involved in the explanation generation process, which is undesired. Let us imagine that we prompt our EM to explain why a certain house costs 20k more than the average house. The first time we prompt the EM, it tells us "the above-average size increased the value of the house by 10k, and the existence of a pool by another 10k". The second time we prompt the EM for the same house, it tells us "the excellent location increased the value of the house by 15k and the existence of a garage by another 5k". For the user of the explanation method, this would be highly confusing and inconsistent, as it would not be possible to understand why exactly the price is 20k above average. Moreover, an EM that does not return identical explanations for identical objects cannot possibly be accurate, due to the random element involved in the explanation.

\noindent The separability axiom is somewhat more difficult to illustrate, however, through examples it should also become evident. We start with the more straightforward example of regression. Let us imagine that we have two houses, of which one costs 500k and the other costs 450k. There must be some differences between these two houses; otherwise, our ML model would have predicted the same price for both houses. If the houses are different, then the explanations why one costs 500k and the other 450k must be different. When we say different, we mean that the EM could still use, e.g. size and location as features to explain both prices, however, the explanation could not be precisely the same regarding feature importances or weights for each of the features. For the 500k house the size could e.g. have contributed 80k and location 40k, whereas, for the 450k house, the size could have only contributed 50k and the location 10k.
\\Now let us imagine a new situation, with two houses both costing 400k. These two houses do have some features in common; however, other features are also different. Nevertheless, the ML model estimated both houses to be worth 400k. Because there are still some differences in the attributes of these two houses, the explanation why one costs 400k and why the other costs 400k would still have to be different concerning feature importances.
\\We can also apply this to a classification example, where we have a ML model that classifies credit card transactions into fraud or no fraud. If we have two different transactions, the prediction model can classify both as fraud; however, the explanation for why one is fraud and why the other is also fraud must be different regarding feature importances.
\\For this axiom to work, there is, however, an important assumption that we must take regarding the ML model. This axiom only holds if the model architecture does not have more degrees of freedom (DOF)\nomenclature{DOF}{Degrees of Freedom} than needed to represent the prediction function \citep{sundararajan2017}. If this is not the case, then there could be distinct configurations of the model architecture that lead to the same prediction function.
\\Applied to our second axiom this means that a situation could occur in which we, e.g. have two different transactions that are both predicted to be fraud, which have the same explanation. This could happen because if the two transactions would only differ in one feature, which is, however, not relevant for the prediction function, then the explanation method could potentially still return the same explanation for these two distinct objects. This would be precisely the opposite of what we wish to achieve with the second axiom.
\\As an example, let us imagine that we have two transactions that are the same in all features, but there is a weather-feature, which in the first transaction says "rain" and in the second transactions says "sun". The weather does not influence a transaction to be related to fraud or no fraud. The ML model thus still predicts fraud for both these transactions. Now, because the ML model has more DOF than it needs (because of the unnecessary weather-feature), our explanation method would still render the same explanation for these two transactions. This is the correct behaviour because if we leave out the irrelevant weather-feature, we have the same object twice, and then the EM should render the same explanation, as established in the first axiom.
\\In sum, it is thus crucial that the DOF assumption is not violated, i.e. the ML model does not have more DOF than necessary to represent the prediction function.

\noindent Finally, the third axiom of stability was inspired on the algorithmic stability concept. An algorithm is said to be stable if slight perturbations in the training data only result in small changes in the predictions of the algorithm \citep{bonnans2013}. Similarly, we define an explanation method to be stable if it returns similar explanations for slightly different objects. It is thereby desirable that the higher the difference is between two objects, the more significant the difference should be between their corresponding explanations, and vice-versa. To return to our housing example, the explanations for two houses, one costing double the amount of the other, should be substantially different (as the houses are likely entirely different regarding features). However, if two houses are very similar regarding their attributes, and thus, the predicted price is also very similar, e.g. 500k and 501k, then the explanation concerning feature importances should also be similar.

With our three axioms defined, we proceed to discuss some more characteristics of the Explanation Consistency Framework. In chapter \ref{ch:Content1:sec:Section2:Subsection3:Subsubsection1}, we established that interpretability comprises the three goals of accuracy, understandability and efficiency. We argue that the above-described axioms influence each of these goal-dimensions of interpretability and that only by at least partially fulfilling these axioms, explanations of an EM can be interpretable to users (figure \ref{figure 3.1}). On the other hand, fulfilling these axioms is not necessarily sufficient to establish interpretability. Hence, we cannot guarantee that if an EM fulfils the above axioms that all users can understand its generated explanations.
%%%%%%%%%%%%% Figure %%%%%%%%%%%%%
\begin{figure}[htp]
  \begin{center}
    \includegraphics[width=1.00\textwidth]{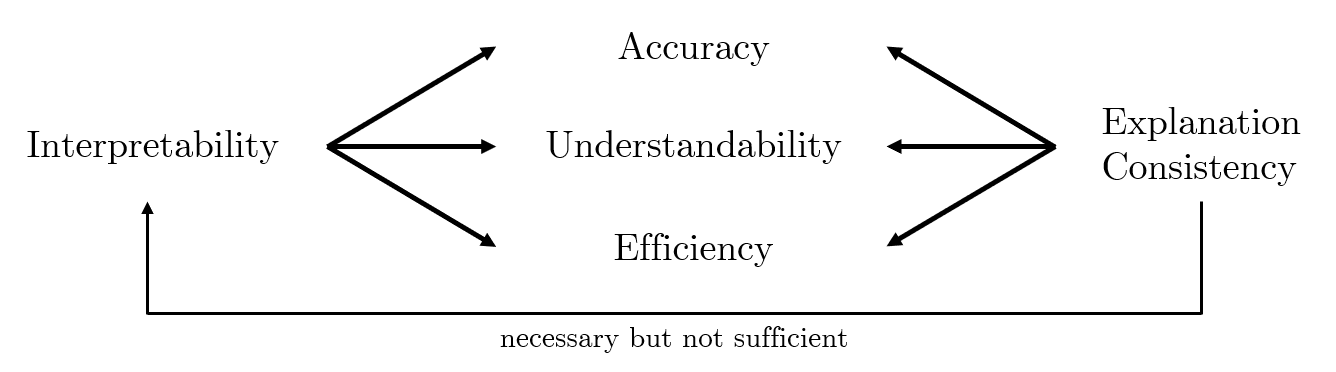}
  \end{center}
  \vspace{-20pt}
  \caption{Explanation Consistency: a Prerequisite for Interpretability}
  \label{figure 3.1}
  \vspace{-10pt}
\end{figure}
%%%%%%%%%%%%% Figure %%%%%%%%%%%%%
The main argument for that is that interpretability remains a partly subjective concept. The axioms are, therefore, necessary but not sufficient to achieve interpretability with an EM.
\\Also, as we see in the experiments chapter, evaluating different explanation methods with the axioms mentioned above makes more sense if we use these as "soft" rather than "hard" criteria. We would thereby not say that a certain EM generally violates axiom 2 if it did so in, e.g. ten out of twenty-thousand cases. It is more meaningful to verify these axioms regarding the degree or percentage of fulfilment. So, a certain EM could, e.g. comply with axiom 1 in 100\% of the cases, with axiom 2 in 95\% of the cases, and with axiom 3 in 90\% of the cases.

Having a good understanding of the explanation consistency framework, and of the axioms, we proceed to the next sections, where we formally define each of these axioms. We start with the regression case and after that, cover the classification case.

%% ===========================
\section{Regression Case}
\label{ch:Content3:sec:Section1}
%% ===========================
Above, we have already defined the axioms for explanation consistency in natural language. In this section, we translate these axioms into mathematical expressions for the regression case, i.e. when the target variable assumes any real value. We start by introducing the necessary notation, and then describe the axioms thereafter.

\noindent {\Large \textbf{Notation:}}
\begin{nscenter} {\textbf{Multisets:}} \end{nscenter}
$X:$ Multiset of all objects \\
$E:$ Multiset of all explanations \\
$\Lambda:$ Multiset of all predictions over $X$ \\
$Z:$ Multiset of all objects with corresponding predictions $\lambda_i$ \\
$P$: Multiset of all Spearman rank correlations between $D_Z$ and $D_E$ \\

\begin{nscenter} {\textbf{Numbers and Arrays:}} \end{nscenter}
$\vec{x_i}:$ Feature vector of the $i$-th object with relevant features \\
$\vec{\varepsilon_i}:$ Explanation vector of the $i$-th object with feature importances for each of $\vec{x_i}$'s features \\
$\lambda_i:$ Prediction for the $i$-th object \\
$D_Z:$ Pairwise distance matrix over $Z$ \\
$D_E:$ Pairwise distance matrix over $E$ \\
$\rho_j:$ Spearman rank correlation for the $j$-th column between $D_Z$ and $D_E$ \\

\begin{nscenter} {\textbf{Functions:}} \end{nscenter}
$d:$ Distance function \\
$p:$ Prediction function, $p:X \longrightarrow \Lambda$ \\
$e:$ Explanation function, $e:Z \longrightarrow E$ \\
$\rho:$ Spearman's rank correlation function, $\rho: D_Z \times D_E \longrightarrow P$ \\

\vspace{-5pt}
\noindent {\Large \textbf{Axioms:}}
\begin{nscenter} {\textbf{Identity:}} \end{nscenter}
\begin{nscenter} $d(\vec{x_a}, \vec{x_b}) = 0 \Longrightarrow d(\vec{\varepsilon_a}, \vec{\varepsilon_b}) = 0,$ \end{nscenter}
\begin{nscenter} $\forall a, b$ \end{nscenter} \\
\vspace{-15pt}
\begin{nscenter} {\textbf{Separability:}} \end{nscenter}
\begin{nscenter} $d(\vec{x_a}, \vec{x_b}) \neq 0 \Longrightarrow d(\vec{\varepsilon_a}, \vec{\varepsilon_b}) > 0,$ \end{nscenter}
\begin{nscenter} $\forall a, b$ \end{nscenter} \\
\vspace{-15pt}
\begin{nscenter} {\textbf{Stability:}} \end{nscenter}
\begin{nscenter} $\rho(D_{Z_j}, D_{E_j}) = \rho_j > 0,$ \end{nscenter}
\begin{nscenter} $\forall j \in |Z|,$ \hspace{5pt} $\rho_j \subset P$ \end{nscenter}

The first axiom is straightforward to interpret: if the distance between two objects is zero, i.e. it is the same object, then the distance between the two corresponding explanations should also be zero, i.e. it should be the same explanation.

The separability axiom specifies the following: if the distance between two objects is not zero, i.e. these are two different objects, then the distance between their two corresponding explanations must be greater than zero, i.e. these must be two distinct explanations. Remember that explanations do not need to use different features to explain the two objects, but different feature weights for some (or at least one) features. To fulfil this axiom, the DOF assumption must not be violated, as mentioned before. In our understanding, the previously discussed axioms of sensitivity and implementation invariance in section \ref{ch:Content1:sec:Section4} by \cite{sundararajan2017}, would ensure the compliance with the DOF assumption. Consequently, we can see these two axioms as an extension or a refinement of the separability axiom, which would guarantee its validity.

Finally, the third axiom of stability makes use of Spearman's rank correlation coefficient. It specifies that if we compute Spearman's rho $\rho_j$, between two corresponding columns of $D_Z$ and $D_E$ that the rank correlation must be greater than zero. The greater $\rho_j$, the better for intuitive interpretation purposes. If $\rho_j$ were equal to 1, it would mean that the two most similar objects in $Z$ also have the two most similar explanations in $E$; the two second-most similar objects in $Z$ also have the two second-most similar explanations in $E$; and so forth. If $\rho_j$ were -1, then the two most similar objects in $Z$ would have the two most dissimilar explanations in $E$, which is precisely the opposite of what we wish to achieve with the third axiom. The value of $\rho_j$ should thus always be greater than 0 for each $\rho_j \in P$, and as mentioned – the closer $\rho_j$ is to 1, the better.

After discussing and understanding the nuts and bolts of the explanation consistency framework for the regression case, we proceed to discuss the classification case hereafter.

%% ===========================
\section{Classification Case}
\label{ch:Content3:sec:Section2}
%% ===========================
The classification case is very similar to the regression case. Actually, only the stability axiom must be reformulated. In classification, the target variable is a class rather than a real number. There can be any finite number of two or more classes. As before, we start with the notation, which features some small changes, and then proceed with the axioms.

\noindent {\Large \textbf{Notation:}}
\begin{nscenter} {\textbf{Multisets:}} \end{nscenter}
$X:$ Multiset of all objects \\
$E:$ Multiset of all explanations \\
$\Lambda:$ Multiset of all predictions over $X$ \\
$Z:$ Multiset of all objects with corresponding predictions $\lambda_i$ \\
$\overline{C}$: Multiset of all explanation clusters with $|\overline{C}| = |\Lambda|$ \\

\begin{nscenter} {\textbf{Numbers and Arrays:}} \end{nscenter}
$\vec{x_i}:$ Feature vector of the $i$-th object with relevant features \\
$\vec{\varepsilon_i}:$ Explanation vector of the $i$-th object with feature importances for each of $\vec{x_i}$'s features \\
$\lambda_i:$ Prediction for the $i$-th object \\
$\overline{c_i}:$ Explanation cluster of the $\lambda_i$-th prediction \\

\begin{nscenter} {\textbf{Functions:}} \end{nscenter}
$d:$ Distance function \\
$p:$ Prediction function, $p:X \longrightarrow \Lambda$ \\
$e:$ Explanation function, $e:Z \longrightarrow E$ \\
$c:$ Clustering function, $c: E \longrightarrow \overline{C}$ \\

\vspace{-5pt}
\noindent {\Large \textbf{Axioms:}}
\begin{nscenter} {\textbf{Identity:}} \end{nscenter}
\begin{nscenter} $d(\vec{x_a}, \vec{x_b}) = 0 \Longrightarrow d(\vec{\varepsilon_a}, \vec{\varepsilon_b}) = 0,$ \end{nscenter}
\begin{nscenter} $\forall a, b$ \end{nscenter} \\
\vspace{-15pt}
\begin{nscenter} {\textbf{Separability:}} \end{nscenter}
\begin{nscenter} $d(\vec{x_a}, \vec{x_b}) \neq 0 \Longrightarrow d(\vec{\varepsilon_a}, \vec{\varepsilon_b}) > 0,$ \end{nscenter}
\begin{nscenter} $\forall a, b$ \end{nscenter} \\
\vspace{-15pt}
\begin{nscenter} {\textbf{Stability:}} \end{nscenter}
\begin{nscenter} $p(\vec{x_a}) = \lambda_i = p(\vec{x_b}) \Longrightarrow c(\vec{\varepsilon_a}) = \overline{c_i} = c(\vec{\varepsilon_b})$ \end{nscenter}
\begin{nscenter} $\forall a, b$, \hspace{5pt} $\lambda_i \subset \Lambda$, \hspace{5pt} $\overline{c_i} \subset \overline{C}$ \end{nscenter}

The first two axioms – identity and separability, remain the same as in regression, and are, therefore, not further discussed. The stability axiom, however, undergoes some notable changes. What the third axiom specifies is the following: if two objects have the same class prediction $\lambda_i$, then the two corresponding explanations must belong to the same explanation cluster $\overline{c_i}$. The reasoning behind this axiom is that, in general, when data points belong to the same cluster these are similar. Therefore, if explanations belong to the same cluster, then these must also be similar. In our previous fraud vs no fraud transactions example, this would mean that if our prediction model classifies two certain transactions as fraud, then the two corresponding explanations should end up in the same "fraud explanation cluster". We would thus have a cluster that contains all explanations for fraud, and another one with all explanations for no fraud. We can also apply this logic to a multiclass problem, where more than two classes exist. The more explanations end up in the right cluster, the better. To measure how many explanations ended up in the correct cluster, we use the Jaccard similarity introduced in chapter \ref{ch:Content1:sec:Section1:Subsection3:Subsubsection1}.
\\Another vital aspect to mention again at this point is that the result of the third axiom heavily depends on the used clustering algorithm. As mentioned in section \ref{ch:Content1:sec:Section1:Subsection2:Subsubsection1}, by using hierarchical clustering with single linkage, we are guaranteed to find one amongst a set of optimal solutions, if we disregard ties \citep{jardine1971}. We thus rely on this form of clustering when possible and feasible in the experiments chapter.

Now that we have a good understanding of both, the regression and classification cases of our explanation consistency framework, we proceed to the next subsection, where we discuss how we can verify these axioms in practice.

%% ===========================
\section{Explanation Consistency Verification with Big Data}
\label{ch:Content3:sec:Section3}
%% ===========================
One of the limitations of our explanation consistency framework is that it relies on distance functions and matrices, which are not always computationally feasible, especially if we are dealing with big data. There are, however, different approximations that we can use to verify the axioms, and get an estimate on the quality of an explanation method regarding interpretability. This section is thereby devoted to different proxies with a higher computational efficiency, for the evaluation of explanation consistency.

The first axiom of identity can be efficiently verified, by repeatedly feeding identical objects to the explanation method and checking if it consistently returns the same explanation. This can be performed on a subset of the data, or even with individual objects. This axiom can, moreover, also be verified by reading the documentation of the EM. If there is some random process involved for the generation of the explanation, then this axiom can most likely not be fulfilled. Also if the user needs to fix a seed value to obtain the same explanation repeatedly, then the axiom is also not fulfilled.

We can verify the second axiom of separability as follows: generate the explanations for at least a representative subset of the dataset and check for duplicates in the explanations. If there are no duplicates in the objects, then we would also not expect to find any duplicate explanations, as long as the DOF assumption holds.

The verification of the third axiom in the classification case can be computationally costly with big data. This is because hierarchical clustering relies on the distance matrix. As an alternative, we could use the k-means clustering algorithm without random initialisation. The initial centroids passed to the algorithm could thereby be the averages of all explanations belonging to the objects of a specific class. We could e.g. take all explanations that correspond to objects of the fraud class and average these to obtain the initial centroid for the fraud explanation cluster. The same would be repeated to get the second centroid for the no fraud explanation cluster. After that, we could efficiently compute the Jaccard similarity, to measure how many explanations ended up in the correct group.
\\To verify the third axiom in a regression case, there are mainly two feasible options to improve the efficiency: (1) perform the verification on a random sub-sample, or (2) transform the regression problem into a clustering problem through a data binning (discretisation). Several well-known data binning rules or algorithms could be used to generate a set of initial bins. With the bins, we could fundamentally proceed as in the classification case. For that, we would apply a k-means clustering on the explanations with initial centroids corresponding to the averages of all explanations belonging to the objects in one bin, similarly as above. Finally, we could also compute the Jaccard similarity to identify how many explanations ended up in the right explanation clusters.

In the experiments chapter, we always indicate if an approximation was implemented to compute the values, or whether we used the original axiom formulation. Having discussed how to reduce the computational complexity when needed to verify the explanation consistency, we proceed to the next chapter, where we put our framework to test in practical experiments.

%% ===========================
%% ===========================
\chapter{Experiments and Evaluation}
\label{ch:Content4}
%% ===========================
%% ===========================

\hspace{\parindent}The goal of the experiments chapter is to apply the developed explanation consistency framework in practice, by evaluating different explanation methods with it. To achieve this, we use it on two currently preferred model-agnostic interpretability tools, namely LIME and SHAP, introduced in section \ref{ch:Content1:sec:Section3:Subsection3:subsubsection2}. Moreover, we evaluate both cases of the explanation consistency framework – regression and classification.

Before continuing to the next section, we give a brief overview of the structure of this chapter. We start by introducing the two datasets used in all experiments. Thereafter, we discuss the prediction models that we applied to each dataset with their architecture and configurations. Moreover, we provide data on their evaluation. After that, we apply the LIME and SHAP explanation methods to the data, to generate the explanations. Finally, we use these explanations to verify the explanation consistency of each explanation method on the two datasets.

%% ===========================
\section{Datasets}
\label{ch:Content4:sec:Section1}
%% ===========================
As mentioned above, we use two different datasets in the experiments, to evaluate both cases of the explanation consistency framework. One dataset, therefore, has a target variable that is real-valued (regression case), and the other dataset has a target variable with multiclass labels (classification case). In this section, we look at the features of each of these datasets, analyse some descriptive statistics and interpret some exploratory graphs. We start with the Seattle House Prices dataset, which is used for the regression case, as it is more intuitive and easier to grasp. After that, we discuss the Machine Component Failures dataset, which is used for the classification case.

%% ===========================
\subsection{Seattle House Prices (SHP)}
\label{ch:Content4:sec:Section1:Subsection1}
%% ===========================
The Seattle House Prices (SHP)\nomenclature{SHP}{Seattle House Prices Dataset} dataset contains the house sale prices for over 21,000 homes in King County, which includes the metropolitan area of Seattle. The data was collected for a year-long period between May 2014 and May 2015, and is available on the Kaggle platform (see appendix section \ref{Appendix:Section:C}), famous for hosting predictive modelling and analytics competitions.

The original dataset contains nineteen features for each house and an additional two columns for id and price. The first step that we took in the preparation of this dataset, was to eliminate highly correlated features, to avoid the multicollinearity problem. After removing correlated features and the date column, which is irrelevant for our purpose, sixteen features remain. These in addition to the price label, can be consulted in table \ref{Table 4.1}, where we also provide some descriptive statistics for each of the features, including minimum, maximum, mean, median and standard deviation.
\\The less self-explanatory features are waterfront, view, condition and grade, which we discuss hereafter. Waterfront is a binary variable that takes on the value of 0 for no waterfront and 1 for waterfront. The view feature is not related to the quality of the view, but rather to the number of times that the house has been viewed, i.e visited by interested parties before the sale. Condition is the overall
%%%%%%%%%%%%% Table %%%%%%%%%%%%%
\begin{table}[h]
\centering
\scalebox{0.80}{
\begin{tabular}{lcccccc}
\toprule
Feature / Label	& Type 			& Minimum 	& Maximum 	& Mean 		& Median 	& STD \\
\midrule
Bedrooms 		& Categorical 	& 0.00 		& 33.00		& 3.37		& 3.00 		& 0.93 		\\
Bathrooms 		& Categorical 	& 0.00 		& 8.00 		& 2.12 		& 2.25 		& 0.77		\\
Sqft\_Living 	& Numerical 	& 290 		& 13,540 	& 2,083 	& 1,920 	& 919		\\
Sqft\_Lot		& Numerical 	& 520 		& 1,651,359 & 15,145 	& 7,617 	& 41,553 	\\
Floors			& Categorical 	& 1.00 		& 3.50 		& 1.50 		& 1.50		& 0.54 		\\
Waterfront 		& Binary	 	& 0.00 		& 1.00 		& 0.01 		& 0.00 		& 0.09 		\\
View 			& Categorical 	& 0.00 		& 4.00 		& 0.24 		& 0.00 		& 0.77 		\\
Condition 		& Categorical 	& 1.00 		& 5.00 		& 3.41 		& 3.00 		& 0.65 		\\
Grade 			& Categorical 	& 1.00 		& 13.00		& 7.66 		& 7.00 		& 1.17 		\\
Sqft\_Above 	& Numerical 	& 290 		& 9,410 	& 1,791 	& 1,560 	& 829 		\\
Sqft\_Basement 	& Numerical 	& 0 		& 4,820 	& 292 		& 0			& 443 		\\
Yr\_Built 		& Categorical 	& 1900.00	& 2015.00	& 1971.00 	& 1975.00 	& 29.38 	\\
Yr\_Renovated 	& Categorical 	& 0.00 		& 2015.00	& 84.79 	& 0.00 		& 402.57 	\\
Zipcode 		& Categorical 	& 98001.00 	& 98199.00 	& 98077.86 	& 98065.00 	& 53.47 	\\
Lat 			& Categorical 	& 47.16 	& 47.78 	& 47.56 	& 47.57 	& 0.14 		\\
Long 			& Categorical 	& -122.52 	& -121.31 	& -122.21 	& -122.23 	& 0.14 		\\
Price 			& Numerical 	& 75,000 	& 7,700,000	& 540,607	& 450,000 	& 367,782 	\\
\bottomrule
\end{tabular}}
\caption{Seattle House Prices: Descriptive Statistics Overview}
\label{Table 4.1}
\vspace{-10pt}
\end{table}
%%%%%%%%%%%%% Table %%%%%%%%%%%%%
condition of the house on a scale from 1 to 5, whereas 5 corresponds to the best condition. Finally, the grade feature is the overall grade that is given to the house, according to King County's grading system, which rates houses on a scale from 1 to 13, where 13 is the best. With this in mind, we can proceed with the analysis of the descriptive data.
\\We can see some curious numbers in table \ref{Table 4.1}, e.g. there was a house that sold with 33 bedrooms. Also, one of the estates had a large lot size of over 1.6M square-feet. Another interesting finding is when we compare the means with the medians. For instance, the mean basement has a size of 292 square-feet. However, the median is 0, which means that at least half of the houses do not even have a basement. A further example of a feature whose mean is not meaningful is the year renovated. If there are no records of a renovation, the year renovated is set to 0. A better solution would have been to set the number to
%%%%%%%%%%%%% Figure %%%%%%%%%%%%%
\begin{wrapfigure}{r}{0.55\textwidth}
  \vspace{-20pt}
  \begin{center}
    \includegraphics[width=0.55\textwidth]{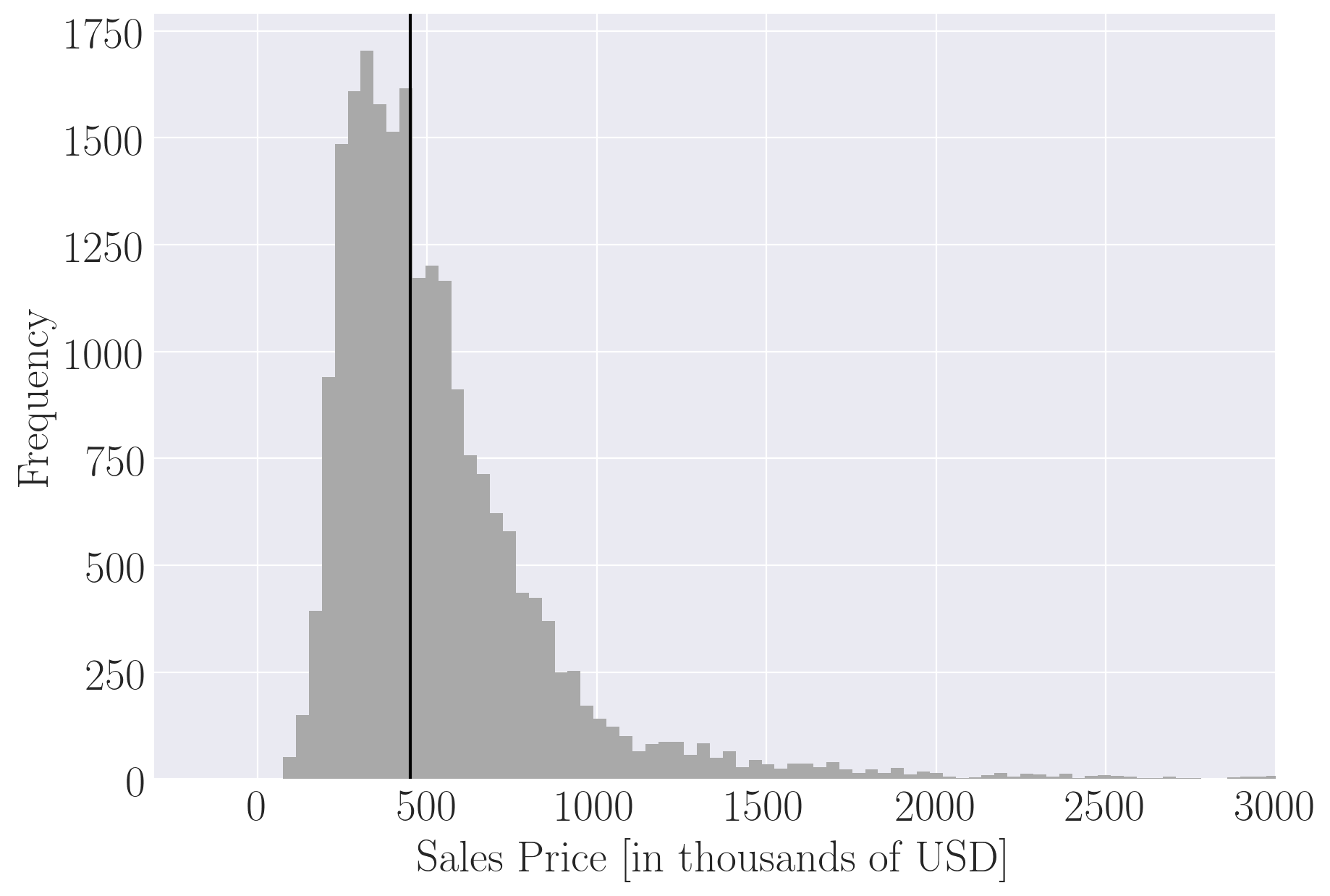}
  \end{center}
  \vspace{-20pt}
  \caption{Sales Price Distribution of Homes in King County}
  \label{figure 4.1}
  \vspace{-10pt}
\end{wrapfigure}
%%%%%%%%%%%%% Figure %%%%%%%%%%%%%
the same year as year built. Consequently, the mean of year renovated is now 84.79, which has no meaning, as we do not know how many houses have been renovated (where Yr\_Built is unequal to zero) and how many houses have not, without further analysis. One last curiosity that we wish to highlight is the price. As we can see in table \ref{Table 4.1}, the mean price is roughly 541k, whereas the median is only 450k. From that data, we can infer that more than half of all houses have a lower price than the mean and that there must be some extraordinarily expensive houses, which increase the mean in such way that it ends up being over 90k higher than the median sales price of a house. We can observe this phenomenon in figure \ref{figure 4.1}, where we set the cut-off price to 3M (the maximum price would be at 7,7M). In the histogram, we observe that there is a high density of prices before the median at 450k, marked by the vertical black line. After the median value, the density is much lower and spread out, creating that long right tail, which explains the far-apart mean and median values.

After numerically and visually exploring the SHP dataset, and understanding all features, we proceed to analyse our second dataset of Machine Component Failures (MCF)\nomenclature{MCF}{Machine Component Failures Dataset}, in the upcoming section.

%% ===========================
\subsection{Machine Component Failures (MCF)}
\label{ch:Content4:sec:Section1:Subsection2}
%% ===========================
The Machine Component Failures dataset contains data on a hundred machines, collected over the course of one year, between 2015 and 2016. Disregarding duplicates, the original dataset contains one measurement per hour and machine for 365 days, amounting to 876k rows. It is composed of five separate datasets: telemetry data (e.g. pressure and vibration); error data (when machines display an error message but are still working); maintenance data (scheduled or unscheduled); machine data (e.g. age and model); and failure data (when machines actually failed, i.e. stopped working). For our purposes, we exclude the maintenance data, as it is a strong machine failure indicator. The MCF dataset is also publicly available on the Kaggle platform (see appendix section \ref{Appendix:Section:C}).

This dataset did not show significant correlations between features, however, due to the highly skewed nature of the data, some feature engineering (FE) was required. Machine failures are generally sporadic events, as compared to machine uptime. We, therefore, created 24-hour rolling mean and standard deviations for the telemetry features, which is an often-used procedure in practice. This means that at a specific time-stamp, e.g. on the 01.02.2015 at 08:00 the rolling mean of vibration would be the mean of the last 24 hours from that time-stamp backwards, rather than only the measured value at 08:00. This enables to represent the short-term history of the telemetry data over the chosen window size (24h in our case).
\\Apart from rolling aggregates for telemetry data, we also used error-based lag features. This is a similar procedure as above, however, as errors are absolute numbers, it makes more sense to look at the sum of errors over a particular lagging window, rather than averages. To be consistent with the FE of the telemetry features, we also choose a window size of 24h. So if an error occurred, e.g. on the 01.02.2015 at 08:00, it would still be visible 23h later, so all rows between the 01.02.2015 at 08:00, and the 02.02.2015 at 07:00, would be marked with the same error code. This helps us to understand longer-term effects of errors, which may co-occur with each other. With this in mind, we can proceed with the analysis of the descriptive data in table \ref{Table 4.2}, in the following.

Through the feature engineering process, some data had to be cut off for each machine to avoid rows with incomplete information, which decreased the total number of rows to roughly 873k. There are a total of nineteen features, and an additional label column.
\\The first curiosity that we spot in table \ref{Table 4.2} is the difference that occurs between the original, and the over 24h aggregated telemetry data. As an example, if we look at the minimum and maximum of the volt feature, and compare these two numbers with the ones
%%%%%%%%%%%%% Table %%%%%%%%%%%%%
\begin{table}[h]
\centering
\scalebox{0.80}{
\begin{tabular}{lcccccc}
\toprule
Feature / Label			& Type 			& Minimum 	& Maximum 	& Mean 		& Median 	& STD 		\\
\midrule
Volt	 				& Numerical 	& 97.33		& 255.12	& 170.78	& 170.61	& 15.51		\\
Rotate	 				& Numerical 	& 138.43	& 965.02	& 446.56	& 447.53	& 52.71		\\
Pressure	 			& Numerical 	& 51.24		& 185.95 	& 100.87 	& 100.43 	& 11.06		\\
Vibration				& Numerical 	& 14.88		& 76.79		& 40.39 	& 40.24 	& 5.37	 	\\
Volt\_Mean\_24h			& Numerical 	& 155.37	& 220.78	& 170.78	& 170.21	& 4.74 		\\
Rotate\_Mean\_24h 		& Numerical	 	& 265.79	& 502.22	& 446.56	& 449.20	& 18.20		\\
Pressure\_Mean\_24h		& Numerical 	& 90.35		& 152.66	& 100.87	& 100.10	& 4.76 		\\
Vibration\_Mean\_24h	& Numerical 	& 35.06		& 61.93		& 40.39		& 40.07		& 2.07 		\\
Volt\_STD\_24h			& Numerical 	& 6.38 		& 28.88		& 14.92		& 14.85		& 2.26 		\\
Rotate\_STD\_24h 		& Numerical 	& 18.39		& 105.33 	& 49.95 	& 49.62 	& 7.68 		\\
Pressure\_STD\_24h 		& Numerical 	& 4.15 		& 28.87		& 10.05		& 9.92		& 1.71 		\\
Vibration\_STD\_24h 	& Numerical 	& 2.07		& 12.66		& 5.00	 	& 4.96	 	& 0.80	 	\\
Error1\_24h 			& Categorical 	& 0.00 		& 2.00		& 0.03	 	& 0.00 		& 0.17	 	\\
Error2\_24h				& Categorical 	& 0.00	 	& 2.00	 	& 0.03	 	& 0.00	 	& 0.17	 	\\
Error3\_24h				& Categorical 	& 0.00	 	& 2.00	 	& 0.02	 	& 0.00	 	& 0.15 		\\
Error4\_24h				& Categorical 	& 0.00	 	& 2.00	 	& 0.02	 	& 0.00	 	& 0.14 		\\
Error5\_24h				& Categorical 	& 0.00	 	& 2.00	 	& 0.01	 	& 0.00	 	& 0.10 		\\
Model	 				& Categorical 	& 1.00	 	& 4.00	 	& 2.83	 	& 3.00	 	& 1.05 		\\
Age 					& Categorical 	& 0.00	 	& 20.00 	& 11.33 	& 12.00 	& 5.83 		\\
Failure 				& Categorical 	& 0.00	 	& 4.00		& 0.05		& 0.00	 	& 0.37	 	\\
\bottomrule
\end{tabular}}
\caption{Machine Component Failures: Descriptive Statistics Overview}
\label{Table 4.2}
\vspace{-10pt}
\end{table}
%%%%%%%%%%%%% Table %%%%%%%%%%%%%
of the Volt\_Mean\_24h feature, we see that range of the former is about 158, whereas the range of the latter is only about 65. This means that the aggregated data is much smoother, as compared to the original data, which is also confirmed by the lower standard deviation. This phenomenon likewise occurs for the other telemetry features of rotation, pressure and vibration. The only measure that must remain equal for both – the original and aggregated features, is the mean value, which can also be observed in the table.
\\The mean of the 24h rolling standard deviation features, corresponds approximately to the STD of the original telemetry features. Like above, these values are also smaller than the original ones, due to the smoothing effect of computing rolling aggregates.
\\The lagging error features all vary between zero and two errors, whereas error codes 1 and
%%%%%%%%%%%%% Figure %%%%%%%%%%%%%
\begin{wrapfigure}{l}{0.45\textwidth}
  \vspace{-20pt}
  \begin{center}
    \includegraphics[width=0.45\textwidth]{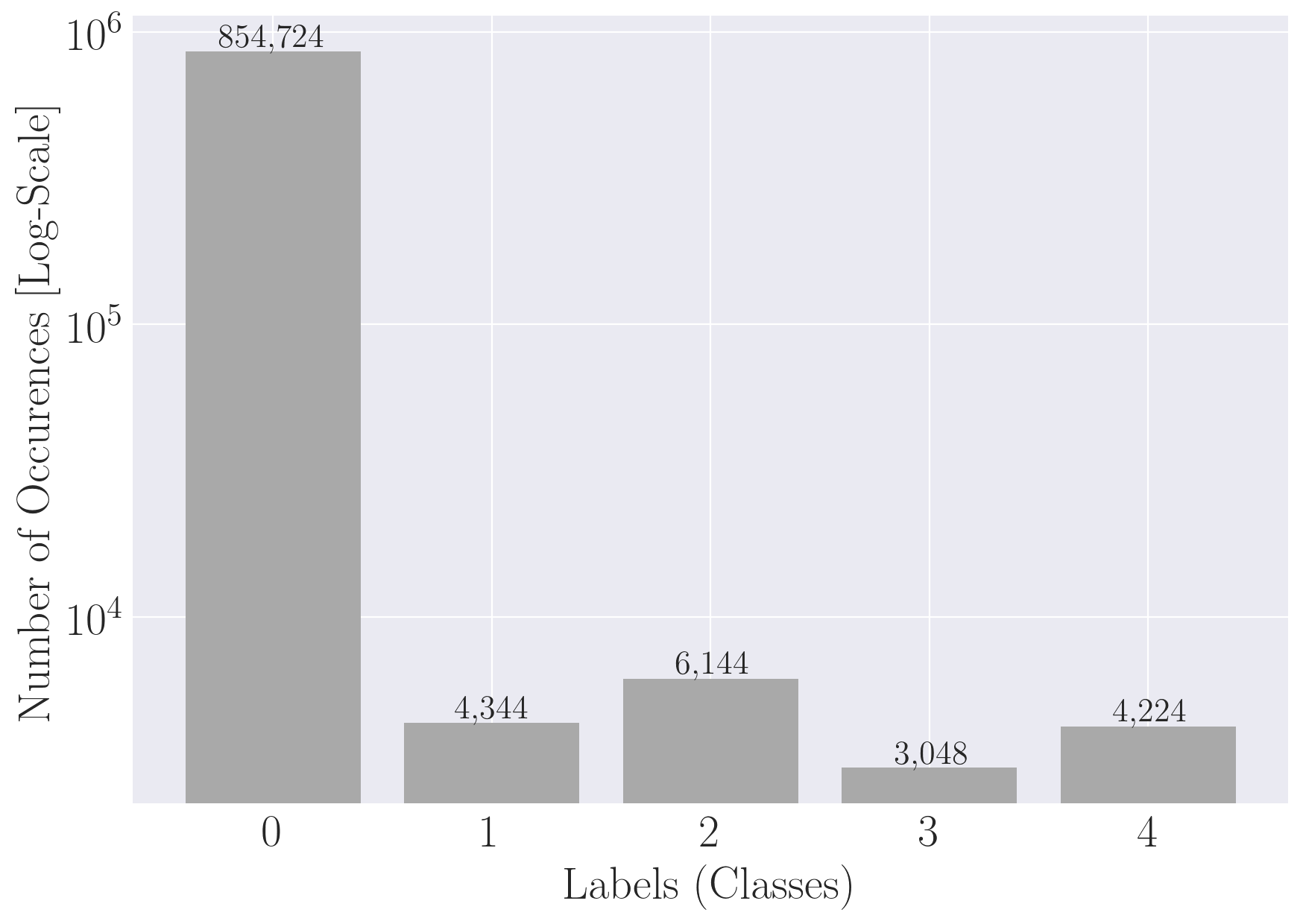}
  \end{center}
  \vspace{-20pt}
  \caption{Machine Failure Type Distribution}
  \label{figure 4.2}
  \vspace{-10pt}
\end{wrapfigure}
%%%%%%%%%%%%% Figure %%%%%%%%%%%%%
2 occur most frequently, and error code 5 occurs least frequently, which we can infer from the mean values. The median of all errors is zero, which we expected, as errors occur very rarely. Lastly, the standard deviation of errors is higher for more frequent errors, which seems intuitive.
\\The model feature shows that machines of model 1 and 2 occur less frequently than models 3 and 4, as the median corresponds to model 3, and as the mean is noticeably larger than 2. The mean age between all machines and models is 11.33 years, whereas the standard deviation is quite high at 5.83 years.
\\Finally, the failure feature takes on integer values from 0 to 4. A value of 0 corresponds to no failure and integers from 1 to 4 represent 4 different types of machine failures (damages). As we can see in table \ref{Table 4.2}, the median is zero as failures occur in less than 50\% of all objects. The mean and STD values are not meaningful to interpret, as the features take on integer values from 0 to 5, and not only binary values. The label and failure distribution over all timestamps (objects) are shown in figure \ref{figure 4.2}. It is noteworthy that the y-axis in the figure is on a log-scale; otherwise, the occurrences of labels 1 to 4 would not be visible. The absolute number of occurrences for each label is written on top of each bar and is dominated by label 0, i.e. no failure. Furthermore, failure of type 2 occurs most frequently and about double as much as failure 3. The two remaining failure types, 1 and 4 occur roughly equally frequently.

Having visually and descriptively explored the MCF dataset, and gained an understanding of its features, we proceed to the next section, in which we fit and evaluate different ML models on the two introduced datasets.

%% ===========================
\section{Prediction Models}
\label{ch:Content4:sec:Section2}
%% ===========================
In this section, we elaborate on different ML models that we employed on each of our two datasets. The process of finding a high performing model is often lengthy and requires to not only test different parameter combinations for the same models, but also completely different models. Furthermore, certain models and configurations that perform well on one dataset, may achieve a poor predictive performance on the next dataset.
\\We do not review every single model that we apply in-depth, as that would go beyond the scope of this thesis. Nevertheless, we elaborate on the best-performing models for each of our two datasets.

%% ===========================
\subsection{Model Fitting}
\label{ch:Content4:sec:Section2:Subsection1}
%% ===========================
To get meaningful results with explanation methods, it is imperative that the underlying prediction models are very accurate. This is so because EMs explain predictions from ML models. So if many predictions are already inaccurate, then the EMs explain inaccurate predictions, which obviously does not make much sense. We thus only select prediction models that are highly accurate, but at the same time also do not overfit to the training data.

In the following two paragraphs, we first discuss the best-performing models for the Seattle House Prices dataset, and after that, we repeat the same process for the Machine Component Failures dataset.

%% ===========================
\paragraph{Seattle House Prices Dataset: Models, Architectures and Configurations}
\label{ch:Content4:sec:Section2:Subsection1:Paragraph1}
\vspace{-7pt}
%% ===========================
We applied two models to the SHP dataset, namely XGB and LightGBM. The former was already discussed in section \ref{ch:Content1:sec:Section1:Subsection1:Subsubsection1}, and the latter is disregarded for further consideration, as it did not achieve a high enough predictive performance, as compared to XGB. More concretely, there was an overfitting tendency with a significant accuracy loss, when adding regularization to control overfitting. In the following, we thus only review the XGB model.

There are many settings that the original model-API provides for configuration, of which we highlight the few ones that we consider to be most important, hereafter.
\\\textit{Objective}: as the labels of the SHP dataset are continuous values, we need to choose regression (reg:linear) as the objective function. Other options include, e.g. multi-softmax, which we would use for a multiclass problem, such as our MCF dataset.
\\\textit{Boosting rounds}: this number corresponds to the trees used in the ensemble. We set this number to a hundred. If we look back at figure \ref{figure 2.1}, it would mean that we use a hundred trees trained in sequential form, each improving on the error of the last tree.
\\\textit{Eta}: corresponds to the learning rate, i.e. how much the algorithm adjusts the weights of our trees in each training step. We set this parameter to 0.10, to make the boosting process more conservative, thereby avoiding overfitting.
\\\textit{Evaluation Metric}: this setting specifies the measure that is used by the model when evaluating its performance during training. As we have a regression objective, we set this to the root mean square error (RMSE)\nomenclature{RMSE}{Root Mean Square Error}, which is just the square root of the previously discussed MSE in section \ref{ch:Content1:sec:Section1:Subsection1}.
\\\textit{Max Depth} and \textit{Min Child Weight}: these two final parameters that we discuss, both serve to control overfitting, similarly to eta. The former corresponds to the maximum depth that each boosted tree is allowed to have amongst the hundred trees. We set this measure to five. The latter relates to the minimum number of instances that need to be in each leaf-node of the tree, which we set to three instances. The higher this number is, the more conservative and less prone to overfitting is the ML model.

With a good overview of the most critical configurations that we chose for our XGB model, we proceed to discuss the settings of the three best-performing prediction models on the MCF dataset.

%% ===========================
\paragraph{Machine Component Failures: Models, Architectures and Configurations}
\label{ch:Content4:sec:Section2:Subsection1:Paragraph2}
\vspace{-7pt}
%% ===========================
We applied multiple models to the MCF dataset, starting with a logistic regression, which due to the non-linear relationships in the data, did not return satisfactory results. Next, we trained a support vector machine (SVM)\nomenclature{SVM}{Support Vector Machine} with a non-linear radial basis function (RBF)\nomenclature{RBF}{Radial Basis Function} kernel. This model is, however, not very scalable to big data, and as our training set had over 600k instances, we could not even train the model in a feasible time. We thus decided to use a tree-based approach, starting with a random forest and a XGB model thereafter. The latter outperformed the random forest by several orders of magnitude, without overfitting to the training set. After that, we applied a MLP model, which achieved a similar performance as the XGB. Finally, we fitted a LSTM to the data, which scored an almost identical performance as compared to the XGB and MLP models, however, without any feature engineering. In the following, we briefly discuss the chosen configurations and architectures of each of the three models – XGB, MLP and LSTM.

We start with the XGB model, as we already discussed its parameters above in the last paragraph. We set the \textit{objective} to multi-softprob, which is used for multiclass classification problems, where we wish to also obtain the estimated probabilities for each class, rather than only the predicted class. As before we also used 100 \textit{boosting rounds}, i.e. trees, and a learning rate \textit{eta} of 0.10. As the MCF dataset is of classification, we could not use the RMSE as \textit{evaluation metric}, but used log-loss (cross entropy loss) instead. Finally, the \textit{max depth} was set to three, as any higher number lead to increased overfitting.

For the multilayer perceptron, different parameters exist, as it is a feed-forward neural network, rather than a tree-based approach. In an artificial neural network, we have first to define the architecture of the network. This includes choosing the number of units (neurons) per hidden layer and the number of hidden layers. We found that for the MCF dataset two hidden layers lead to the best result, with 100 neurons in the first, and 50 neurons in the second hidden layer. As activation function for the hidden layers, we used a rectified linear unit (RELU)\nomenclature{RELU}{Rectified Linear Unit}, which improved the predictive performance the most. Finally, we employed a constant learning rate of 0.001, which is much lower than before, but the MLP achieved much more stable results with a lower learning rate than the XGB model.

The long short-term memory neural network has a similar architecture to the MLP. A crucial difference, however, is that the LSTM operates on temporal data. That means that we must choose a sequence length, which is a similar concept to the window size, as discussed in section \ref{ch:Content4:sec:Section1:Subsection2}. To have a fair comparison with the other ML models, we choose the sequence length to be 24, as we used lagging and rolling windows with size 24 in the feature engineering process. Concerning the architecture, we also choose two hidden layers, with 100 units in the first and 50 units in the second layer. Moreover, we added a 0.20 dropout to each hidden layer to control overfitting. Lastly, as objective, we used sparse categorical cross entropy, which is for multiclass problems, and a softmax activation function in the dense layer.

After talking about the specific architectures and configurations of our ML models for each of the two datasets, we proceed to the next section where we evaluate each of the discussed models.

%% ===========================
\subsection{Model Evaluation}
\label{ch:Content4:sec:Section2:Subsection2}
%% ===========================
In this section we continue with the evaluation of the introduced ML models. We maintain the same structure as in the previous sub-chapter, first evaluating the performance of the XGB model on the SHP dataset, and thereafter the performance of the XGB, MLP and LSTM models on the MCF dataset.

%% ===========================
\paragraph{Seattle House Prices Dataset: Model Evaluation}
\label{ch:Content4:sec:Section2:Subsection2:Paragraph1}
\vspace{-7pt}
%% ===========================
As the SHP dataset features a regression target label, we must use evaluation metrics such as the previously introduced coefficient of determination (R\textsuperscript{2} measure) and the RMSE. The unbiased performance estimate on the test set for R\textsuperscript{2} is 0.89. That means that our XGB model explains 89\% of the variability of the data around its mean. We consider this to be a good value, however, it may be more meaningful to look at the R\textsuperscript{2} measure together with the RMSE, which is 120,478. That means that the average deviation from our predicted value to the real value is roughly 120k. Considering a mean house price of around 540k, this represents a 22\% deviation. An aspect to keep in mind, however, is that outliers greatly influence both of the discussed measures, meaning that if these would have been removed, both measures – R\textsuperscript{2} and RMSE, could improve significantly.

A more visually appealing evaluation of the XGB model quality provides the following scatterplot in figure \ref{figure 4.3}. On the x-axis, we plot the actual price of houses in the test set, and on the y-axis the predicted house price. The ideal prediction line is represented by the black diagonal. The closer all points are to this diagonal, the better is the
%%%%%%%%%%%%% Figure %%%%%%%%%%%%%
\begin{wrapfigure}{r}{0.50\textwidth}
  \vspace{-20pt}
  \begin{center}
    \includegraphics[width=0.50\textwidth]{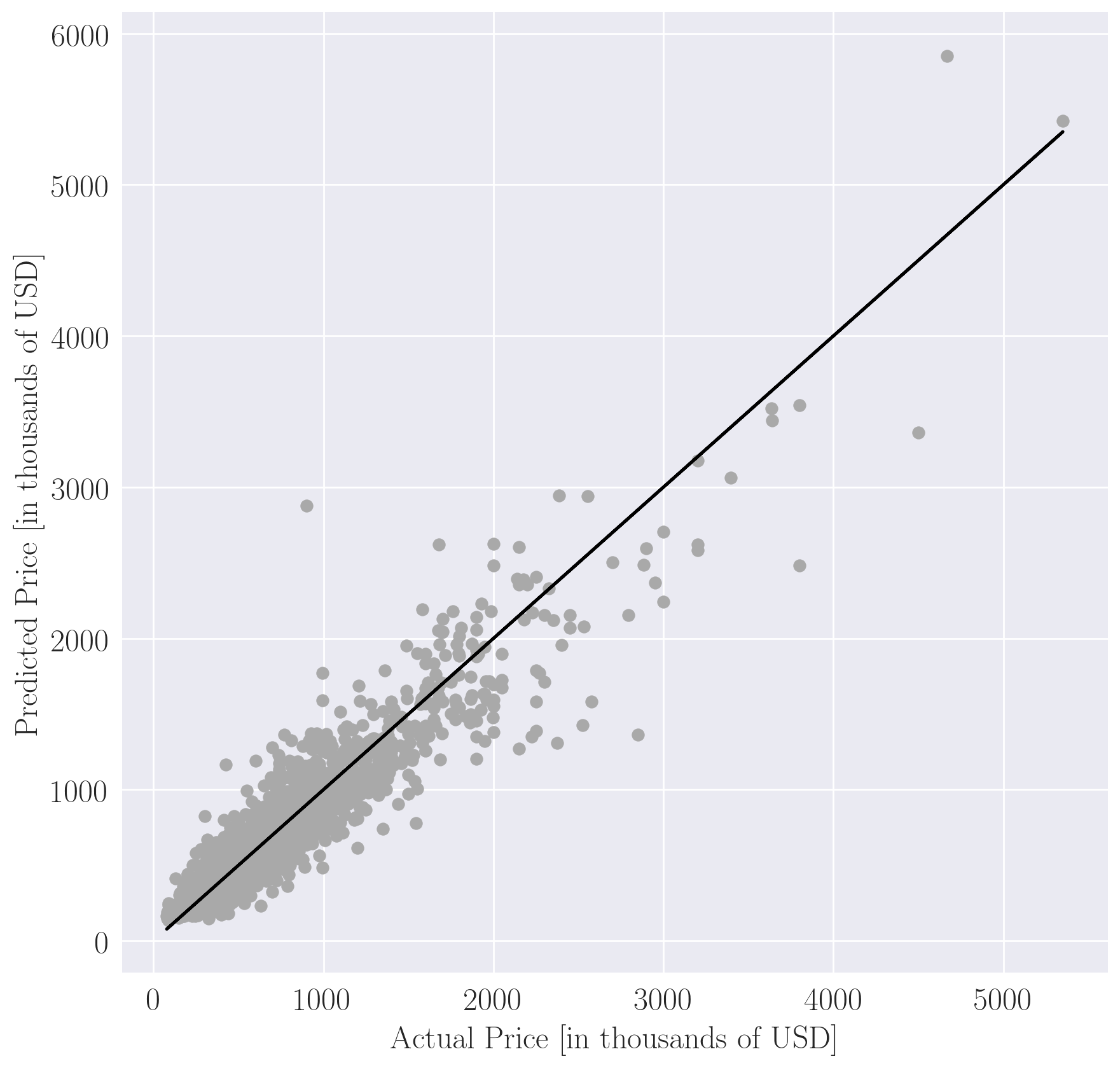}
  \end{center}
  \vspace{-20pt}
  \caption{Predicted vs Actual Prices of King County Homes}
  \label{figure 4.3}
  \vspace{-10pt}
\end{wrapfigure}
%%%%%%%%%%%%% Figure %%%%%%%%%%%%%
performance of our prediction model. We explain this with the following example. Imagine that we predict the price of a house that costs 1M, and let us say that our XGB model predicts 950k. These two prices represent the coordinates of a point in our scatter plot, namely (x, y) = (1M, 950k). If we would have predicted 1M, then the point would be (1M, 1M), i.e. it would lie on the black diagonal. However, our point is not exactly on the diagonal, but a little below. It is therefore desirable that all predictions are on, or as close as possible to the black diagonal.
\\Figure \ref{figure 4.3} moreover allows us to analyse the distribution of our predictions vs real values. We can easily spot outliers, where our predictions were particularly bad, such as the point that has coordinates of approximately (1M, 3M), where our estimate for the house price was nearly 2M off. Even though this could potentially represent an interesting investment opportunity, in reality, it are exactly these types of points that have a significant influence on the scores of the above-discussed measures (R\textsuperscript{2} and RMSE).

An advantage of using the XGB model is that it provides easily accessible information regarding the overall importance of all features used in the model. This improves the global model interpretability as discussed in section \ref{ch:Content1:sec:Section2:Subsection2:Subsubsection2}. For the SHP dataset, we plotted these global feature importance scores in figure \ref{figure 4.4}. On the y-axis, we plotted each of the sixteen features in descending order of their relevance. On the x-axis, the
%%%%%%%%%%%%% Figure %%%%%%%%%%%%%
\begin{wrapfigure}{l}{0.55\textwidth}
  \vspace{-20pt}
  \begin{center}
    \includegraphics[width=0.55\textwidth]{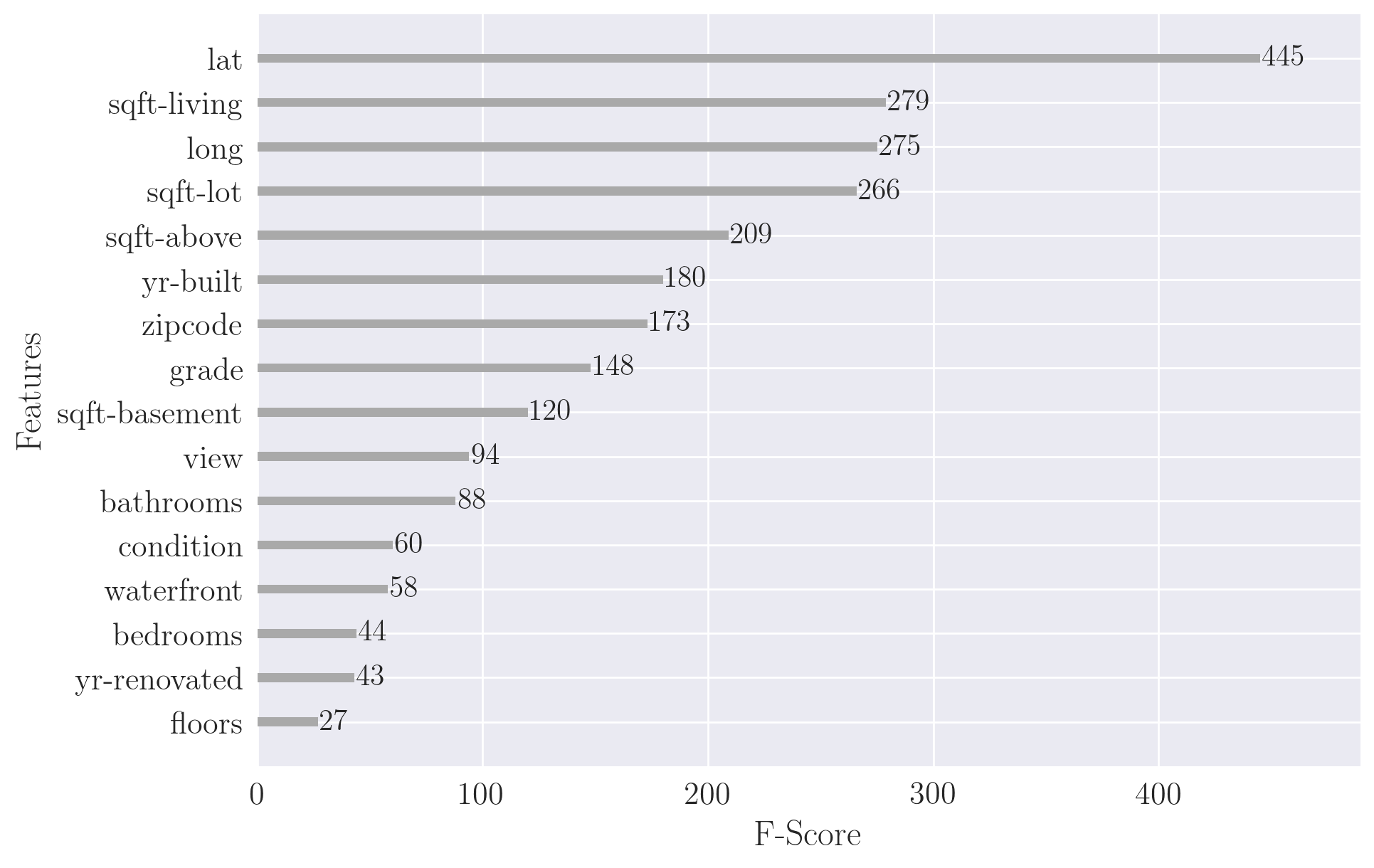}
  \end{center}
  \vspace{-20pt}
  \caption{Global Feature Importances of the XGB model and SHP Dataset}
  \label{figure 4.4}
  \vspace{-10pt}
\end{wrapfigure}
%%%%%%%%%%%%% Figure %%%%%%%%%%%%%
importance of each feature is expressed through the F-Score. We determine this score as the sum of the times that the feature was split on in all trees of our XGB model. As we can see, latitude is the feature with the highest F-score and longitude the feature with the third-highest score. This means that in the SHP dataset, the location of a house generally has a high influence on its price. Note that this is only the general feature importance. If we look at the predictions of an individual house, the order of feature significances may be distinct to the order shown in figure \ref{figure 4.4}, due to the non-linearity of the prediction model. Further globally important features are the living area and the area of the lot in square-feet.

With a robust performance evaluation of the XGB model on the SHP dataset, we continue to the next paragraph, where we analyse how the three models – XGB, MLP and LSTM perform on the MCF dataset.

%% ===========================
\paragraph{Machine Component Failures: Model Evaluation}
\label{ch:Content4:sec:Section2:Subsection2:Paragraph2}
\vspace{-7pt}
%% ===========================
The MCF dataset features a multiclass target variable, which means that we must use the confusion matrix (CM), and measures such as precision, recall and the f1-score to asses the performance of our ML classifiers.
\\The CM compares actual classes with predicted classes and enables the calculation of the other mentioned measures. Precision indicates how often our classifier is correct when it predicts a certain label. An example for precision in our MCF dataset would be that if our classifier predicts the failure of component 1 ten times, and it would only be that type of failure in eight of those cases, then the precision would be 8 out of 10, or 80\%. Recall, also known as \textit{sensitivity}, measures how often our classifier predicts a certain class, out of the total occurrences of that class. As an example, if the failure of component 1 occurs ten times in total, and our model only recognises that failure nine times, then recall would be 9 out of 10, or 90\%. In sum, recall tells us "when something happens, how often do we recognise it" and precision tells us "when we predict something, how often is our prediction correct". Both of these measures are crucial in the performance analysis of a ML classifier. However, it is sometimes challenging to compare distinct classifiers based on these two criteria. As an example, it is challenging to say what is better, classifier 1 with 98\% precision and 92\% recall, or classifier 2 with 95\% precision and 96\% recall. The f1-score (not to be confused with the previous F-Score) provides an answer to this question: it is the harmonic mean between recall and precision and varies between 0 and 1, whereas 1 would mean perfect precision and recall. To answer the previous question, the f1-scores for classifier 1 is 94,91\%, and for classifier 2 is 95,50\%. As a side-note, we want to emphasise that the f1-score is only a means to get a combined view of precision and recall. In practice, whether we optimise a ML model for one or the other measure is often a question that we must answer with regards to the context of the situation or problem.

We start by evaluating the XGB classifier with the previously mentioned CM and performance measures. The results for this model can be seen in the following table \ref{Table 4.3}.
%%%%%%%%%%%%% Table %%%%%%%%%%%%%
\begin{table}[h]
\centering
\scalebox{0.75}{
\begin{tabular}{cccccccccc}
\toprule
& & \multicolumn{5}{c}{Predicted} & \multicolumn{3}{c}{Classification Report} \\
& &	No Failure & Failure C1 & Failure C2 & Failure C3 & Failure C4	& Precision & Recall & F1-Score \\
\midrule
\parbox[t]{2mm}{\multirow{5}{*}{\rotatebox[origin=c]{90}{Actual}}} 	& No Failure & 1.00 & 0.00 & 0.00 & 0.00 & 0.00 & 0.9989 & 0.9983 & 0.9986 \\
                        											& Failure C1 & 0.06 & 0.87 & 0.03 & 0.00 & 0.03 & 0.8007 & 0.8741 & 0.8358 \\
                        											& Failure C2 & 0.04 & 0.01 & 0.91 & 0.00 & 0.04 & 0.9214 & 0.9117 & 0.9165 \\
                        											& Failure C3 & 0.05 & 0.02 & 0.00 & 0.92 & 0.01 & 0.9221 & 0.9229 & 0.9225 \\
                        											& Failure C4 & 0.04 & 0.01 & 0.03 & 0.01 & 0.90 & 0.8640 & 0.9025 & 0.8828 \\
\bottomrule
\end{tabular}}
\caption{MCF Dataset: XGB Confusion Matrix and Classification Report}
\label{Table 4.3}
\vspace{-10pt}
\end{table}
%%%%%%%%%%%%% Table %%%%%%%%%%%%%
The first part of the table corresponds to the confusion matrix, and the last three columns to the performance measures, i.e. classification report. To interpret the CM we can take a row-based view. As an example, when "Failure C2" (failure of component 2) actually occurred, our classifier correctly predicted this failure in 91\% of those cases. However, out of all "Failure C2" cases, it also wrongly predicted: "No Failure" with 4\%, "Failure C1" with 1\%, and "Failure C4" with 4\% frequency. This means that in sum, the classifier correctly predicted 91\% of all "Failure C2" cases, but also returned incorrect predictions for the other 9\%.
\\The XGB classifier achieved good results overall. The most difficult label to predict for this model seems to be "Failure C1", which was only correctly predicted in 87\% of all cases, whereas all other values on the diagonal of the CM are 90\% or higher. 
\\The performance measures allow us to gain some deeper insight, and to uncover further curiosities. As we can see in the precision column, the lowest values were achieved for "Failure C1" and "Failure C4", with respectively 80\% and 86\%. Furthermore, in the recall column, we notice that the only value below 90\% is "Failure C1". Expectedly, the f1-score is thus the lowest for "Failure C1", with 84\%. We thereby conclude that the XGB model is strong in correctly predicting the "No Failure", "Failure C2" and "Failure C3" labels, but shows some weakness for the "Failure C4", and especially for "Failure C1" labels.
\\Finally, in analogy to figure \ref{figure 4.4}, we briefly look at the global importance scores of the features of the XGB classifier on the MCF dataset in figure \ref{figure 4.5}. As we can see in the diagram,
%%%%%%%%%%%%% Figure %%%%%%%%%%%%%
\begin{wrapfigure}{r}{0.55\textwidth}
  \vspace{-20pt}
  \begin{center}
    \includegraphics[width=0.55\textwidth]{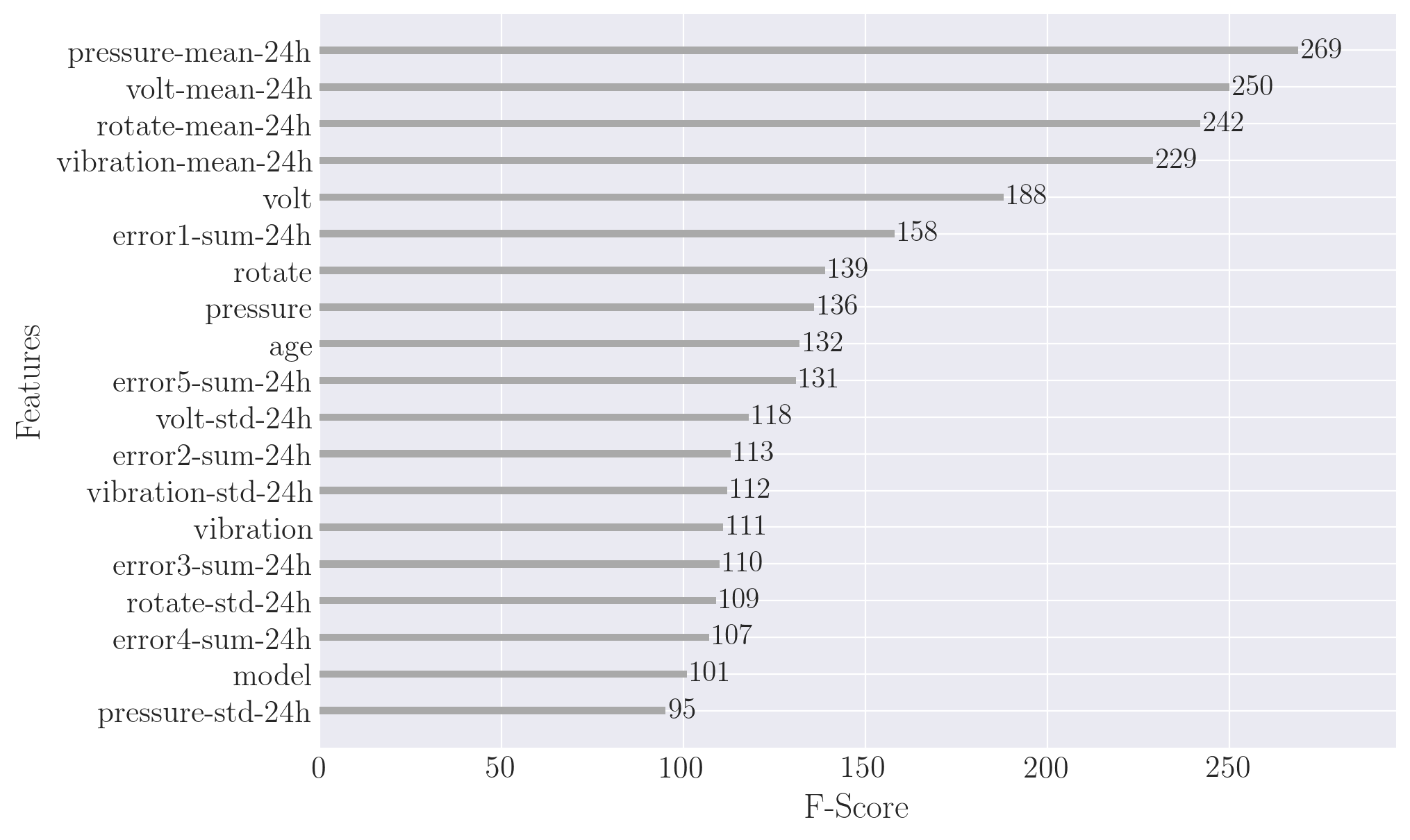}
  \end{center}
  \vspace{-20pt}
  \caption{Global Feature Importances of the XGB model and MCF Dataset}
  \label{figure 4.5}
  \vspace{-10pt}
\end{wrapfigure}
%%%%%%%%%%%%% Figure %%%%%%%%%%%%%
the four telemetry-based features with 24h rolling aggregates, created through the feature engineering, achieved the highest F-Scores. Curiously, in the non-aggregated measures, the volt feature has a higher importance than the pressure feature, which is exactly the opposite of the aggregated features. Age and errors 1 and 5 also still have a strong global influence, which seems intuitive. Lastly, the model type and some of the STD features do not seem to be globally as important for the classifier. Once again, these are only global feature importances, the significance of these features for individual predictions may look very different due to the non-linearity of the prediction model.
\\With these performance insights on the XGB model, we proceed to analyse how the MLP and LSTM models stack up against it in the following.

We continue by evaluating the MLP classifier with the same methodology as above. The results for this model can be seen in table \ref{Table 4.4} on the following page. We start by looking at the diagonal of the confusion matrix and compare it to the CM of the XGB model. There is no difference for the "No Failure" feature, however, the "Failure C1", "Failure C2" and "Failure C3" features all score 0.94, which is higher than the values for those labels of the XGB model, respectively, 0.87, 0.91 and 0.92. A significant difference occurs in the "Failure C4" label, in which the MLP model only recognizes 81\% of all cases, as compared to 90\% of the XGB classifier. We note that the MLP model often predicts "Failure C2",
%%%%%%%%%%%%% Table %%%%%%%%%%%%%
\begin{table}[h]
\centering
\scalebox{0.75}{
\begin{tabular}{cccccccccc}
\toprule
& & \multicolumn{5}{c}{Predicted} & \multicolumn{3}{c}{Classification Report} \\
& &	No Failure & Failure C1 & Failure C2 & Failure C3 & Failure C4	& Precision & Recall & F1-Score \\
\midrule
\parbox[t]{2mm}{\multirow{5}{*}{\rotatebox[origin=c]{90}{Actual}}} 	& No Failure & 1.00 & 0.00 & 0.00 & 0.00 & 0.00 & 0.9991 & 0.9976 & 0.9983 \\
                        											& Failure C1 & 0.04 & 0.94 & 0.02 & 0.00 & 0.00 & 0.7313 & 0.9398 & 0.8225 \\
                        											& Failure C2 & 0.04 & 0.02 & 0.94 & 0.00 & 0.00 & 0.8922 & 0.9415 & 0.9162 \\
                        											& Failure C3 & 0.04 & 0.02 & 0.00 & 0.94 & 0.00 & 0.8942 & 0.9362 & 0.9147 \\
                        											& Failure C4 & 0.04 & 0.04 & 0.09 & 0.02 & 0.81 & 0.9324 & 0.8097 & 0.8667 \\
\bottomrule
\end{tabular}}
\caption{MCF Dataset: MLP Confusion Matrix and Classification Report}
\label{Table 4.4}
\vspace{-10pt}
\end{table}
%%%%%%%%%%%%% Table %%%%%%%%%%%%%
when it is actually "Failure C4", namely in 9\% of the cases. This seems to represent the main weakness of the MLP classifier. 
\\This first suspicion is confirmed when we look at the classification report. In general, the MLP model scores high recall values, all being above the values of the XGB classifier, apart from the last "Failure C4" label. The precision values on the other hand, are all lower as compared to the XGB model, apart from the last "Failure C4" label. This means that the MLP model, in general, has a higher rate of false-alarms, but consequently also detects more failures as compared to the XGB classifier. A last analysis of the f1-scores shows that the XGB model is slightly superior to the MLP model over all labels.
\\Unfortunately, there is no global feature importance map for both – the MLP and LSTM models. It would have been interesting to compare the importances that these distinct classifiers attribute to each of the features.
\\With the evaluation of the predictive performance of the MLP model, we continue to analyse how the LSTM stacks up against the MLP and XGB models in the following.

We finally evaluate the performance of the LSTM classifier, which like the MLP is also a type of neural network. The difference is, however, that the LSTM operates on temporal data, which is enabled through its memory function, as mentioned in section \ref{ch:Content1:sec:Section1:Subsection1:Subsubsection3}. Also, the training dataset of the LSTM only contains the raw input features, without any feature engineering whatsoever. This means that instead of having access to the nineteen features shown in figure \ref{figure 4.5} to learn from, the LSTM only has access to eleven features. These include the 4 telemetry-based features (volt, rotate, pressure and vibration), the 5 types of error codes (without lagging aggregates), and the model type and age.
\\The results for this model can be seen in the following table \ref{Table 4.5}. Regarding the confusion matrix, the model achieves a similar predictive performances as compared to the MLP classifier. 
%%%%%%%%%%%%% Table %%%%%%%%%%%%%
\begin{table}[h]
\centering
\scalebox{0.75}{
\begin{tabular}{cccccccccc}
\toprule
& & \multicolumn{5}{c}{Predicted} & \multicolumn{3}{c}{Classification Report} \\
& &	No Failure & Failure C1 & Failure C2 & Failure C3 & Failure C4	& Precision & Recall & F1-Score \\
\midrule
\parbox[t]{2mm}{\multirow{5}{*}{\rotatebox[origin=c]{90}{Actual}}} 	& No Failure & 1.00 & 0.00 & 0.00 & 0.00 & 0.00 & 0.9989 & 0.9988 & 0.9989 \\
                        											& Failure C1 & 0.06 & 0.92 & 0.02 & 0.01 & 0.00 & 0.8156 & 0.9176 & 0.8636 \\
                        											& Failure C2 & 0.04 & 0.00 & 0.93 & 0.00 & 0.03 & 0.9854 & 0.9254 & 0.9545 \\
                        											& Failure C3 & 0.02 & 0.00 & 0.00 & 0.98 & 0.00 & 0.9443 & 0.9778 & 0.9608 \\
                        											& Failure C4 & 0.08 & 0.02 & 0.00 & 0.01 & 0.89 & 0.9360 & 0.8864 & 0.9105 \\
\bottomrule
\end{tabular}}
\caption{MCF Dataset: LSTM Confusion Matrix and Classification Report}
\label{Table 4.5}
\vspace{-10pt}
\end{table}
%%%%%%%%%%%%% Table %%%%%%%%%%%%%
For the second and third labels (failure of C1 and C2), the performance of the LSTM is slightly below the MLP, but above the XGB. For the fourth and fifth labels (failure of C3 and C4), its prediction accuracy is above the MLP. Compared to the XGB, it scores better in the "Failure C3" label, and slightly worse in the "Failure C4" label.
\\With a comparative look at the classification report, we see that the LSTM model is well-balanced overall, achieving similar scores for both – precision and recall. When comparing the f1-scores of the LSTM with those of the XGB and MLP models, we see that it outperforms both models over all five labels, which is extremely remarkable. As a sanity check, we applied the MLP model to the same dataset as the LSTM, without any feature engineering, and the results were disastrous. The MLP model predicted "No Failure" for every single instance in the test set, thereby performing worse than random guessing. This should further underline the predictive power of the LSTM classifier.
\\Even though the LSTM achieved great results in terms of predictive performance, we do not use it to evaluate the explanation consistency due to its complex architecture, which currently makes it very difficult to apply any explanation methods to it.

Having a broad overview of the evaluation and predictive performance of each of the considered models, we continue to the next section, where we highlight which EMs we apply to which dataset and model.

%% ===========================
\section{Application of Explanation Methods}
\label{ch:Content4:sec:Section3}
%% ===========================
The aim of this brief sub-chapter is to specify exactly, which explanation methods we use in the following section to test our explanation consistency framework. Table \ref{Table 4.6} below provides an overview for this purpose. We start with the SHP dataset and continue with the MCF dataset thereafter.
%%%%%%%%%%%%% Table %%%%%%%%%%%%%
\begin{table}[h]
\centering
\scalebox{0.90}{
\begin{tabular}{lll}
\toprule
Dataset										& Prediction Models 						& Explanation Method		\\
\midrule
\multirow{2}{*}[-3pt]{Seattle House Prices} & \multirow{2}{*}[-3pt]{Extreme Gradient Boosting}& LIME Tabular Explainer	\\
\cmidrule(l){3-3}
                  							&                   						& Tree SHAP 				\\
\midrule
\multirow{4}{*}[-7pt]{Machine Component Failures} & \multirow{2}{*}[-3pt]{Extreme Gradient Boosting}& LIME Tabular Explainer 	\\
\cmidrule(l){3-3}
                 							&                  							& Tree SHAP					\\
\cmidrule(l){2-3}
                  							& \multirow{2}{*}[-3pt]{Multilayer Perceptron} 	& LIME Tabular Explainer	\\
\cmidrule(l){3-3}
                  							&                   						& SHAP Kernel Explainer		\\
\bottomrule
\end{tabular}}
\caption{Overview of Applied Prediction Models and Explanation Methods}
\label{Table 4.6}
\vspace{-10pt}
\end{table}
%%%%%%%%%%%%% Table %%%%%%%%%%%%%

As shown in table \ref{Table 4.6} and already discussed in the model evaluation in \ref{ch:Content4:sec:Section2:Subsection2}, we solely applied the XGB model to the SHP dataset, as it already achieved a high predictive performance, thereby making it redundant to train further models.
\\With regards to the explanation methods discussed in \ref{ch:Content1:sec:Section3:Subsection3:subsubsection2}, we apply both – LIME and SHAP. For the former, we implemented the \textit{Lime Tabular Explainer} \citep{ribeiro2016}, which we use to generate all LIME explanations of the SHP test set. For the latter, we use the \textit{Tree SHAP} algorithm \citep{lundberg2017consistent} to generate all SHAP explanations of the SHP test set. The Tree SHAP algorithm is especially useful, as it enables a rapid estimation of SHAP values for all types of tree ensembles \citep{lundberg2017consistent}. When we say rapid, it means that we can generate over 200k SHAP explanations in a matter of seconds for tree ensembles such as XGB, from our personal experience.

For the MCF dataset, we trained two ML models, namely XGB and MLP, as shown in table \ref{Table 4.6}. For the XGB model, we applied the same two algorithms, LIME tabular explainer and Tree SHAP, as before on the SHP dataset. For the MLP model, we also employed the LIME tabular explainer, however, as the MLP is not a tree ensemble, we could not apply the Tree SHAP algorithm. Consequently, we had to use the SHAP Kernel Explainer, which was computationally very intense. The time that this algorithm takes to generate an explanation mainly depends on the number of features in the dataset. Because the MCF test set has 19 features and over 200k instances, the algorithm requires over five minutes to generate a single explanation for a standard machine with 16GB of RAM and Core i5 processor. We thus generated a random sample of the test set with roughly 5k instances, while maintaining its label distribution. In this sample, we generated all LIME and SHAP explanations, which still took over two weeks, even while using three virtual machines in the cloud. As a side note, to generate an explanation on the same test set for the LIME algorithm only took several seconds, rather than minutes. Nevertheless, it is still interesting to observe whether some changes occur concerning the explanation consistency, by using two different underlying prediction models.
\\In sum, for the XGB model on the MCF dataset, everything remains equal as in the SHP dataset. For the MLP model, we generated the LIME and SHAP explanations on a sample of around 5k instances due to the time complexity of not using a tree ensemble with the SHAP explanation method.

With a good summary of which explanation methods were applied to which prediction models and datasets, we continue to the next section. This upcoming last sub-chapter is the most critical part of the experiments chapter, as it evaluates our developed explanation consistency framework from chapter \ref{ch:Content3}.

%% ===========================
\section{Evaluation of the Explanation Consistency}
\label{ch:Content4:sec:Section4}
%% ===========================
In this last section of the experiments and evaluation chapter, we finally put our developed explanation consistency framework (ECF) to test. Our goal is to find out whether the defined axioms and framework make sense and if it can really be used to compare different explanation methods amongst each other. We choose to proceed with the same order as before, first evaluating the explanation consistency on the SHP dataset, i.e. regression case of the ECF (as defined in \ref{ch:Content3:sec:Section1}), and thereafter on the MCF dataset, i.e. classification case of the ECF (as defined in \ref{ch:Content3:sec:Section2}).

%% ===========================
\subsection{Regression Case – SHP Dataset}
\label{ch:Content4:sec:Section4:Subsection1}
%% ===========================
To evaluate the axioms on the SHP dataset, we can use their exact formulation as defined in \ref{ch:Content3:sec:Section1}, rather than approximations as discussed in \ref{ch:Content3:sec:Section3}. This is so because the SHP dataset is small enough for us to work with distance matrices. The results for the regression case are shown in table \ref{Table 4.7} on the following page. It is structured into two parts: the upper half corresponds to the evaluation of the explanation consistency for the LIME EM, and conversely, the lower half for the evaluation of the SHAP EM. Each of the EMs is evaluated based on how many times for all generated explanations, they either satisfy the established axioms in \ref{ch:Content3:sec:Section1}, or violate it, i.e. fail to comply with it.

A first look at the table reveals that LIME violates axiom 1 for ever single explanation. This is because we can only generate equal explanations with LIME for the same object, if we set a fixed seed value for the algorithm. Otherwise, the explanations are different every single time, due to the probabilistic nature of the EM. SHAP on the other hand fulfils the first axiom in 100\% of all generated explanations on the SHP dataset.
%%%%%%%%%%%%% Table %%%%%%%%%%%%%
\begin{table}[ht]
\centering
%\scalebox{0.90}{
\begin{tabular}{cllll}
\toprule
Explanation Method     	& Axiom 			& \#violated 	& \#satisfied  	& \% satisfied 	\\
\midrule
\multirow{3}{*}[-5pt]{LIME}  	& 1. Identity      	& 5,355        	& 0				& 0\%    		\\
\cmidrule(l){2-5}
		      			& 2. Separability	& 134    		& 28,670,536	& 99.9995\%   	\\
\cmidrule(l){2-5}
			    		& 3. Stability      & 4		        & 5,351			& 99.9252\%		\\
\midrule
\multirow{3}{*}[-5pt]{SHAP}  	& 1. Identity       & 0		        & 5,355			& 100\%    		\\
\cmidrule(l){2-5}
                  		& 2. Separability   & 28        	& 28,670,642  	& 99.9999\%   	\\
\cmidrule(l){2-5}
			            & 3. Stability      & 0         	& 5,355 		& 100\%   		\\
\bottomrule
\end{tabular}%}
\caption{Explanation Consistency Evaluation: SHP Dataset and XGB Model}
\label{Table 4.7}
\vspace{-10pt}
\end{table}
%%%%%%%%%%%%% Table %%%%%%%%%%%%%
\\With regards to the second axiom, if we look at the \% satisfied column, there seems not to be a significant performance difference between LIME and SHAP. When we look at the \#violated column, however, we see that LIME violates the separability axiom almost five times as often as SHAP. Nevertheless, due to the sheer number of checks for this axiom, the value becomes insignificant in relative terms, but in absolute terms, it is still a noteworthy discrepancy.
\\Similarly, the results for the stability axiom, regarding performance differences between LIME and SHAP, also seem insignificant in relative terms. The aggregated numbers for this axiom thus only allow us to gain a shallow understanding. To achieve a deeper level
%%%%%%%%%%%%% Table %%%%%%%%%%%%%
\begin{wraptable}{l}{0.37\textwidth}
\vspace{-5pt}
\scalebox{0.84}{ %To resize the table "zoom-out"
\centering
\begin{tabular}{lcc}
\toprule
Spearman's Rho 	& LIME 		& SHAP \\
\midrule
Minimum:		& -0.0086	& 0.1685	\\
Maximum:		& 0.8001	& 0.8934	\\
Mean:			& 0.4902	& 0.7020	\\
Median:			& 0.5037	& 0.7150	\\
\bottomrule
\end{tabular}}
\vspace{-5pt}
\caption{Stability Axiom: Rho Analysis}
\label{Table 4.8}
\vspace{-10pt}
\end{wraptable}
%%%%%%%%%%%%% Table %%%%%%%%%%%%%
of insight on the performance discrepancy for the stability axiom, we can look at table \ref{Table 4.8}. This table describes the results in a more numerical form, by comparing the minimum, maximum, mean and median values for Spearman's rank correlation coefficients between LIME and SHAP. The minimum for LIME is negative, which is where the four violations of axiom 3 in table \ref{Table 4.7} come from. As expected, the minimum value for SHAP is positive, and well-above LIME's. Furthermore, SHAP also achieves a higher maximum for the stability axiom. Considering that the maximum value for Spearman's Rho is one, the 0.89 achieved by SHAP is remarkable. Regarding explanation stability, this means that there is an almost perfect positive correlation between the similarities of objects and their corresponding explanations. In other words, the more similar two objects are, the more similar their corresponding explanations generated by SHAP, and vice-versa. Finally, we can also see a pronounced difference in the mean and median values between LIME and SHAP. The latter achieves rank correlations that are more than 0.20 higher than the ones of LIME.
\\In sum, SHAP always complies with identity, incurs some minor violations regarding separability, and achieves high scores for stability. LIME, on the other hand, is not capable of complying with the identity axiom, shows greater weakness than SHAP for separability, and achieves a notably lower performance regarding explanation stability.
\\We thereby conclude that for the regression case, the SHAP explanation method has a higher explanation consistency than LIME overall. Consequently, on the basis of what we previously established in \ref{ch:Content1:sec:Section2:Subsection3:Subsubsection1} and \ref{ch:Content3}, SHAP achieves a higher interpretability quality. This means that explanations generated by SHAP should generally be more accurate and easier for humans to intuitively understand than those of LIME, which is consistent with the findings by \cite{lundberg2017}, who reach the same result through a user study.

After successfully evaluating the LIME and SHAP EMs for the regression case of our framework, we continue with the analysis for the classification case in the following.

%% ===========================
\subsection{Classification Case – MCF Dataset}
\label{ch:Content4:sec:Section4:Subsection2}
%% ===========================
From \ref{ch:Content3:sec:Section2} we know that between the regression and classification cases, only the definition of the third axiom changes. However, as the test set of the MCF dataset has over 220k objects, we must use some heuristics, as described in \ref{ch:Content3:sec:Section3}, to compute the explanation consistency results for all three axioms. Furthermore, as mentioned in \ref{ch:Content4:sec:Section3}, this subsection consists in two parts for which we use two different prediction models, namely XGB and MLP. The test set for the XGB model contains the full 220k objects as we can use the Tree SHAP algorithm to quickly compute all explanations. However, the test set for the MLP model only includes roughly 5k objects, as we must use a slower algorithm for the explanation generation.

We start with the XGB part, following the same structure as in \ref{ch:Content4:sec:Section4:Subsection1}. The main difference is that we do not fall back on Spearman's Rho for the evaluation of the third axiom, but rather use the Jaccard similarity and k-means clustering. The results for the classification case and XGB model are shown in table \ref{Table 4.9} below. The table follows the same
%%%%%%%%%%%%% Table %%%%%%%%%%%%%
\begin{table}[ht]
\centering
%\scalebox{0.90}{
\begin{tabular}{cllll}
\toprule
Explanation Method     	& Axiom 			& \#violated 	& \#satisfied  	& \% satisfied 	\\
\midrule
\multirow{3}{*}[-5pt]{LIME}  	& 1. Identity      	& 218,115       & 0				& 0\%    		\\
\cmidrule(l){2-5}
		      			& 2. Separability	& 22,163		& 195,952		& 88.6896\%	  	\\
\cmidrule(l){2-5}
				   		& 3. Stability		& 164		    & 217,951		& 99.9248\%		\\
\midrule
\multirow{3}{*}[-5pt]{SHAP}  	& 1. Identity       & 0		        & 218,115		& 100\%    		\\
\cmidrule(l){2-5}
                  		& 2. Separability   & 312        	& 217,803	 	& 99.8568\%   	\\
\cmidrule(l){2-5}
			            & 3. Stability		& 316         	& 217,799 		& 99,8549\%   	\\
\bottomrule
\end{tabular}%}
\caption{Explanation Consistency Evaluation: MCF Dataset and XGB Model}
\label{Table 4.9}
\vspace{-10pt}
\end{table}
%%%%%%%%%%%%% Table %%%%%%%%%%%%%
familiar layout as table \ref{Table 4.7}, which should ease its interpretation.
\\Once again, LIME fails to comply with the first axiom, due to the already known reasons, whereas SHAP satisfies it in 100\% of all cases.
\\To measure the second axiom, we recurred to the following heuristic: if there are no duplicates in the input data of the EM, then there should be no duplicated explanations, given that the DOF-assumption is not violated, as discussed in \ref{ch:Content3}. LIME, however, generated over 20k duplicated explanations in the total of roughly 220k explanations, thereby violating the second axiom in more than 11\% of all cases. This represents a significant difference compared to the SHAP EM, which still violates the second axiom, however, only 312 times, which translates into less than 0,15\% of all cases.
\\For the analysis of the third axiom, as mentioned before, we used the k-means clustering algorithm as a heuristic, because the test set was too large to apply an agglomerative hierarchical clustering, which operates based on the distance matrix. Nevertheless, we use a semi-informed initialisation procedure for the initial centroids, as described in \ref{ch:Content1:sec:Section1:Subsection2:Subsubsection2}. The results for the stability axiom in table \ref{Table 4.9} seem to imply that LIME achieved better explanation stability than SHAP. Again, this is one of the caveats of using aggregated measures. To gain a deeper insight into the evaluation of the third axiom, we analyse table \ref{Table 4.10}.
%%%%%%%%%%%%% Table %%%%%%%%%%%%%
\begin{wraptable}{r}{0.39\textwidth}
\vspace{-5pt}
%\scalebox{0.95}{ %To resize the table "zoom-out"
\centering
\begin{tabular}{lcc}
\toprule
				& \multicolumn{2}{c}{K-Means}			\\
                \cmidrule(lr){2-3}
Cluster		 	& LIME		& SHAP 						\\
\midrule
No Failure:		& 0.9998	& 0.9993					\\
Failure C1:		& 0.9252	& 0.9296					\\
Failure C2:		& 0.9270	& 0.9579					\\
Failure C3:		& 0.9736	& 0.9929					\\
Failure C4:		& 0.9092	& 1.0000					\\
\bottomrule
\end{tabular}%}
\vspace{-5pt}
\caption{Stability Axiom: Cluster similarities XGB}
\label{Table 4.10}
\vspace{-10pt}
\end{wraptable}
%%%%%%%%%%%%% Table %%%%%%%%%%%%%
This table compares the similarities between object and explanation clusters, based on their constituents. As an example, if there are 10 data points in the "Failure C1" object cluster (i.e. with prediction label "Failure C1"), then we check how many of the 10 corresponding explanations of those 10 data points belong to the "Failure C1" explanation cluster. This means that we fundamentally analyse similarities between object clusters and explanation clusters. Now, if e.g. only 9 corresponding explanations out of the 10 would be in the "Failure C1" explanation cluster, and at the same time, no further explanations from any other data point were to belong to that "Failure C1" explanation cluster, then the similarity between these two would be 90\%, as calculated by the Jaccard similarity.
\\With that in mind, we continue with the analysis of table \ref{Table 4.10}. For the first "No Failure" cluster, LIME achieves a higher similarity than SHAP, which explains the better scores of LIME for the stability axiom in table \ref{Table 4.9}. Even though the similarity difference is only 0.005 for that cluster, it is enough to make LIME appear more powerful in aggregated measures and regarding explanation stability, because the "No Failure" label occurs hundreds of times more often than any of the failure events. It is, however, much more valuable if the explanation stability is high for all labels, rather than only the majority label. Table \ref{Table 4.9} highlights exactly that – even though LIME has a slightly higher similarity score for the "No Failure" cluster, SHAP achieves clearly higher similarity scores for all other clusters, thereby outperforming LIME regarding explanation stability.
\\In sum, SHAP again delivers 100\% compliance with the first axiom, whereas LIME fails to satisfy it. Moreover, the number of violations of the separability axiom for SHAP is notably lower than for LIME. Finally, LIME has fewer violations of the third axiom; however, SHAP achieves higher explanation stability for minority-features.
\\We thereby conclude that for the classification case, while using the XGB as an underlying prediction model, SHAP overall scores a higher explanation consistency than LIME. This is consistent with our previous findings in \ref{ch:Content4:sec:Section4:Subsection2}, and means that explanations generated by the SHAP explanation method have higher interpretability, also for classification.

With the evaluation of the first part of the classification case, we proceed to the second part where we repeat the same analysis as above, however, while using the MLP as the underlying prediction model.

%%%%%%%%%%%%%%%%% Second Case: Classification with MLP %%%%%%%%%%%%%%%%%

The multilayer perceptron, which we use as the underlying prediction model for this second part of the evaluation of the classification case, is a type of neural network rather than a tree-based algorithm. This means that as discussed in \ref{ch:Content4:sec:Section3}, we must use the SHAP kernel explainer instead of the Tree SHAP, which is computationally expensive. We, therefore, only evaluate the ECF on a subset with around 5k data points, instead of 220k. Nevertheless, we hope to find interesting insights and differences regarding explanation consistency, which originate from using two different underlying prediction models.
\\We show the results for the second part of the classification case with the MLP model in table \ref{Table 4.11} below, which also follows the already-known layout.
%%%%%%%%%%%%% Table %%%%%%%%%%%%%
\begin{table}[ht]
\centering
%\scalebox{0.90}{
\begin{tabular}{cllll}
\toprule
Explanation Method     	& Axiom 			& \#violated 	& \#satisfied  	& \% satisfied 	\\
\midrule
\multirow{3}{*}[-5pt]{LIME}  	& 1. Identity      	& 5,456		    & 0				& 0\%    		\\
\cmidrule(l){2-5}
		      			& 2. Separability	& 0				& 5,456			& 100\%	  		\\
\cmidrule(l){2-5}
				   		& 3. Stability		& 26		    & 5,430			& 99.5211\%		\\
\midrule
\multirow{3}{*}[-5pt]{SHAP}  	& 1. Identity       & 0		        & 5,456			& 100\%    		\\
\cmidrule(l){2-5}
                  		& 2. Separability   & 0	        	& 5,456		 	& 100\%   		\\
\cmidrule(l){2-5}
			            & 3. Stability		& 2         	& 5,454 		& 99,9633\%   	\\
\bottomrule
\end{tabular}%}
\caption{Explanation Consistency Evaluation: MCF Dataset and MLP Model}
\label{Table 4.11}
\vspace{-10pt}
\end{table}
%%%%%%%%%%%%% Table %%%%%%%%%%%%%
\\The identity axiom is, for the third time, fully satisfied by the SHAP explanation method, and 0\% by LIME, due to the already-discussed reasons. We emphasise the shortcomings of LIME for the first axiom as its main weakness.
\\With respect to the separability axiom, LIME and SHAP both achieve a 100\% satisfied score. This is a notable difference from the first part of the classification case, where LIME failed to comply with the second axiom in more than 11\% of all cases. This finding further suggests that LIME starts to generate more duplicated explanations when the total amount of explanations to compute increases. The same also applies to SHAP, however, in significantly smaller proportions.
\\To produce the evaluation scores for the third axiom, we used the AGNES hierarchical clustering method described in \ref{ch:Content1:sec:Section1:Subsection2:Subsubsection1}, as the data was small enough for this algorithm to work feasibly. We first applied the single linkage function in combination with AGNES, but quickly ran into the chaining problem described in \cite{hartigan1981}. This means that we ended up with one explanation cluster containing all explanations and four singletons. We thus had to use a different linkage function and recurred to the ward method described in \ref{ch:Content1:sec:Section1:Subsection2:Subsubsection1}, which is based on variance rather than distance \citep{ward1963}. With this linkage function, the AGNES clustering algorithm produced well-formed clusters of higher quality, as measured by the Calinski-Harabaz score \citep{calinski1974}. Moreover, to have a sanity check, we also applied the k-means clustering algorithm without random initialisation, as in the first part.
\\When we look at table \ref{Table 4.11}, we see that in relative terms there also seems to only be an insignificant difference between LIME and SHAP regarding the performance for the third axiom. It is noteworthy, however, that in absolute terms SHAP only violates the stability axiom in 2 cases, and LIME in 26 cases, which is 13 times more often. This represents a further difference from the first part of the classification case, where LIME incurred fewer violations of the third axiom. To gain an even deeper insight into the evaluation of the stability axiom, we can look at table \ref{Table 4.12}, which is the equivalent of table \ref{Table 4.10}, but for the second part of the classification case. As mentioned, we used two clustering algorithms, namely AGNES and k-means, which is why table \ref{Table 4.12} also features two comparisons
%%%%%%%%%%%%% Table %%%%%%%%%%%%%
\begin{wraptable}{l}{0.55\textwidth}
\vspace{-5pt}
\scalebox{0.95}{ %To resize the table "zoom-out"
\centering
\begin{tabular}{lcccc}
\toprule
				& \multicolumn{2}{c}{K-Means} & \multicolumn{2}{c}{AGNES}	\\
				\cmidrule(lr){2-3} 		\cmidrule(lr){4-5}
Cluster		 	& LIME		& SHAP		& LIME 		& SHAP 				\\
\midrule
No Failure:		& 0.9989	& 0.9998	& 0.9974	& 0.9998			\\
Failure C1:		& 0.7368	& 0.9697	& 0.5952	& 0.9697			\\
Failure C2:		& 0.9333	& 1.0000	& 0.8913	& 1.000				\\
Failure C3:		& 0.9375	& 1.0000	& 0.5306	& 1.000				\\
Failure C4:		& 0.7826	& 1.0000	& 0.7727	& 1.000				\\
\bottomrule
\end{tabular}}
\vspace{-5pt}
\caption{Stability Axiom: Cluster similarities MLP}
\label{Table 4.12}
\vspace{-10pt}
\end{wraptable}
%%%%%%%%%%%%% Table %%%%%%%%%%%%%
between LIME and SHAP. For both clustering algorithms, SHAP achieves higher cluster similarities  than LIME over all clusters, which is a notable difference as compared to table \ref{Table 4.10} in the previous part. A further curiosity that we highlight in table \ref{Table 4.12} is that SHAP's explanation stability does not change, no matter if we use AGNES or k-means as clustering algorithms. For LIME on the other hand, the cluster similarities are higher when the k-means algorithm is used. This is a further indication that SHAP's explanation stability is greater than LIME's.
\\In sum, SHAP again fully complies with the identity axiom, as opposed to LIME. Moreover, both explanation methods achieve a 100\% satisfied score for the second axiom. Finally, for the stability axiom SHAP incurs markedly less violations than LIME, and also achieves higher cluster similarities over all clusters and clustering methods.
\\Herewith, we conclude that also in the second part of the classification case, SHAP generally achieved higher explanation consistency scores than LIME. Again, this is in line with our two previous findings, and a strong indication that explanations generated by SHAP, in general, have a higher interpretability quality than those of LIME.

Having finished the evaluation of our explanation consistency framework, in the next and last sub-chapter of the experiments, we briefly elaborate on and summarise the strengths and weaknesses of each of the two used explanation methods.

%% ===========================
\subsection{Strengths and Weaknesses of LIME and SHAP}
\label{ch:Content4:sec:Section4:Subsection3}
%% ===========================
In this last evaluation subsection of the explanation consistency, we summarize the main findings regarding strengths and weaknesses of each of the two used explanation methods – LIME and SHAP. We start with LIME and analyse SHAP after that.

The local surrogates model (LIME) may appear inferior regarding explanation consistency from our findings in \ref{ch:Content4:sec:Section4}. However, as measured by our ECF, LIME's performance was not weak. In fact, LIME also features many strengths compared to SHAP. First and most importantly, LIME is computationally efficient and can be applied to large datasets. In practice, this is often not a negligible factor to consider, as it can change the decision on what explanation method to use. Moreover, LIME returns an actual prediction model, which enables users to prompt it with different values and see how these affect the explanation. Lastly, LIME is applicable to all types of data – tabular, image and text.
\\LIME does, however, also have some shortcomings. First of all, it was never able to fulfil the identity axiom, due to its random sampling nature. To elucidate why failing to comply with the first axiom is very undesirable we use the following example. Imagine that we prompt LIME twice to explain why a specific house was predicted to cost 550k. The first time, the explanation says that the two most influential factors for that house price were the above-average size of the house and the existence of a swimming pool. The second time, the explanation tells us that the two most influential factors were the excellent location of the house and the existence of a garage. Situations like the one described occur with explanations generated by LIME, which is highly confusing and unintuitive for a human to understand.
\\Furthermore, several parameters can be adjusted in LIME such as the kernel width (i.e. size of the neighbourhood), the type of distance measure to use, and how many points to sample in the neighbourhood. The choice of these parameters heavily influences the explanations generated by LIME. Choosing an appropriate set of parameters requires some degree of specialised knowledge, thereby making this EM unsuitable to be used by non-technical people, or people without specialised domain knowledge.
\\Finally, LIME assumes that a linear model can locally approximate a complex non-linear ML model. There is, however, no solid theory, research or evidence that explains why this should work or not \citep{molnar2018}.

The SHAP explanation method features some strengths where shortcomings for LIME exist and vice-versa. Most importantly, SHAP is the only EM with a solid theory that has been around for a long time, as discussed in \ref{ch:Content1:sec:Section3:Subsection3:subsubsection2}. It is provably the only method that satisfies the three properties of feature attributions, namely: dummy, symmetry, and efficiency. This means that SHAP may be the only compliant EM to use in situations that demand explainability by law, such as the GDPR \citep{molnar2018}.
\\Furthermore, SHAP scored very high in our ECF, which is probably because it has a solid theoretical foundation. Scoring high values in the ECF does not only stand for better human interpretability, but also for higher accuracy. If we take the example of the identity axiom, if an EM does not fulfil this condition, then it cannot possibly be accurate. These findings derived in our evaluation of the explanation consistency reach the same conclusion as the findings by \cite{lundberg2017}.
\\Lastly, we highlight the computational efficiency of the Tree SHAP algorithm. This means that when we use tree ensembles as prediction models, it may be worthy to use SHAP as an explanation method. The Tree SHAP algorithm enables exceptionally high speeds in generating thousands of explanations, and simultaneously, as discussed before, offers higher accuracy and interpretability for the explanations compared to LIME.
\\This last strength is simultaneously the most daunting weakness of SHAP, as soon as we use any prediction model other than a tree ensemble. Because the method follows an exact approach that generates all possible coalitions of features ($2^k$) to determine their exact attributions, it can be computationally costly to apply, especially when there are many features. Hence for most practical applications, approximated explanations such as the ones generated by LIME would probably suffice.
\\Moreover, as described by \cite{molnar2018}, Shapley values can easily be misinterpreted, as these do not express total feature contributions. They instead explain how to get from the mean prediction to the actual prediction \citep{lundberg2017}. Applied to our housing price example, this means the following: if e.g. the Shapley value for latitude is 50k, then these 50k represent the increase from the "average contribution" of the latitude feature to this current prediction. To get the total feature contribution of the latitude feature for this particular data point, we would thus have to sum the "average contribution" of the latitude feature with the Shapley value of the latitude feature for this particular data point that we are looking at.
\\Finally, SHAP does not return a prediction model unlike LIME, which is a significant shortcoming, as the user is not able to quickly change some feature values and see how these affect the explanation.

With a good understanding and overview of the strengths and weaknesses of each of the two explanation methods – LIME and SHAP, we conclude the experiments chapter. In the following and final chapter, we finalise this thesis project by discussing the contribution of this work, limitations and future directions for research.
%\include{text/evaluation} #Milo
%% ===========================
%% ===========================
\chapter{Discussion and Conclusion}
\label{ch:Conclusion}
%% ===========================
%% ===========================

\hspace{\parindent}In this last chapter of our thesis, we discuss and conclude our work, starting with a review of the context and scope of our research.
\\The fundamental starting point of this thesis project was the problem of loss in interpretability, which occurs when we use and apply black box machine learning models. The complex and intricate inner workings of these black boxes make them less interpretable and in some cases even completely opaque regarding human understandability. Nevertheless, the reward of using black box models – a high-performance potential regarding predictive power – is often too tempting to pass.
\\Driven by the needs of industry and practice, the research community has recognised the necessity to develop methods that explain predictions made by these black boxes, to recover some of the lost interpretability. Hence, over the past two to three years, numerous attempts and different so-called explanation methods have been developed and published by research to address this problem. These EMs are useful tools as they help development engineers and other project stakeholders to understand and follow predictions of even the most complex black box ML models such as neural networks.
\\With the proliferation of many different types of EMs and new ones being frequently released, it remains often unclear, however, which EM is superior concerning explanation quality, or best-suited for a particular situation. Moreover, we noticed that no method exists to assess the quality of different EMs amongst each other in terms of their strengths and weaknesses. We thereby identified our research gap – developing a method to asses the quality of different EMs. Over the course of this thesis, we developed such a framework that enables us to rank EMs regarding the desired interpretability goals of accuracy, understandability and efficiency as discussed in \ref{ch:Content1:sec:Section2:Subsection3:Subsubsection1}. By means of this framework, it is possible to determine which is the most understandable and accurate explanation method for a specific situation.

The remainder of this chapter continues by highlighting and summarising the most important research results that we obtained with our explanation consistency framework and experiments. Furthermore, we elaborate on our contribution to the literature in the interpretable machine learning domain. After that, we point out the limitations of our chosen approach and the developed framework, thereby setting future research impulses. Lastly, we conclude this thesis by revisiting and answering our initial research questions.

%% ===========================
\section{Results Summary}
\label{ch:Conclusion:Section1}
%% ===========================
In the experiments chapter, we evaluated our developed explanation consistency framework for both cases – classification and regression. To achieve this, we prepared a dataset for each case and trained a range of ML models on them, selecting the ones with the highest unbiased prediction performance for each dataset. As guidance and reference for the data and model preparation, we used the CRISP-DM process, introduced in \ref{ch:Introduction1:sec:Section2:Subsection2} (see appendix section \ref{Appendix:Section:C} for the codebase of this thesis project). After that, we applied two explanation methods, namely LIME and SHAP, to each of the two datasets.

We start by reviewing the results of the regression case. Regarding the identity axiom, SHAP outperformed LIME, by achieving a 100\% satisfied score, whereas LIME incurred a 100\% violation. As we have seen before in \ref{ch:Content1:sec:Section3:Subsection3:subsubsection2:Paragraph1}, LIME cannot possibly fulfil the first axiom, due to the random sampling process involved in the generation of the neighbourhood, based on which the linear approximative model is trained \citep{ribeiro2016}. We emphasise that this is LIME's greatest weakness, playing a significant role in the loss of explanation consistency as defined by our axioms in \ref{ch:Content3}. Not fulfilling this axiom means that LIME generates distinct explanations for the same input (object and prediction) if prompted multiple times, which is confusing for the user and far from being consistent.
\\SHAP also achieved a higher score for the separability axiom, incurring almost five times fewer violations than LIME. Failing to comply with this axiom means that for some cases, the EMs generate the same explanation to different objects, which is also unintuitive and thus undesired. Nevertheless, in relative terms, both analysed EMs achieved high scores for the second axiom.
\\Finally the stability axiom was never violated by SHAP for the regression case, and LIME also only incurred four violations in total. This means that both EMs achieved high scores for this axiom, whereas SHAP once again, achieved higher scores than LIME. By looking at the Rho analysis in table \ref{Table 4.8}, we can study this axiom in greater detail. We highlight that SHAP achieved a median rank correlation of almost 0.72, which is a remarkable result, and about 0.20 better than for LIME. This high positive correlation indicates that SHAP generates very similar explanations for similar objects, thereby achieving high explanation consistency.
\\In sum, we conclude the analysis of the regression case with the central finding that SHAP achieves a higher explanation consistency over all defined axioms. Consequently, based on our framework we argue that explanations by SHAP are more interpretable and accurate than those of LIME. This result is consistent with the findings by \cite{lundberg2017}, who reach the same conclusion by means of a user study.

For the classification case, we applied two prediction models to the same classification dataset, namely XGB and MLP, which allowed us to validate our framework further. For the first axiom, as in the regression case, SHAP achieved a 100\% satisfied score for both prediction models, whereas LIME, once again, failed to comply with this axiom.
\\For the separability axiom, the results are different for the two underlying prediction models. For the XGB model, SHAP achieved a distinctively higher score than LIME, reaching almost 100\% satisfaction, whereas LIME only complies with the second axiom in roughly 89\% of all cases. For the MLP model, LIME and SHAP both incur no violations regarding separability, thus achieving a 100\% satisfied score. It is noteworthy, however, that the underlying test set on which the explanations were generated for the MLP model, is significantly smaller than the underlying test set of the XGB model. With this result, we derived the hypothesis that both – LIME and SHAP start to generate more duplicate explanations, with an increasing total number of explanations to generate. For LIME, however, the increase in duplicates is significantly more accentuated than for SHAP.
\\For the third axiom, the results for different prediction models were mixed again regarding compliance. For the XGB model, LIME incurred fewer violations than SHAP, whereas, for the MLP model, the opposite was the case. By analysing the results of the third axiom in greater detail, we found that LIME only incurs fewer violations for the XGB model in the majority label. For all minority labels, SHAP achieved significantly higher scores than LIME. Moreover, when we used the MLP as underlying prediction model, we found that SHAP outperformed LIME in all labels – majority and minority, concerning explanation stability.
\\In sum, we thereby also conclude that for the classification case, SHAP achieved a higher explanation consistency than LIME overall. Based on our framework, we thus reach the same conclusion as for the regression case and argue that explanations generated by SHAP are more interpretable and accurate than those of LIME. This is consistent with the findings of the user study by \cite{lundberg2017}.

With a good overview of the achieved results in this work, we continue discussing its limitations and future directions for research in the following.

%% ===========================
\section{Limitations and Future Work}
\label{ch:Conclusion:Section2}
%% ===========================
In this section, we discuss and highlight two different types of limitations of our research: limitations of the approach, and limitations of the developed explanation consistency framework itself. Moreover, we explore topics that could be covered in future research projects in this domain.

Regarding the approach, there are mainly three limitations that we emphasise in the following. First, we only tested our ECF with two explanation methods, namely LIME and SHAP. It would be interesting to apply the ECF to more EMs, maybe also to some which are not model-agnostic, to analyse if any differences occur in the results.
\\Second, for the evaluation of our framework, we solely relied on two datasets: one for the regression case, and the other for the classification case. To further strengthen the validity of the ECF, it is necessary to evaluate it on more datasets for each case.
\\Finally, the third limitation is that we did not evaluate our framework with real people. We argue that achieving a higher explanation consistency implies a higher interpretability quality; however, even if the formulated axioms are intuitive and consistent with expectations regarding explanations, this last step is necessary to confirm the validity of the ECF fully. This type of validation could also uncover further insights on desirable characteristics of explanations in general, which we could then translate into further (sub-)axioms.

Concerning the limitations of the developed framework, we also highlight the three main ones. First, possibly more restrictive (sub-)axioms could have been defined, as especially in relative terms, the axioms sometimes returned very close results for LIME and SHAP.
Second, the exact validation of explanation consistency is computationally complex. In section \ref{ch:Content3:sec:Section3} we defined heuristics to validate each of the axioms for situations with big data, however, when we use approximations the meaningfulness of the results may suffer.
Finally, the third limitation of the ECF is that we may lose some valuable information by using aggregated distances. When e.g. computing the distance between two explanations, we only look at the final distance value. However, distances between individual feature attributions or importances may also be strong indicators to measure explanation consistency. We could possibly even define more specific (sub-)axioms based on these individual distances rather than on aggregated ones.

We also see the different types of limitations discussed above as opportunities or directions for future research. First of all, it would be interesting to conduct a more extensive experiment, where a more significant number of datasets and explanation methods are used. Also, it would be crucial to validate these results or the results of a more extensive study with real people in the form of field or lab experiments. Through such a study, the explanation consistency framework could be further validated, improved and extended.
\\Lastly, researchers could also take completely different approaches to compare the explanation quality amongst different EMs, by e.g. using a variance-based rather than a distance-based approach as in this thesis.

Having a good overview of the limitations of the present work, and further research gaps that can be explored in subsequent research, we continue to the next section, where we emphasise the contribution of our work.

%% ===========================
\section{Contribution}
\label{ch:Conclusion:Section3}
%% ===========================
After discussing the results, limitations and future directions for research, we continue by underlining the contribution of this thesis project in what follows.
\\The developed explanation consistency framework represents some first research in this sub-domain. Moreover, the results obtained in our experiments, which are consistent with related research, speak for the feasibility of the ECF to be used to compare different EMs and their generated explanations concerning interpretability quality. This was the main goal of this research and simultaneously represents its biggest contribution. We further highlight that the ECF is applicable to any EM, which operates with absolute or relative feature importance scores, and is, moreover, employable to datasets with either – classification or regression target variables.
\\Apart from the contribution that our research makes to the domain of interpretable machine learning, we hope that we can also motivate and persuade other researchers to invest their time and efforts in this sub-domain. As we have seen in the limitations, the developed ECF is far from being perfect and much work remains to be done. Nevertheless, we firmly believe that our research raises awareness for this area, and on the other hand, provides a solid foundation on which further research can be carried out.

%% ===========================
\section{Conclusion}
\label{ch:Conclusion:Section4}
%% ===========================
In this last sub-chapter of our thesis, we finally revisit our overall research question and the two refined research questions as defined in \ref{ch:Introduction1:sec:Section2:Subsection1}, providing answers to them. Our overall research question stated the following:

\vspace{-7pt}
\begin{center}
\textit{"How can an adequate framework be designed, which enables to assess the quality of explanation methods, used to explain predictions made by black box models?"}
\end{center}
\vspace{-7pt}

\noindent Moreover, our first refined question asked how a good approach to compare different explanation methods can be characterized, and the second refined question asked how we can evaluate our developed framework in terms of yielding meaningful results.
\\After finalising our thesis, and through the validation of our developed explanation consistency framework, we mainly reach two conclusions. First, axioms are a feasible vehicle to compare explanation methods amongst each other. Second, the validity and usefulness of our framework, to measure and assess the quality of different EMs, is supported by the meaningful results of our experiments, and further confirmed by related research.
\\Axioms as defined in \ref{ch:Content3}, were findable, applicable (as we have seen in \ref{ch:Content4}) and common practice (as we have seen e.g. in \ref{ch:Content1:sec:Section4}). Moreover, by means of our axiomatic framework, we reached the same conclusion as related research, namely that the SHAP explanation method is superior to LIME, in terms of interpretability \citep{lundberg2017}. We thereby conclude that the design of our framework overall is adequate.

Finally, there are certainly some limitations in our approach and framework, but we strongly believe that it is a solid basis for this new research sub-domain. We thereby hope to inspire and convince other researchers to join and invest their expertise and efforts in this emerging field.
%% ===========================
%% ===========================
\iflanguage{english}
{\chapter{Declaration}}
{\chapter{Erklärung}}
\label{ch:Declaration}
%% ===========================
%% ===========================

Ich versichere hiermit wahrheitsgemäß, die Arbeit selbstständig verfasst und keine anderen als die angegebenen Quellen und Hilfsmittel benutzt, die wörtlich oder inhaltlich übernommenen Stellen als solche kenntlich gemacht und die Satzung des Karlsruher Instituts für Technologie (KIT) zur Sicherung guter wissenschaftlicher Praxis in der jeweils gültigen Fassung beachtet zu haben.

\vspace*{1cm}
\hspace*{4cm} Karlsruhe, den \submissiontime \hspace*{0.5cm}\hrulefill \\
\hspace*{10.5cm} \myname \\

%% ----------------
%% |   Appendix   |
%% ----------------
% IM Style: No additional blank page
% \cleardoublepage

%% ===========================
%% ===========================
\appendix
\iflanguage{english}
{\addchap{Appendix}}	% english style
{\addchap{Anhang}}	% german style
%% ===========================
%% ===========================

\section{LIME: Formal Model Definition}
\label{Appendix:Section:A}
In section \ref{ch:Content1:sec:Section3:Subsection3:subsubsection2} we have introduced LIME and discussed its explanation generation process. The goal of this appendix section is thus to briefly describe the mathematical model behind LIME, to gain a more in-depth understanding of this method.

As defined by \cite{ribeiro2016}, explanations generated by LIME are obtained through the following:
\[
\xi(x) = \operatorname*{argmin}_{g \in G} \mathcal{L}(f, g, \pi_x) + \Omega(g),
\]
\noindent whereas $\xi(x)$ represents the explanation for object $x$. The other part of the equation is composed of $\mathcal{L}$, which is a measure of unfaithfulness for the approximation (similar to a loss function), and  $\Omega$, which is a complexity measure (complexity of the generated explanation). The explanation is thereby defined as a model $g \in G$, where $G$ is a class of potentially interpretable models, such as linear models (linear or logistic regression) or decision trees. Now, the goal of our explanation is to minimise both – the unfaithfulness $\mathcal{L}(f, g, \pi_x)$ and the level of complexity $\Omega(g)$, to remain interpretable to humans. The function $f$ corresponds to our prediction model and $\pi_x(z)$ is a proximity measure between the instance $x$ that we want to explain and instances $z$ of the locality around $x$, which is defined by $\pi_x$. So the function $\pi_x$ basically performs the weighting step as described in \ref{ch:Content1:sec:Section3:Subsection3:subsubsection2}, and uses an exponential kernel function for that purpose.

The model described above constitutes the main idea and essence of LIME. For more specific information regarding this explanation method, we recommend the original paper by \cite{ribeiro2016}, or the {\href{https://github.com/marcotcr/lime}{GitHub repository of LIME}} for further details on its implementation.

\section{SHAP: Formal Definition of the Shapley Values}
\label{Appendix:Section:B}
In this appendix section, we further explore the Shapley value model from cooperative game theory \citep{shapley1953}, which is the basis of the SHAP explanation method introduced in \ref{ch:Content1:sec:Section3:Subsection3:subsubsection2}. The goal of this section is thus to briefly describe the mathematical concept of Shapley values, to gain a deeper understanding of the SHAP EM.

As defined by \cite{lundberg2017}, the classic Shapley value estimation is obtained as follows:
\[
\phi_i = \sum\limits_{S \subseteq F \setminus \{i\}} \frac{|S|!(|F|-|S|-1)!}{|F|!}[f_{S \cup \{i\}}(x_{S \cup \{i\}}) - f_S(x_S)],
\]
\noindent whereas $\phi_i$ corresponds to the Shapley value of the $i$-th feature. $F$ thereby represents the set of all features, and $S \subseteq F$ corresponds to all feature subsets on which the model needs to be retrained. The method assigns an importance value to each feature, which represents the effect of that feature on the model prediction when this feature is included. To achieve this, we need to train two models: one that contains that feature ($f_{S \cup \{i\}}$), and the other which does not ($f_S$). Thereafter, we compare the predictions of our two models for the current input $x_S$, with input features in the set $S$, by subtracting them from each other: $f_{S \cup \{i\}}(x_{S \cup \{i\}}) - f_S(x_S)$. Now, because the effect of holding back a feature depends on the other features in the model, we must compute the other differences for all possible subsets $S \subseteq F \setminus \{i\}$. Finally, the Shapley values are then computed as the weighted averages of all possible differences and used as feature importances in SHAP explanations.

With this further elaboration, it becomes more evident why the determination of Shapley values is computationally very expensive, as the complexity increases exponentially with $F$. The SHAP explanation method applies sampling approximations to the model above and further approximates the effect of removing a variable from the model, which substantially decrease its complexity \citep{lundberg2017}. Nevertheless, the method remains computationally costly for any prediction model not belonging to the family of tree ensembles \citep{lundberg2017consistent}.

The model described above extended with some clever approximations, constitutes the main idea behind SHAP. For more specific information regarding this explanation method, we recommend the original paper by \cite{lundberg2017}, or the {\href{https://github.com/slundberg/shap}{GitHub repository of SHAP}} for further details on its implementation.

\section{Codebase and Datasets}
\label{Appendix:Section:C}
The entire codebase, consisting of Python Jupyter Notebooks, which was implemented and developed over the course of this master thesis is publicly available under the MIT license, and can be consulted {\href{https://github.com/MHonegger/explanation_consistency}{on this GitHub repository}}\footnote{\url {https://github.com/MHonegger/explanation_consistency}}.

The datasets used for the experimental validation of the explanation consistency framework are also publicly available on the Kaggle platform. The house sales in King County dataset (Seattle house prices) can be found {\href{https://www.kaggle.com/harlfoxem/housesalesprediction}{here}}\footnote{\url {https://www.kaggle.com/harlfoxem/housesalesprediction}}, and the dataset for predictive maintenance (machine component failures)  can be found {\href{https://www.kaggle.com/yuansaijie0604/xinjiang-pm}{here}}\footnote{\url {https://www.kaggle.com/yuansaijie0604/xinjiang-pm}}.

%% --------------------
%% |   Bibliography   |
%% --------------------
\cleardoublepage
\phantomsection
\addcontentsline{toc}{chapter}{\bibname}

% IM Style
\iflanguage{english}
{\bibliographystyle{apacite}}	% english style
{\bibliographystyle{apacite}}	% german style

%

% Informatik-Style
%\iflanguage{english}
%{\bibliographystyle{IEEEtranSA}}	% english style
%{\bibliographystyle{babalpha-fl}}	% german style
												  
% Use IEEEtran for numeric references
%\bibliographystyle{IEEEtranSA})

\bibliography{thesis}

\end{document}